%% file: paper.tex
\documentclass[runningheads]{llncs}

\usepackage{eccv}

\usepackage{eccvabbrv}

\usepackage{graphicx}
\usepackage{amsmath}
\usepackage{amssymb}
\usepackage{booktabs}
\usepackage{soul}
\usepackage[dvipsnames]{xcolor}
\usepackage{colortbl}
\usepackage{lipsum}
\usepackage{subcaption}
\usepackage{multirow}
\usepackage{enumitem}
\usepackage{wrapfig}
\usepackage{nicefrac}
\usepackage{makecell}

\input{tex/plots}

\usepackage[accsupp]{axessibility}  %

\usepackage{hyperref}

\usepackage{orcidlink}

\begin{document}

\title{AMES: Asymmetric and Memory-Efficient Similarity Estimation for Instance-level Retrieval}

\author{Pavel Suma\inst{1}\and
Giorgos Kordopatis-Zilos\inst{1} \and
Ahmet Iscen\inst{2} \and
Giorgos Tolias\inst{1}}

\titlerunning{AMES: Asymmetric and Memory-Efficient Similarity}
\authorrunning{P.~Suma et al.}

\institute{$^1$VRG, FEE, Czech Technical University in Prague \quad
$^2$Google DeepMind}

\input{tex/abbrev}
\input{fig/pgfplotsdata}

\maketitle
\begin{figure}[h!]
\vspace{-20pt}
  \centering
  \begin{tabular}{c@{\lsp}c}
  \includegraphics[clip, trim=0 0 0 20, height=0.25\linewidth]{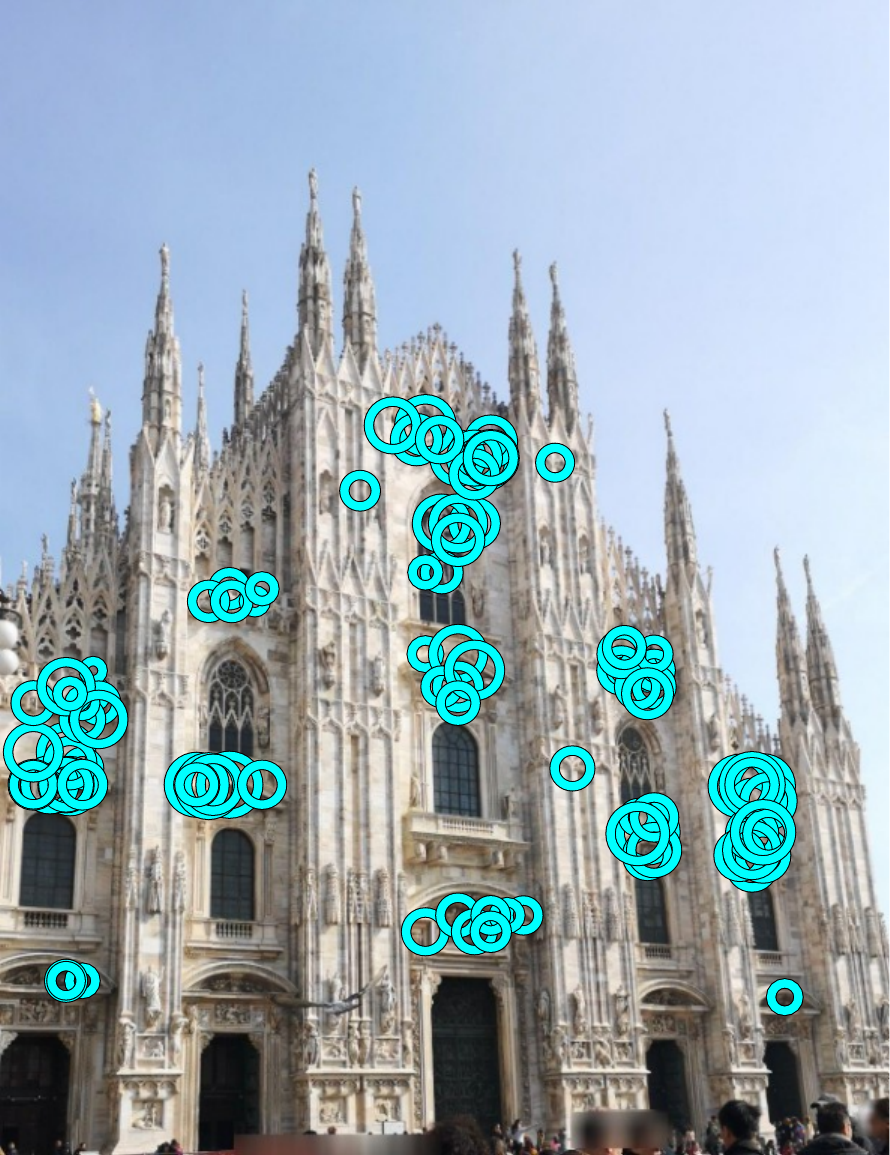}&
  \includegraphics[clip, trim=0 0 0 0, height=0.25\linewidth]{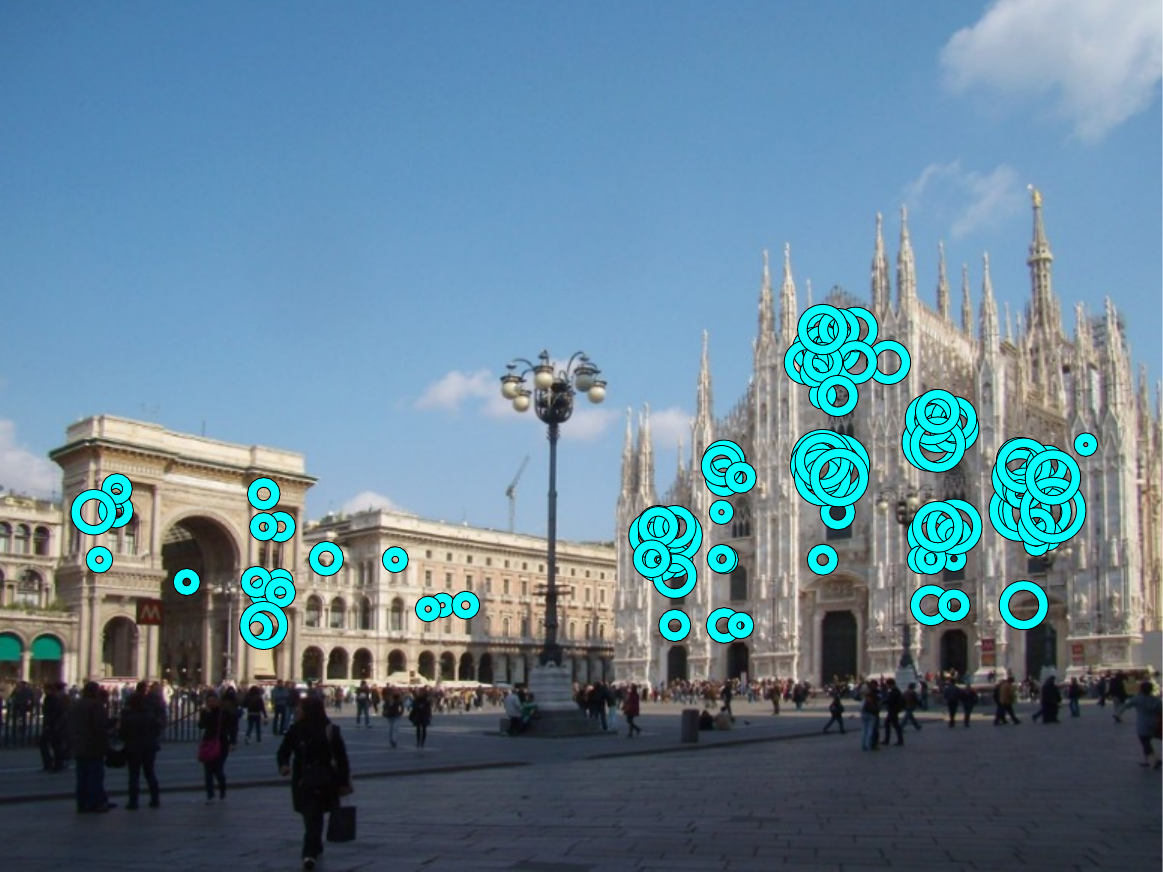}\\
  \includegraphics[clip, trim=0 0 0 20, height=0.25\linewidth]{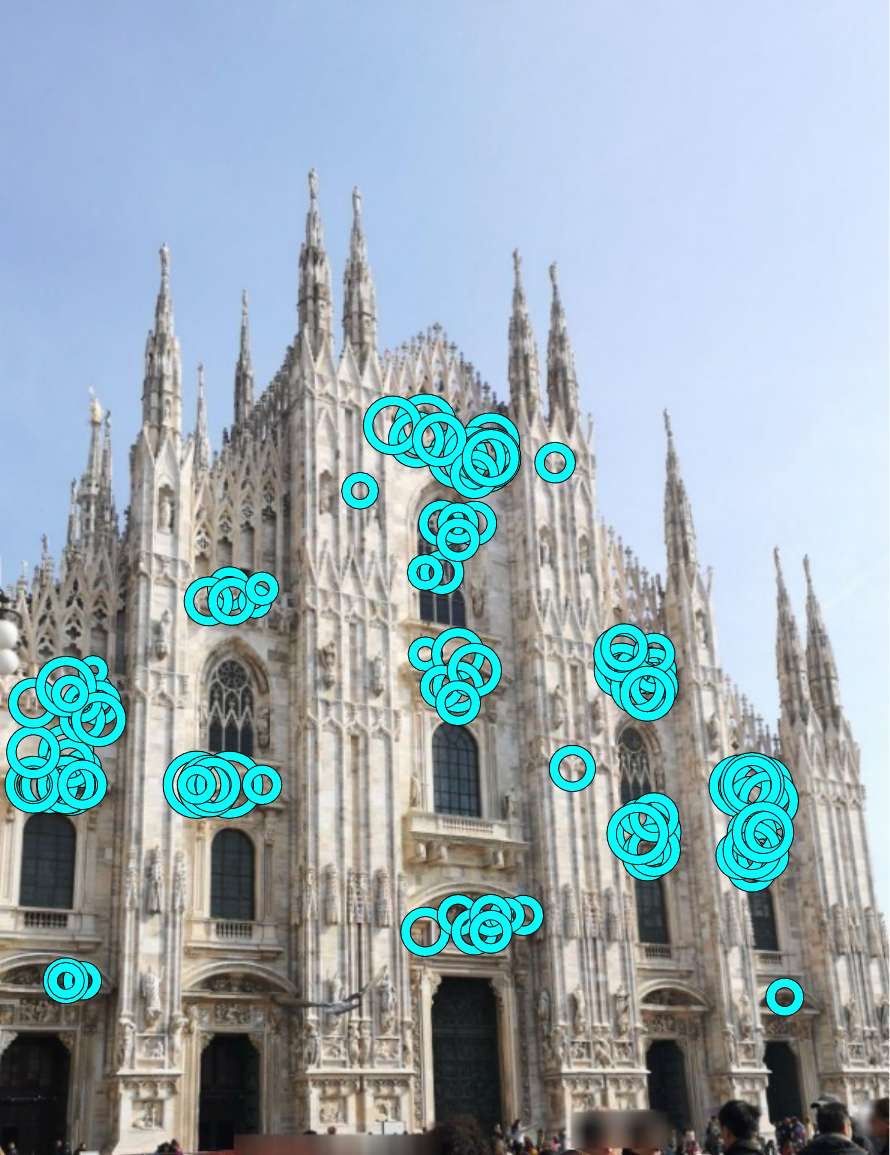}&
  \includegraphics[clip, trim=0 0 0 0, height=0.25\linewidth]{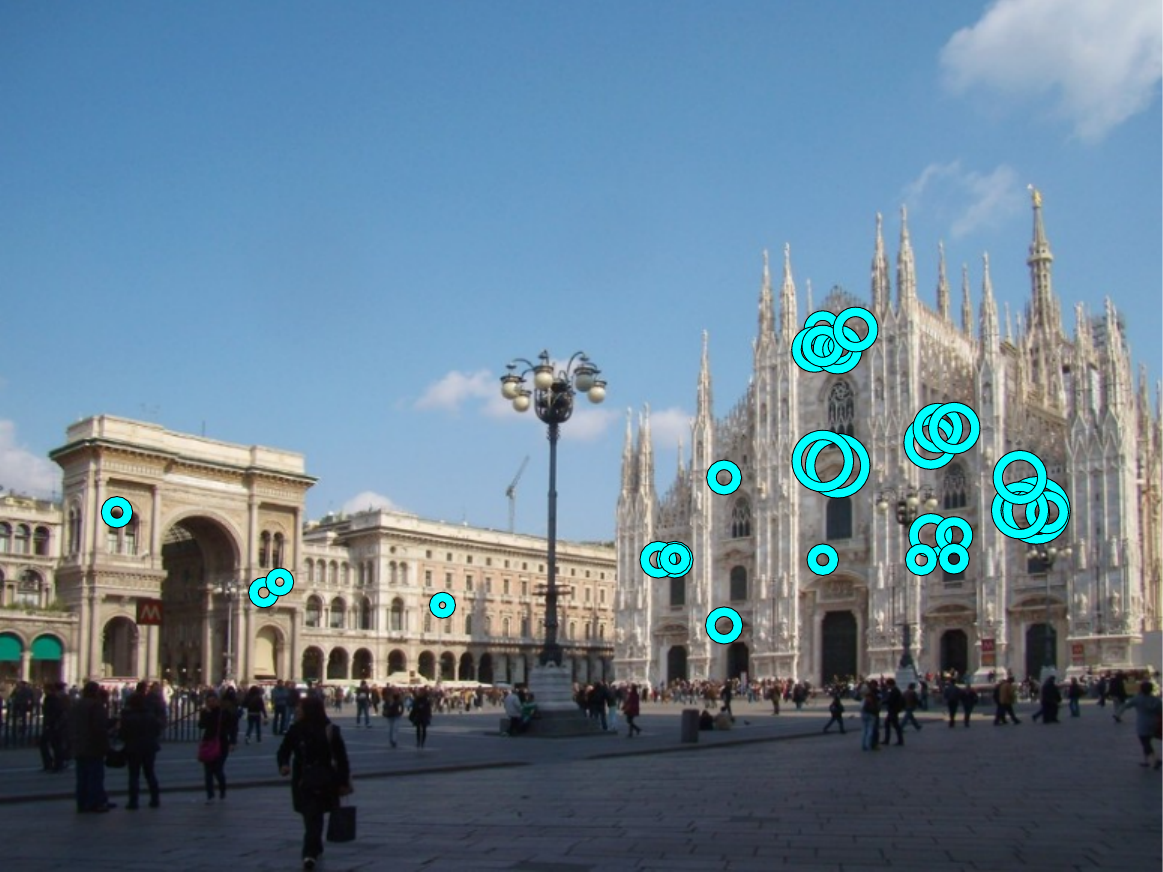}
  \end{tabular}
  \vspace{-10pt}
  \caption{\ours estimates the image-to-image similarity based on local descriptor sets.
  Top: 100 (query) \vs 100 (database) descriptors. 
  Bottom: memory-efficient and asymmetric variant with 100 \vs 30 local descriptors.
  Circle size reflects descriptor importance within \ours; descriptors of the common object get higher importance.
  \label{fig:teaser}
\vspace{-20pt}
}
\end{figure}

\begin{abstract}
\input{tex/abstract}

\end{abstract}

\input{tex/intro}
\input{tex/related}
\input{tex/method}
\input{tex/exp}

\input{tex/conclusions}

\smallskip\small\noindent\textbf{Acknowledgments:}
This work was supported by the Junior Star GACR GM 21-28830M and the Czech Technical University in Prague grant No. SGS23/173/OHK3/3T/13. \tolerance=10000

{\small
\bibliographystyle{splncs04}
\bibliography{tex/bib}
}

\clearpage
\newpage
\appendix
\input{tex/supplementary}

\end{document}

%% file: tex/plots.tex
\usepackage{tikz}
\usetikzlibrary{arrows,shapes,calc,matrix,fit,backgrounds}
\usepackage{pgfplots}
\usepackage{pgfplotstable}
\pgfplotsset{compat=1.9}
\usepgfplotslibrary{fillbetween}
\usepackage{xstring}

\tikzset{every mark/.append style={solid}}
\pgfplotsset{%
	grid=both, width=\columnwidth, try min ticks=5,
	every axis/.append style={font=\scriptsize},
	every axis plot/.append style={thick,mark=none,mark size=1.2,tension=0.18},
	legend cell align=left, legend style={fill opacity=1},
}

\pgfplotsset{
    legend image with text/.style={
        legend image code/.code={%
            \node[anchor=west] at (3pt,0pt) {#1};
        }
    },
}

\pgfplotsset{
	dash/.style={mark=o,dashed,opacity=0.7},
	dott/.style={mark=o,dotted,opacity=0.7},
}

\newcommand{\ourscol}{cyan}
\newcommand{\asmkcol}{OliveGreen}
\newcommand{\rrtcol}{red}
\newcommand{\globalcol}{BrickRed}

\newcommand{\rtfcol}{Orange}

\newcommand{\fullmark}{triangle*}
\newcommand{\binmark}{diamond*}
\newcommand{\pqmark}{square*}
\newcommand{\globalmark}{square*}

\pgfmathsetmacro{\tradeoffmarkersize}{3.5}
\pgfmathsetmacro{\tradeoffmarkersizeb}{2.5}
\pgfmathsetmacro{\asymmarkersize}{2}

\pgfplotsset{
    oursfull/.style={mark options={draw=black},thick, color=\ourscol, mark=\fullmark,mark size=\tradeoffmarkersize pt},
    oursbin/.style={mark options={draw=black},thick, color=\ourscol, mark=\binmark,mark size=\tradeoffmarkersize pt},
    oursreal/.style={mark options={draw=black},thick, color=\ourscol, mark=\fullmark,mark size=\tradeoffmarkersize pt},
    oursdist/.style={mark options={draw=black},thick, color=\ourscol, mark=*, mark size=\tradeoffmarkersizeb pt},
    globalfull/.style={mark options={draw=black},thick, color=\globalcol, mark=\globalmark,mark size=\tradeoffmarkersizeb pt},
    sgfull/.style={mark options={draw=black},thick, color=\ourscol, mark=\globalmark,mark size=\tradeoffmarkersizeb pt},
    dinofull/.style={mark options={draw=black},thick, color=\rtfcol, mark=\globalmark,mark size=\tradeoffmarkersizeb pt},
    globalpq/.style={mark options={draw=black},thick, color=\globalcol, mark=\pqmark,mark size=\tradeoffmarkersize pt},
    rrt/.style={mark options={draw=black},thick, color=\rrtcol, mark=\fullmark,mark size=\tradeoffmarkersize pt},
    cvnetdist/.style={mark options={draw=black},thick, color=\rtfcol, mark=*,mark size=\tradeoffmarkersizeb pt},
    cvnetreal/.style={mark options={draw=black},thick, color=\rtfcol, mark=\fullmark,mark size=\tradeoffmarkersize pt},
    asmk/.style={mark options={draw=black},thick, color=\asmkcol, mark=\binmark,mark size=\tradeoffmarkersize pt},
    rtf/.style={mark options={draw=black},thick, color=\rtfcol, mark=\fullmark,mark size=\tradeoffmarkersize pt},
    rtfdist/.style={mark options={draw=black},thick, color=\rtfcol, mark=*, mark size=\tradeoffmarkersizeb pt},
    ours-univ/.style={mark options={draw=black},thick, color=\ourscol, mark=*, mark size=\asymmarkersize pt, sharp plot},
    asymlarge/.style={mark options={draw=black},thick, color=\asmkcol, mark=*, mark size=\asymmarkersize pt, only marks, sharp plot},
    asymsmall/.style={mark options={draw=black},thick, color=\rtfcol, mark=*, mark size=\asymmarkersize pt, only marks, sharp plot},
    specific/.style={mark options={draw=black},thick, color=\globalcol, mark=*, mark size=\asymmarkersize pt, only marks, sharp plot},
    rrt-univ/.style={mark options={draw=black},thick, color=\rrtcol, mark=*, mark size=\asymmarkersize pt, only marks, sharp plot},
    ourssmall/.style={mark options={draw=black},thick, color=\rrtcol, mark=\binmark,mark size=\tradeoffmarkersize pt},
    oursbig/.style={mark options={draw=black},thick, color=\asmkcol, mark=\binmark,mark size=\tradeoffmarkersize pt},
    oursmed/.style={mark options={draw=black},thick, color=\rtfcol, mark=\binmark,mark size=\tradeoffmarkersize pt},
    oursdinoreal/.style={mark options={draw=black},thick, color=\rrtcol, mark=\fullmark,mark size=\tradeoffmarkersize pt},
    oursdinodist/.style={mark options={draw=black},thick, color=\rrtcol, mark=*, mark size=\tradeoffmarkersizeb pt},
}

%% file: tex/abbrev.tex
\def\Lqtrain{\ensuremath{L_q^{\text{\tiny \raisebox{2pt}{train}}}}\xspace}
\def\Lxtrain{\ensuremath{L_x^{\text{\tiny \raisebox{2pt}{train}}}}\xspace}
\def\Lqtest{\ensuremath{L_q^{\text{\tiny \raisebox{2pt}{test}}}}\xspace}
\def\Lxtest{\ensuremath{L_x^{\text{\tiny \raisebox{2pt}{test}}}}\xspace}
\def\Lxasmk{\ensuremath{L^{\text{\tiny \raisebox{2pt}{asmk}}}_x}\xspace}

\newcommand{\ione}{i\hspace{-.05em}+\hspace{-.07em}1}

\newcommand{\mypartight}[1]{\noindent {\bf #1}}
\newcommand{\myparagraph}[1]{\vspace{3pt}\noindent\textbf{#1}\xspace}

\newcommand{\optional}[1]{{#1}}
\newcommand{\alert}[1]{{\color{red}{#1}}}
\newcommand{\gt}[1]{{\color{purple}{GT: #1}}}
\newcommand{\gtt}[1]{{\color{purple}{#1}}}
\newcommand{\gtr}[2]{{\color{purple}\st{#1} {#2}}}

\newcommand{\gkz}[1]{{\color{cyan}{GKZ: #1}}}
\newcommand{\gkzt}[1]{{\color{cyan}{#1}}}
\newcommand{\gkzr}[2]{{\color{cyan}\st{#1} {#2}}}

\newcommand{\ps}[1]{{\color{brown}{PS: #1}}}
\newcommand{\pst}[1]{{\color{brown}{#1}}}
\newcommand{\psr}[2]{{\color{brown}\st{#1} {#2}}}

\newcommand{\ai}[1]{{\color{blue}{AI: #1}}}
\newcommand{\ait}[1]{{\color{blue}{#1}}}
\newcommand{\air}[2]{{\color{blue}\st{#1} {#2}}}

\newcommand{\gray}[1]{{\color{gray}{#1}}}

\def\roxf{$\mathcal{R}$Oxford\xspace}
\def\rox{$\mathcal{R}$Oxf\xspace}
\def\ro{$\mathcal{R}$O\xspace}
\def\rpar{$\mathcal{R}$Paris\xspace}
\def\rpa{$\mathcal{R}$Par\xspace}
\def\rp{$\mathcal{R}$P\xspace}
\def\rdis{$\mathcal{R}$1M\xspace}

\newcommand\resnet[3]{\ensuremath{\prescript{#2}{}{\mathtt{R}}{#1}_{\scriptscriptstyle #3}}\xspace}

\newcommand{\ours}{AMES\xspace} %
\newcommand{\rtf}{$R^2$Former\xspace}

\newcommand{\stddev}[1]{\scriptsize{$\pm#1$}}

\newcommand{\diffup}[1]{{\color{OliveGreen}{($\uparrow$ #1)}}}
\newcommand{\diffdown}[1]{{\color{BrickRed}{($\downarrow$ #1)}}}

\newcommand{\comment} [1]{{\color{orange} \Comment     #1}} %

\def\nmsp{\hspace{-6pt}}
\def\nssp{\hspace{-3pt}}
\def\nxssp{\hspace{-1pt}}
\def\zsp{\hspace{0pt}}
\def\xssp{\hspace{1pt}}
\def\ssp{\hspace{3pt}}
\def\msp{\hspace{6pt}}
\def\lsp{\hspace{12pt}}
\def\xlsp{\hspace{20pt}}

\newcommand{\head}[1]{{\smallskip\noindent\bf #1}}
\newcommand{\equ}[1]{(\ref{equ:#1})\xspace}

\newcommand{\nn}[1]{\ensuremath{\text{NN}_{#1}}\xspace}
\def\l1{\ensuremath{\ell_1}\xspace}
\def\l2{\ensuremath{\ell_2}\xspace}

\newcommand{\tran}{^\top}
\newcommand{\mtran}{^{-\top}}
\newcommand{\zcol}{\mathbf{0}}
\newcommand{\zrow}{\zcol\tran}

\newcommand{\ind}{\mathds{1}}
\newcommand{\expect}{\mathbb{E}}
\newcommand{\nat}{\mathbb{N}}
\newcommand{\zahl}{\mathbb{Z}}
\newcommand{\real}{\mathbb{R}}
\newcommand{\proj}{\mathbb{P}}
\newcommand{\prob}{\mathbf{Pr}}

\newcommand{\mif}{\textrm{if }}
\newcommand{\other}{\textrm{otherwise}}
\newcommand{\minimize}{\textrm{minimize }}
\newcommand{\maximize}{\textrm{maximize }}

\newcommand{\id}{\operatorname{id}}
\newcommand{\const}{\operatorname{const}}
\newcommand{\sgn}{\operatorname{sgn}}
\newcommand{\erf}{\operatorname{erf}}
\newcommand{\var}{\operatorname{Var}}
\newcommand{\mean}{\operatorname{mean}}
\newcommand{\trace}{\operatorname{tr}}
\newcommand{\diag}{\operatorname{diag}}
\newcommand{\vect}{\operatorname{vec}}
\newcommand{\cov}{\operatorname{cov}}

\newcommand{\softmax}{\operatorname{softmax}}
\newcommand{\clip}{\operatorname{clip}}

\newcommand{\defn}{\mathrel{:=}}
\newcommand{\peq}{\mathrel{+\!=}}
\newcommand{\meq}{\mathrel{-\!=}}

\newcommand{\floor}[1]{\left\lfloor{#1}\right\rfloor}
\newcommand{\ceil}[1]{\left\lceil{#1}\right\rceil}
\newcommand{\inner}[1]{\left\langle{#1}\right\rangle}
\newcommand{\norm}[1]{\left\|{#1}\right\|}
\newcommand{\frob}[1]{\norm{#1}_F}
\newcommand{\card}[1]{\left|{#1}\right|\xspace}
\newcommand{\diff}{\mathrm{d}}
\newcommand{\der}[3][]{\frac{d^{#1}#2}{d#3^{#1}}}
\newcommand{\pder}[3][]{\frac{\partial^{#1}{#2}}{\partial{#3^{#1}}}}
\newcommand{\ipder}[3][]{\partial^{#1}{#2}/\partial{#3^{#1}}}
\newcommand{\dder}[3]{\frac{\partial^2{#1}}{\partial{#2}\partial{#3}}}

\newcommand{\wb}[1]{\overline{#1}}
\newcommand{\wt}[1]{\widetilde{#1}}

\newcommand{\cA}{\mathcal{A}}
\newcommand{\cB}{\mathcal{B}}
\newcommand{\cC}{\mathcal{C}}
\newcommand{\cD}{\mathcal{D}}
\newcommand{\cE}{\mathcal{E}}
\newcommand{\cF}{\mathcal{F}}
\newcommand{\cG}{\mathcal{G}}
\newcommand{\cH}{\mathcal{H}}
\newcommand{\cI}{\mathcal{I}}
\newcommand{\cJ}{\mathcal{J}}
\newcommand{\cK}{\mathcal{K}}
\newcommand{\cL}{\mathcal{L}}
\newcommand{\cM}{\mathcal{M}}
\newcommand{\cN}{\mathcal{N}}
\newcommand{\cO}{\mathcal{O}}
\newcommand{\cP}{\mathcal{P}}
\newcommand{\cQ}{\mathcal{Q}}
\newcommand{\cR}{\mathcal{R}}
\newcommand{\cS}{\mathcal{S}}
\newcommand{\cT}{\mathcal{T}}
\newcommand{\cU}{\mathcal{U}}
\newcommand{\cV}{\mathcal{V}}
\newcommand{\cW}{\mathcal{W}}
\newcommand{\cX}{\mathcal{X}}
\newcommand{\cY}{\mathcal{Y}}
\newcommand{\cZ}{\mathcal{Z}}

\newcommand{\vA}{\mathbf{A}}
\newcommand{\vB}{\mathbf{B}}
\newcommand{\vC}{\mathbf{C}}
\newcommand{\vD}{\mathbf{D}}
\newcommand{\vE}{\mathbf{E}}
\newcommand{\vF}{\mathbf{F}}
\newcommand{\vG}{\mathbf{G}}
\newcommand{\vH}{\mathbf{H}}
\newcommand{\vI}{\mathbf{I}}
\newcommand{\vJ}{\mathbf{J}}
\newcommand{\vK}{\mathbf{K}}
\newcommand{\vL}{\mathbf{L}}
\newcommand{\vM}{\mathbf{M}}
\newcommand{\vN}{\mathbf{N}}
\newcommand{\vO}{\mathbf{O}}
\newcommand{\vP}{\mathbf{P}}
\newcommand{\vQ}{\mathbf{Q}}
\newcommand{\vR}{\mathbf{R}}
\newcommand{\vS}{\mathbf{S}}
\newcommand{\vT}{\mathbf{T}}
\newcommand{\vU}{\mathbf{U}}
\newcommand{\vV}{\mathbf{V}}
\newcommand{\vW}{\mathbf{W}}
\newcommand{\vX}{\mathbf{X}}
\newcommand{\vY}{\mathbf{Y}}
\newcommand{\vZ}{\mathbf{Z}}

\newcommand{\va}{\mathbf{a}}
\newcommand{\vb}{\mathbf{b}}
\newcommand{\vc}{\mathbf{c}}
\newcommand{\vd}{\mathbf{d}}
\newcommand{\ve}{\mathbf{e}}
\newcommand{\vf}{\mathbf{f}}
\newcommand{\vg}{\mathbf{g}}
\newcommand{\vh}{\mathbf{h}}
\newcommand{\vi}{\mathbf{i}}
\newcommand{\vj}{\mathbf{j}}
\newcommand{\vk}{\mathbf{k}}
\newcommand{\vl}{\mathbf{l}}
\newcommand{\vm}{\mathbf{m}}
\newcommand{\vn}{\mathbf{n}}
\newcommand{\vo}{\mathbf{o}}
\newcommand{\vp}{\mathbf{p}}
\newcommand{\vq}{\mathbf{q}}
\newcommand{\vr}{\mathbf{r}}
\newcommand{\Vs}{\mathbf{s}}
\newcommand{\vt}{\mathbf{t}}
\newcommand{\vu}{\mathbf{u}}
\newcommand{\vv}{\mathbf{v}}
\newcommand{\vw}{\mathbf{w}}
\newcommand{\vx}{\mathbf{x}}
\newcommand{\vy}{\mathbf{y}}
\newcommand{\vz}{\mathbf{z}}

\newcommand{\vone}{\mathbf{1}}
\newcommand{\vzero}{\mathbf{0}}

\newcommand{\valpha}{{\boldsymbol{\alpha}}}
\newcommand{\vbeta}{{\boldsymbol{\beta}}}
\newcommand{\vgamma}{{\boldsymbol{\gamma}}}
\newcommand{\vdelta}{{\boldsymbol{\delta}}}
\newcommand{\vepsilon}{{\boldsymbol{\epsilon}}}
\newcommand{\vzeta}{{\boldsymbol{\zeta}}}
\newcommand{\veta}{{\boldsymbol{\eta}}}
\newcommand{\vtheta}{{\boldsymbol{\theta}}}
\newcommand{\viota}{{\boldsymbol{\iota}}}
\newcommand{\vkappa}{{\boldsymbol{\kappa}}}
\newcommand{\vlambda}{{\boldsymbol{\lambda}}}
\newcommand{\vmu}{{\boldsymbol{\mu}}}
\newcommand{\vnu}{{\boldsymbol{\nu}}}
\newcommand{\vxi}{{\boldsymbol{\xi}}}
\newcommand{\vomikron}{{\boldsymbol{\omikron}}}
\newcommand{\vpi}{{\boldsymbol{\pi}}}
\newcommand{\vrho}{{\boldsymbol{\rho}}}
\newcommand{\vsigma}{{\boldsymbol{\sigma}}}
\newcommand{\vtau}{{\boldsymbol{\tau}}}
\newcommand{\vupsilon}{{\boldsymbol{\upsilon}}}
\newcommand{\vphi}{{\boldsymbol{\phi}}}
\newcommand{\vchi}{{\boldsymbol{\chi}}}
\newcommand{\vpsi}{{\boldsymbol{\psi}}}
\newcommand{\vomega}{{\boldsymbol{\omega}}}

\newcommand{\rLambda}{\mathrm{\Lambda}}
\newcommand{\rSigma}{\mathrm{\Sigma}}

\makeatletter
\DeclareRobustCommand\onedot{\futurelet\@let@token\@onedot}
\def\@onedot{\ifx\@let@token.\else.\null\fi\xspace}
\def\eg{\emph{e.g}\onedot} \def\Eg{\emph{E.g}\onedot}
\def\ie{\emph{i.e}\onedot} \def\Ie{\emph{I.e}\onedot}
\def\vs{\emph{vs\onedot}}
\def\cf{\emph{cf}\onedot} \def\Cf{\emph{C.f}\onedot}
\def\etc{\emph{etc}\onedot} \def\vs{\emph{vs}\onedot}
\def\wrt{w.r.t\onedot} \def\dof{d.o.f\onedot}
\def\etal{\emph{et al}\onedot}
\makeatother

\newcommand\rurl[1]{%
  \href{https://#1}{\nolinkurl{#1}}%
}

\newcommand{\bentarrow}[1][]{%
  \begin{tikzpicture}[#1]%
    \draw (0,0.7ex) -- (0,0) -- (0.75em,0);
    \draw (0.55em,0.2em) -- (0.75em,0) -- (0.55em,-0.2em);
  \end{tikzpicture}%
}

%% file: tex/abstract.tex
This work investigates the problem of instance-level image retrieval re-ranking with the constraint of memory efficiency, ultimately aiming to limit memory usage to 1KB per image. 
Departing from the prevalent focus on performance enhancements, this work prioritizes the crucial trade-off between performance and memory requirements.
The proposed model uses a transformer-based architecture designed to estimate image-to-image similarity by capturing interactions within and across images based on their local descriptors.
A distinctive property of the model is the capability for asymmetric similarity estimation. 
Database images are represented with a smaller number of descriptors compared to query images, enabling performance improvements without increasing memory consumption.
To ensure adaptability across different applications, a universal model is introduced that adjusts to a varying number of local descriptors during the testing phase. 
Results on standard benchmarks demonstrate the superiority of our approach over both hand-crafted and learned models.
In particular, compared with current state-of-the-art methods that overlook their memory footprint, our approach not only attains superior performance but does so with a significantly reduced memory footprint. The code and pretrained models are publicly available at: \url{https://github.com/pavelsuma/ames}

%% file: tex/intro.tex
\section{Introduction}
\label{sec:intro}
Image retrieval approaches typically focus on optimizing performance metrics, often effectively due to modern representations~\cite{agt+15,rtc19,gar+17,bl15}.
Higher-dimensional representations provide a better estimation of the image-to-image similarity and result in better retrieval accuracy~\cite{rms+20, mbl20}.
However, what is typically overlooked is the memory footprint requirements of storing such representations.
Storing high-dimensional, and usually dense, representations results in a large memory footprint, making them not applicable for web-scale applications.
Therefore, it becomes important to consider performance alongside the memory footprint.
This work focuses precisely on this dual aspect, which has been underestimated in the literature.
We analyze the trade-off between performance and memory usage in the realm of instance-level image retrieval.
Specifically, we investigate and improve this trade-off assuming the use of both local and global descriptors, ideally with both provided by a universal model~\cite{cas20}.

Global descriptors offer a convenient way to handle image retrieval in large databases due to computational and memory benefits. 
They provide a fast way to perform retrieval and allow additional speed-up with approximate nearest neighbor search techniques~\cite{ka14, bl12}.
At the same time, storing the representation of database images comes with a low footprint, which is further reduced by dimensionality reduction or other compression techniques~\cite{jc12,rjc15,jed+11}.
Nevertheless, despite the high performance achieved in multiple benchmarks due to deep models, global descriptors have drawbacks.
This is especially the case when it comes to severe background clutter, such as small objects and large occlusions~\cite{ita+16}. 

One way to improve performance is to represent an image with a set of local descriptors while using a similarity function to compare two local descriptor sets.
The similarity function can be hand-crafted~\cite{ras+14,taj15} or include learnable parts~\cite{tyo+21,lsl+22}.
An important aspect of such approaches is that the local descriptors are usually used at a \emph{re-ranking stage}~\cite{tyo+21,pci+07}, after a more efficient initial ranking stage, \eg retrieval with global descriptors.
In this fashion, the high computational cost of using local descriptors becomes less of a weakness if the number of images to re-rank is limited. Their large memory footprint, however, still remains a major drawback.

We explore and optimize the trade-off between retrieval performance and the memory requirements, especially in a low memory footprint regime such as 1KB per image, to allow scalable solutions. 
Our methodology targets both major factors regarding memory, namely operating with a low number of local descriptors and the footprint per descriptor.
We design an image-to-image similarity model that uses transformers to capture within and across image interactions.
The underlined trade-off is optimized in a four-fold way: (i) through asymmetric similarity estimation where the query image is represented by a higher number of local descriptors, without affecting the database footprint (Figure~\ref{fig:teaser}), (ii) training the model to operate on binary input vectors learned in an end-to-end manner, (iii) with a distillation process where the teacher model uses a richer representation than the student, 
(iv) by a proper combination of the similarity given by global and local descriptors during the re-ranking stage.

The number of local descriptors per database or query image may need to vary in a practical system to control both the memory and the retrieval speed. 
We obtain a universal model, tackling all cases, by varying their number per image during the training. 
This approach successfully alleviates the sensitivity in transformers observed during our experiments, where performance tends to decline due to discrepancies in the number of input tokens between the training and testing phases.
The list of contributions is outlined below.
\vspace{-5pt}
\begin{itemize}[leftmargin=*]
  \setlength\itemsep{0pt} %
    \item We present the first systematic study that investigates the trade-off between performance and memory consumption in the context of deep models and their representations for instance-level image retrieval.
    \item We introduce \ours, a new similarity estimation architecture and training strategy inspired by existing techniques, specifically tailored to address our unique requirements and objectives.
   \item We reveal the sensitivity of transformers to changes in the number of the input tokens and develop a universal model that exhibits robustness to such changes, ensuring stable performance across different operating conditions.
\end{itemize}

%% file: tex/related.tex
\section{Related work} 
\label{sec:related}

\textbf{Similarity using global descriptors. }
Traditional methods in image retrieval extract global representations by aggregating hand-crafted local features, \eg BoW~\cite{sz03, cdf+04}, Fisher vectors~\cite{spm+13, pls+10}, VLAD~\cite{jds+10}, or Triangulation embedding~\cite{jz14}, and compute a simple metric, \eg cosine similarity, to perform the retrieval.
The memory footprint of these representations is further improved by reducing their dimensionality with different variants of PCA~\cite{rjc15,jc12}, or product quantization (PQ)~\cite{jed+11}.
Since the advance of deep networks, utilizing learned, \ie not hand-crafted, descriptors is also a popular choice for image retrieval~\cite{agt+15, mls+17}.
Early work in the field employs neural networks trained for classification and transfers them to image retrieval by aggregating internal activation maps~\cite{bl15,rtc19}.
More recent methods directly train a neural network specifically for image retrieval by optimizing for robust global representations through different metric learning losses~\cite{rar+19, skp+15}, sampling of the training data~\cite{lxz+19, sohn16, sxj+15, wms+17}, and architectures~\cite{song2023boosting,song2024train}.
In recent work, SuperGlobal~\cite{sck+23} descriptors are extracted via a repurposed architecture of an already trained retrieval model that becomes effective with appropriate hyper-parameter tuning.
In addition to global, some models derive local descriptors too through separate network branches~\cite{yhf+21, szy+22, cas20}.

\textbf{Similarity with local descriptors. }
In contrast to aggregating local representations into a global one, specific metrics are designed to compare the local descriptors and estimate the image-to-image similarity.
Traditional approaches follow the BoW paradigm but utilize match kernels to compute the similarity between local features assigned to the same visual words~\cite{jed+08, taj15}.
Hamming Embedding~\cite{jed+08} opts for binary codes to significantly reduce the memory as it requires indexing additional features. 
The kernels are combined with weighing to reduce the burstiness effect~\cite{jed+09} or to account for local density in the embedding space~\cite{az14}.
ASMK~\cite{taj15} takes a middle route and aggregates only the local features belonging to the same visual word. This reduces the memory footprint of the stored database features, which is further reduced with binarization.
These methods are shown to be outperformed by learnable re-ranking approaches~\cite{tyo+21} or improved with appropriate representation learning~\cite{wll+22}.

Different forms of geometric constraints are shown helpful for image similarity estimation, \eg consistency in spatial neighborhoods~\cite{sz03}, angles and scales~\cite{jed+10}, or in relative spatial ordering~\cite{wki+09, hgx+21}. 
Geometric verification through RANSAC-like~\cite{fib+81} procedures is one of the most popular methods to estimate the similarity of not only hand-crafted local descriptors~\cite{pci+08, pci+07} but also local features extracted from deep networks~\cite{nas+17, cas20, sac+19}. 
This is a time-consuming re-ranking approach with an additional memory overhead due to storing local feature geometry.

Recent advancements in deep learning have led to the development of models that estimate image-to-image similarity based on local descriptors~\cite{tyo+21, zyc+23}. 
These models allow for the direct optimization of similarity computation for specific tasks, enhancing performance.
RRT~\cite{tyo+21} and $R^2$Former~\cite{zyc+23} improve over spatial verification in performance and processing time but require to store several hundred full-precision descriptors. 
CVNet~\cite{lsl+22} processes cross-scale correlation between dense feature maps with a 4D convolutional network, requiring even more memory.
The proposed model draws inspiration from the RRT architecture and effectively extends it. 
To benchmark our model, we directly compare its performance to RRT, R$^2$Former, and CVNet, demonstrating its superiority in terms of performance and memory.

Furthermore, our architecture design incorporates ideas from image matching networks~\cite{ssw+21,sdm+20,lsp+23}. 
These approaches utilize a combination of self-attention and cross-attention layers to facilitate information exchange within individual images as well as across the two images.
By integrating these concepts, our model allows for a more effective similarity estimation approach.

\textbf{Asymmetric similarity} is used in prior work to optimize the performance and memory trade-off by comparing binary to full-precision vectors~\cite{jjg11}. This approach is complementary to our contributions. Recently, a variety of deep learning methods optimizes the performance \vs query time trade-off using light-weight query processing~\cite{ba21,wwz+22,st23} or model ensembling on the database side~\cite{wwz+23}. Our universal model is able to control query time by operating on a varying number of query local descriptors.

\textbf{Transformer sensitivity to token-set size discrepancy.} 
In natural language processing (NLP), tasks such as question answering inherently involve sentences of varying lengths during training and testing~\cite{vb21}.
Models exhibit some robustness to such variations unless significantly longer sentences are encountered during testing~\cite{cwc+23,rdg+23}.
Inspired by this observation, we propose to train our models with a varying number of input tokens. 
A similar phenomenon is observed in Flexible Vision Transformers (ViTs)~\cite{bik+23} and CAPE~\cite{lxs+21}, where variations in patch or image size are considered that consequently affect the token-set size.

%% file: tex/method.tex
\section{Method for memory-efficient similarity estimation}
\label{sec:method}
We first introduce the task of memory-efficient retrieval and then present the proposed approach for training an image-to-image similarity model that operates on two sets of local descriptors as input while keeping the memory footprint low.

\subsection{Problem formulation}
The objective of image retrieval is to search a database of images $\cX$ using a given query image $q$ and retrieve relevant images.
In this work, relevance is defined in terms of depicting the same specific object, focusing on instance-level retrieval.
The image-to-image similarity between the query and an image $x \in \cX$ is measured by the function $s(x,q) \in \real$. Subsequently, the results are sorted to generate a ranking of the database images.

In contrast to the majority of prior work, which predominantly emphasizes performance metrics, we argue that the quality of a retrieval approach should be considered as a composite of both performance and memory footprint. 
The latter concerns the memory required to store the representation of image $x$ necessary for similarity estimation.
Memory allocation for the query is less crucial as it is released at the end of the retrieval process. 
Consequently, we explore and find merit in asymmetric setups within this work.

\subsection{Preliminaries}
A single-vector image representation, referred to as \emph{global descriptor}, is a memory-efficient and computationally efficient way of computing similarity.
The \l2-norma\-lized global descriptors for database image $x$ and query image $q$ are denoted by $\vx \in \real^{d_g}$ and $\vq \in \real^{d_g}$, respectively. 
Image similarity is then simplified to the dot product
$s_g(x,q) = \vx^\top \vq,$
also referred to as \emph{global similarity}.

To enhance retrieval quality, a common approach involves representing each image with multiple vectors, such as \emph{local descriptors}.
Each database image $x$ is then represented by $L_x$ $D$-dimensional vectors in $X \in \real^{D \times L_x}$, and the query $q$ by $L_q$ vectors in $Q \in \real^{D \times L_q}$. 
Although working with order-less vector sets, we employ matrix notation for the sake of clarity. 
Image-to-image similarity is defined as $    s_l(x,q) = S(X, Q), $
referred to as \emph{local similarity}, 
where $S: \real^{D \times L_x} \times \real^{D \times L_q} \rightarrow \real$ is a similarity function operating on two vector sets.
The two sets do not need to have the same size. %
While various options exist for designing and implementing $S$, the computational complexity is inevitably higher compared to the global similarity estimation.
A practical solution is to estimate global similarity in an initial ranking stage, and local similarity during re-ranking of the most similar images according to the first step. 
However, this approach requires storing both global and local representations for image $x$. 
Depending on the values of $d_g$, $D$, $L_x$, 
this may quickly become impractical for large databases.
Therefore, maintaining a low memory footprint is crucial.

\subsection{Proposed approach - \ours}
Given the local descriptor sets $X$ and $Q$ for images $x$ and $q$, respectively, we design the image-to-image similarity function $S(X,Q)$ with a transformer-based network. 
Together with the training and testing strategy, we refer to the proposed approach as ``Asymmetric and Memory-Efficient Similarity'' or \ours.

\begin{figure*}[t]
  \centering
  \includegraphics[width=0.99\textwidth]{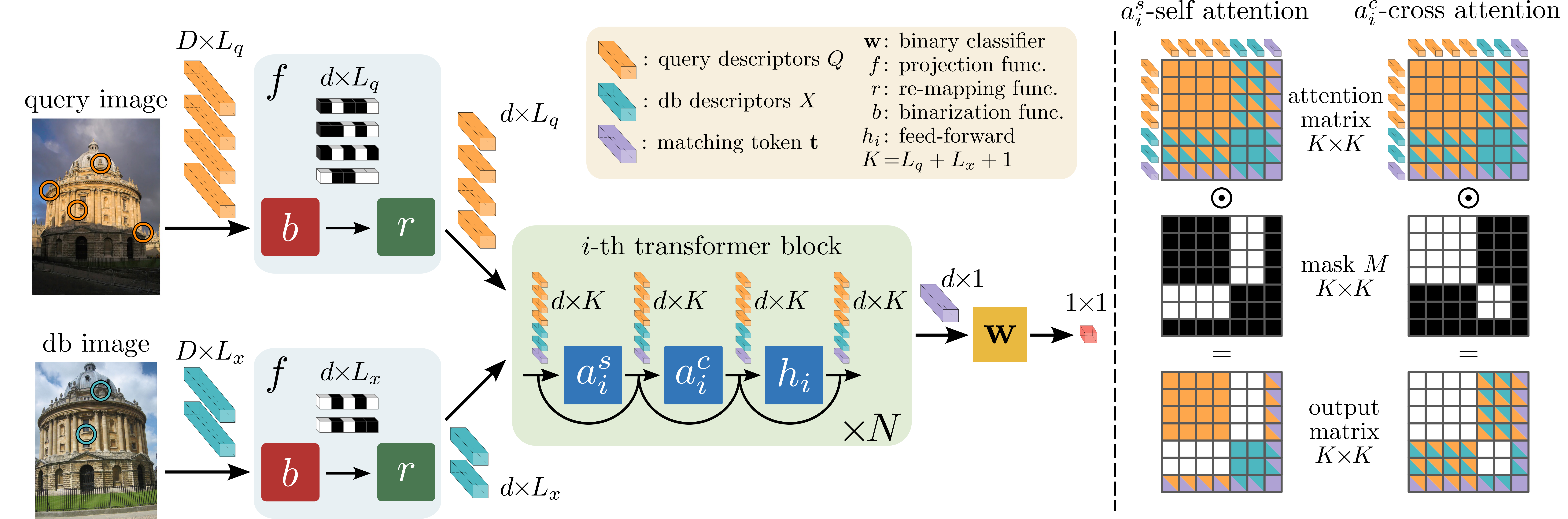}
  \vspace{-5pt}
  \caption{\textbf{Overview of the \ours model designed for estimating image-to-image similarity} when provided two local descriptor sets, where one set has a smaller size. Descriptors are processed by projection ($f$) comprised of binarization ($b$) and re-mapping to the real coordinate space ($r$). 
  During testing, binarization (re-mapping) is performed offline (online); therefore, we need to store only the binary vectors for the database images. 
  Descriptors, together with a learnable matching token, form the input token set for a transformer-based architecture, which, together with a binary classifier, estimate the similarity. 
  Sequentially, inter-image (self) and intra-image (cross) attention is performed by standard self-attention blocks alongside appropriate masking.
  \vspace{-10pt}
  \label{fig:overview}}
\end{figure*}

\textbf{\ours architecture. }
We use a transformer-based architecture~\cite{vsp+17} in a processing pipeline that interchanges between self-attention and cross-attention to capture both intra-image and inter-image interactions between local descriptors. 
As a first step, the representation space undergoes transformation, optionally involving dimensionality reduction, through a projection function $f:\real^{D}\rightarrow \real^{d}$.
This transformation is applied individually to each local descriptor, with the application to the entire set represented as $f(X)$ and $f(Q)$.

Each local descriptor becomes a transformer token whose input is composed by concatenation of the projected $X$ and $Q$, and a \emph{matching token} $\vt \in \real^{d}$. 
We include this token by following common practice, similar to the classification token~\cite{vsp+17, tyo+21}, initialize it by a random vector, and treat it as a learnable variable. 
This plays the role of information aggregation from both images through the attention mechanism.
In summary, the input to \ours consists of $K=L_x + L_q + 1$ tokens and is denoted by
\begin{equation}
    Z_0 = [~ f(X) ~~ f(Q) ~~ \vt ~].
\label{equ:input}
\end{equation}

The network consists of $N$ consecutive processing blocks that are identical in design. 
Each block $i$, $i\leq N$, comprises two consecutive attention operations and a feed-forward layer with residual connections. 
The output of the $i$-th block 
\begin{equation}
    Z_i = [~ X_i ~~ Q_i ~~ \vt_i ~],
\label{equ:inout}
\end{equation}
matches the input to $(i{+}1)$-th block,
while processing by a block is given by
\begin{eqnarray} 
    \hat{Z_{i}} =& a_{\ione}^s( Z_{i} ; M^s ) +Z_{i} \label{equ:self} \\
    \tilde{Z_{i}} =& a_{\ione}^c( \hat{Z_{i}} ; M^c )+ \hat{Z_{i}}  \label{equ:cross} \\
    Z_{\ione} =& h_{\ione}( \tilde{Z_{i}} ) +\tilde{Z_{i}} \label{equ:ff}, 
\end{eqnarray}
where \equ{self} and \equ{cross} correspond to self and cross attention, respectively. 
In the former (latter), local descriptors across images (within the same image) do not attend to each other.
Masks $M^s, M^c \in \{0, 1\}^{K \times K}$ are binary matrices that allow to perform both operations via masked attention across all input tokens\footnote{Both $a^s_i$ and $a^c_i$ perform attention across all input tokens, but it is the masking that makes them reduce down to a type of self-attention (within image) and cross-attention (across image). Super-scripts are there to indicate different learnable parameters; each block has a different set of learnable parameters.}.
Mask $M^s$ ($M^c$) has zeros for elements across the two images (within the same image) and ones otherwise.
The feed-forward process in \equ{ff} is an MLP operating independently per token.
Note that the matching token $\vt$ is never masked in order to aggregate information from both images and both operations simultaneously.
We treat the output matching token $\vt_N$, at the last block, as a representation of the image pair and pass it to a linear binary classifier 
represented by vector $\vw \in \real^{d}$. The final similarity, with the use of a sigmoid function $\sigma$, ranges in $[0,1]$ and is given by 
\begin{equation}
    S(X,Q) = \sigma(\gamma \vt_N^\top \vw ),
\label{equ:final}
\end{equation}
where $\gamma$ is a temperature variable that is set and fixed to 1 during training.
The overall architecture is shown in Figure~\ref{fig:overview}.

\textbf{Local descriptor footprint. }
The memory-demanding variant of \ours implements the projection function $f$ with a linear layer.
The output dimensionality $d$ controls the memory footprint since $f(X)$ is part of the model input and is stored for each database image. 
We refer to it as \emph{full-precision} (fp) variant.

To reduce the memory footprint, we consider an alternative projection function that 
consists of binarization function $b$ mapping local descriptors to the Hamming space and re-mapping function $r$  projecting binary descriptors back to the real coordinate space. 
The projection is given by $f = r\circ b$, with the benefit of only storing $b(X)$, which is more compact, and applying $r$ during query time to recover the real-valued vector.
This is what we refer to as \emph{binary} (bin) variant. \looseness=-1

Formally, given an arbitrary local descriptor $\vu$, the binarization function $b\colon \real^{D}\to\{-1, 1\}^{d}$ performs dimensionality reduction and vector binarization by
$    b(\vu) = \sgn\left(\vu W\right),$
where $W\in \real^{D\times d}$ is a matrix of trainable parameters, and $\sgn$ denotes the element-wise sign function, which is not differentiable. 
Therefore, we use a smooth approximation~\cite{ktp+22} given by
$b(\vu) = \erf\left(\nicefrac{\vu W}{\sqrt{2\delta^2}}\right),$
where $\erf$ is the error function, and $\delta$ a hyper-parameter that controls the smoothness of the approximation.
During training, we use the smooth variant, allowing gradients to flow, while during inference, we use sign-based binarization.
The re-mapping function $r$ consists of a linear layer and Layer Normalization~\cite{bkh16}. 

\textbf{\ours testing. }
We describe the retrieval stage with \ours and the memory requirements of different variants.

\emph{Global-local similarity ensemble: }
We assume access to both global and local descriptors but incorporate local similarity only during the re-ranking stage, where the image-to-image similarity is then expressed as a linear combination of global and local similarities
\begin{equation}
    s(x,q) = \lambda s_g(x,q) + (1-\lambda) s_l(x,q),
\label{equ:combo}
\end{equation}
referred to as \emph{ensemble similarity}.
Although most prior work~\cite{nas+17,sac+19,tyo+21} considers $\lambda=0$, \ie re-ranking solely with local similarity, the ensemble~\eqref{equ:combo} is shown to bring benefits in recent work~\cite{cas20,lsl+22}.
To effectively perform the similarity combination, we tune two hyper-parame\-ters, namely $\lambda$ and temperature $\gamma$ in \equ{final} to properly align and balance the two similarity distributions. The tuning is performed with grid search according to performance on a validation set.

\emph{Asymmetric similarity: }
We differentiate between the number of local descriptors per image used during training and testing by \Lqtrain, \Lxtrain,  and \Lqtest, \Lxtest, respectively. 
The footprint scales linearly with \Lxtest, whereas a higher number proves advantageous~\cite{tyo+21}, up to a certain extent, thus establishing the performance \vs memory trade-off.
Asymmetric similarity estimation, achieved by setting $\Lqtest > \Lxtest$, is a viable option that avoids compromising database memory and demonstrates performance improvements in our experiments.

\emph{Varying number of local descriptors: }
We treat \Lqtest as a hyper-parameter that increases the model complexity according to the underlined limits of query time, \eg per user or environment.
Despite \Lxtest being fixed for a particular database, a different system may need to use a different value.
Therefore, a universal model handling different values both for \Lxtest and \Lqtest is necessary.
The training of a universal model and its challenges are discussed in the following section.

\textbf{\ours training. }
We perform supervised training based on labels of image pairs, optionally combined with a distillation process.

\emph{Universal \ours: } 
When there is a mismatch in the local descriptor set size between the training and testing phases, we observe a performance decline.
The sensitivity is attributed to the attention blocks and is due to the change in the number of tokens.
A naive solution is to separately train a different model for each size, \ie many different models trained with $\Lqtest=\Lqtrain$ and $\Lxtest=\Lxtrain$.
We refer to these models as \emph{specific models}.
Nevertheless, this solution significantly increases the training complexity and requires maintaining multiple models for inference of different combinations of $(\Lxtest, \Lqtest)$.
Instead, we train a single model by randomly sampling values for \Lxtrain and \Lqtrain per batch and using a subset of the images' local descriptors\footnote{The order of the input images does not matter; differentiating between sizes for image $x$ and $q$ is important for the testing setup, \eg asymmetry, but not for the training.}.
This \emph{universal model} performs well when tested across different set sizes, even compared to the specific models.

\emph{Supervised training: } 
An image pair $(x,q)$ in the training set is associated with label $y_{xq}=1$ if the two images are relevant (matching) or label \mbox{$y_{xq}=0$}, otherwise.
Using the pair labels, we perform label-balanced training with backpropagation through binary cross-entropy loss, which for a specific pair is given by
$    \cL_{\text{\scriptsize bce}}(x,q) = -y_{xq} \log s_l(x,q) - (1-y_{xq}) \log (1 - s_l(x,q))$.
In our task, this corresponds to a contrastive loss pushing the similarity of matching pairs to 1 and non-matching to 0.
We optimize the parameters of functions $a_i^s, a_i^c, h_i$ for $i \in \{1,...,N\}$, along with $\vw$, $W$ and matching token $\vt$.
We find it beneficial to initialize $W$ of the binarization layer with the result of ITQ~\cite{glg+12} trained on a large set of local descriptors.

\emph{Distillation-guided supervised training: } 
We improve the (student) binary variant of \ours by using an already trained full-precision variant as a teacher. During this distillation procedure, the teacher receives a large set of local descriptors as input for both images, while the student operates with a varying set size that is smaller than that of the teacher. 
At each batch, the student descriptor set is a subset of the teacher one.
The distillation process operates in the latent representation space and not in the output space of scalar similarities, where the supervised loss is applied.
In particular, it operates right before the classifier, at the output of the last transformer block.
Let $Z^{(t)}_N$ and $Z^{(s)}_N$ be those output matrices for the teacher and the student model, respectively, for the same image pair, and $\bar{Z}^{(t)}_N$ a trimmed version of $Z^{(t)}_N$ to contain the tokens that correspond to those of the student model.
Distillation is performed via \l2 loss as
\begin{equation}
    \cL_{\text{\scriptsize dis}} = \frac{1}{d K}\left\lVert \bar{Z}^{(t)}_N - Z^{(s)}_N\right\rVert_F,
\label{equ:kd}
\end{equation}
where $\left\lVert \cdot \right\rVert_F$ is the Frobenius norm. 
In that way, the transformer of the student network is guided to generate similar output tokens as the transformer of the teacher even though it operates with lower precision local descriptors, \ie binarized, and only with a subset of the local descriptors.
To this end, we optimize a weighted sum of the two losses as follows
$  \cL = \cL_{\text{\scriptsize bce}} + \beta \cL_{\text{\scriptsize dis}},$
where $\beta$ is a hyper-parameter that tunes the impact of $\cL_{dis}$.

%% file: tex/exp.tex
\section{Experiments}
\label{sec:exp}

\subsection{Experimental setup}

\textbf{Datasets and evaluation metrics. }
We use Google Landmarks dataset v2~\cite{wac+20} (GLDv2) for training, validation, and testing. From the \textit{clean} part of the training set, we randomly select 24K classes with more than 10 images and at most 500 images per class. The final training set contains 754K images, corresponding to roughly half of the clean part. 
We use the 761K \emph{index} images as the database and the 379 \textit{public} and 750  \textit{private} query images for validation and testing, respectively. Performance is evaluated using mean Average Precision at \mbox{top-100} images (mAP@100).
$\cR$Oxford~\cite{pci+07,rit+18} and $\cR$Paris~\cite{pci+08,rit+18} datasets are used together with the accompanying 1M distractor images. Performance is evaluated with mAP.
We present results across the \textit{medium} and \textit{hard} setups of both datasets by averaging the four performance values, denoted by $\cR$OP+1M.

\textbf{Global and local descriptors. }
\ours is generic and applicable to any type of local descriptor. 
For global descriptors, we use CVNet~\cite{lsl+22} trained on GLDv2 with a ResNet101~\cite{hzr+16} backbone, its SuperGlobal (SG)~\cite{sck+23} enhancement, and pretrained DINOv2~\cite{odm+24} through its CLS token. 
For local descriptors, we extract descriptors from either the CVNet backbone or DINOv2. 
Unless mentioned otherwise, our default choices are the SG global and the CVNet local descriptors. 
We do not consider DINOv2 as the default option since $\cR$Oxford and $\cR$Paris are listed among the datasets used for its self-supervised training. 
Following prior work~\cite{cas20,tjc20,yhf+21}, we train a local feature detector in order to select the $L$ strongest local descriptors per image during testing; see the supplementary material for more details. \looseness=-1

The dimensionality of the global descriptors is equal to $d_g=2048$ or $d_g=768$ for CVNet or DINOv2, and $d=128$ for the local descriptors.
Global descriptors are either full-precision (fp)\footnote{Variant (fp) for global and local descriptors: Half-precision floating points are used.} or are compressed with product quantization (PQ)~\cite{jed+11}. 
PQ is parameterized to work with 1 byte per sub-space, and the sub-space dimension is set equal to 1, 4, or 8, denoted by PQ1, PQ4, or PQ8, respectively.
Unless otherwise stated, PQ8 is used.
The reported memory corresponds to the storage of global and local descriptors for database images.
The storage for the network parameters is not included since it is a constant and a negligible amount. \looseness=-1

\textbf{\ours training and testing. }
During mini-batch sampling for the universal model training, \Lqtrain and \Lxtrain are randomly sampled in range $[10,400]$ and are, therefore, not necessarily equal to each other. We use $N=5$ blocks in \ours for all settings, which has the same number of parameters as RRT for their default choice of 6 blocks. During testing, we re-rank the top-$m$ images obtained with global similarity. 
Local descriptor set sizes \Lxtest and \Lqtest correspond to the database images and query, respectively. 
The main variant used in our experiments, unless otherwise stated, is the universal model with binary representation after distillation, $m=1600$, $\Lqtest=600$, and varying \Lxtest to control the memory.
For each experiment, we train three networks with different seeds and report average performance. 

\textbf{Existing approaches. }
Using the same global/local descriptors, we perform a direct comparison with ASMK~\cite{taj15}\footnote{ASMK’s memory is based on the effective number of descriptors after aggregation.}, RRT~\cite{tyo+21}, and \rtf~\cite{zyc+23}.
We use their official publicly-available implementation and their default model hyper-parameters. All models are trained within our implementation framework with varying descriptor set size and the same projection function $f$ at their input for fair comparison.
The same ensemble similarity tuning strategy is used for all these methods.
We compare with the reported results for CVNet~\cite{lsl+22} with the same representation network as in our setup. We exclude it from the trade-off comparison because it performs dense similarity estimation that is unsuitable for low-memory regimes.

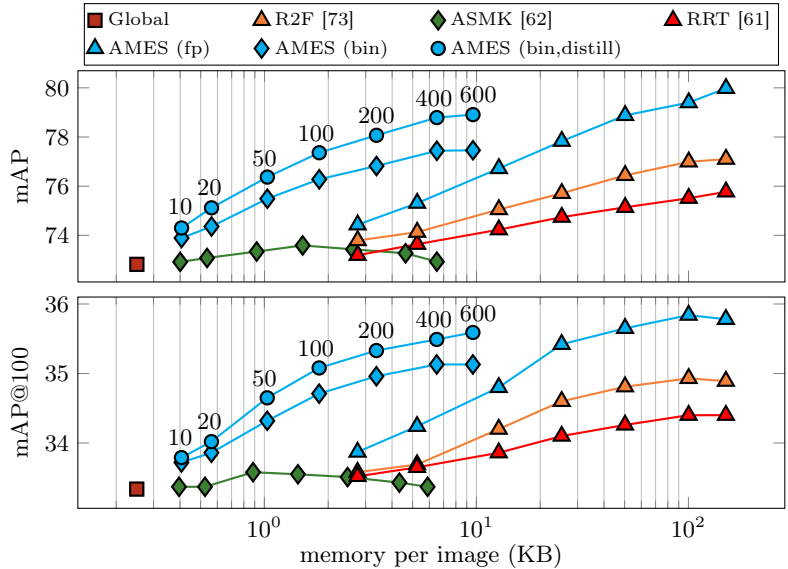
\begin{figure*}[t]
  \centering
  \input{fig/tradeoff_revop.tex}\\
  \vspace{-5pt}
  \hspace{3pt}\input{fig/tradeoff_gld.tex}
  \vspace{-10pt} %
  \caption{\textbf{Performance \vs memory trade-off} on $\cR$OP+1M (top) and GLDv2 (bottom).
  All methods use global descriptors with PQ8 for initial ranking and ensemble similarity to re-rank $m=1600$ images.
  We vary the number of local descriptors \Lxtest for database images, which is shown with text labels, indicatively, for one variant. 
  Binary/full-precision local descriptors denoted by bin/fp.   
  All methods are trained and tested by us, within the same implementation framework.
  \label{fig:tradeoff_revop_gld}
  \vspace{-15pt} %
  }
\end{figure*}

\begin{table}[!b]
  \centering
  \vspace{-30pt}
  \input{tab/sota}
  \caption{\textbf{Comparison with the state-of-the-art.} 
  Backbone architectures are ResNet50 for DELG, ResNet101 for CVNet/SG, and ViT-B/14 for DINOv2.
  Global descriptors are in full precision.
  Scores for SG without and with re-ranking are with our best possible reproduction using the publicly available implementation. 
  $\heartsuit$: reported in the literature, $\spadesuit$: evaluated by us.
  \ours is used with $\Lxtest=600$ and $\Lqtest=600$ local descriptors for the database and query image, respectively.
  \label{tab:sota}
  \vspace{-15pt}
  }
\end{table}

\subsection{Results}

\textbf{Performance/memory comparison.} Figure~\ref{fig:tradeoff_revop_gld} shows the performance of the compared approaches on different memory settings.
\ours achieves the best trade-off; it performs better than all other methods for the same memory or performs the same for much less memory. 
The binary variant significantly decreases memory over the full-precision variant at the cost of a small performance drop. On top of that, our distillation consistently recovers most of this performance loss.
ASMK provides small improvements and does not benefit from more descriptors.
We outperform \rtf and RRT in all cases.

\begin{figure}[!t]
\vspace{0pt}
  \centering
  \input{fig/asymmetry_revop}
  \vspace{-12pt} %
  \caption{\textbf{Universal \vs specific \ours models.} 
  Single specific: one model trained with a fixed number of local descriptors \Lxtrain (value in brackets) and tested with varying $\Lxtest$. 
  Many specific: six models in total, trained and tested per number of local descriptors ($\Lxtest=\Lxtrain$).
  All models are without distillation.
  \label{fig:asymmetry_revop}
  \vspace{-10pt} %
  }
\end{figure}
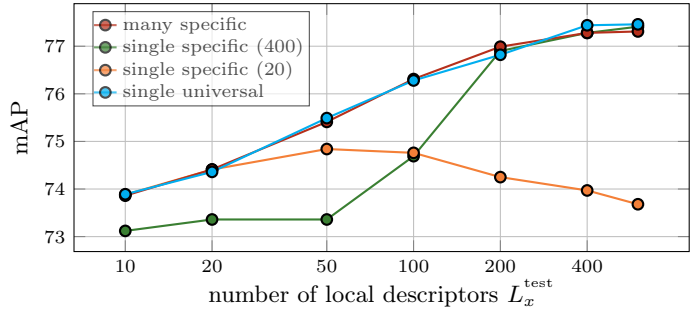

\textbf{State-of-the-art (SoA) comparison.} Reported results of SoA methods are shown in Table~\ref{tab:sota}.
\ours achieves SoA performance when combined with either CVNet or SuperGlobal as global descriptors and using the same value $m$ as the competing approaches. 
In the direct comparison with CVNet re-ranking, we outperform it with a clear margin while our per-image memory footprint is significantly lower, \ie 301KB and $\sim$1400KB for \ours (fp) and CVNet, respectively. Note that CVNet requires the full dense feature map representation. Our binary variant achieves very competitive performance requiring only $\sim$10KB, further optimizing performance-memory trade-off. 
In addition, \ours improves performance over the query expansion re-ranking approach of SuperGlobal~\cite{sck+23} (\ours applied after SG re-rank), highlighting that SG re-rank is complementary in nature to \ours. 
In contrast, CVNet re-ranking does not improve over SG re-ranking (cf. Shao \etal~\cite{sck+23}).

\textbf{Universal \vs specific models. }
In Figure~\ref{fig:asymmetry_revop}, we demonstrate performance for varying number of local descriptors per database image $\Lxtest$ used during testing.
Specific models are trained with fixed $\Lxtrain$ and either tested for $\Lxtest=\Lxtrain$ (many specific) or tested for varying $\Lxtest$ (single specific). 
Our universal model achieves about the same performance as each of the many specific models, even though those are tested for the number of local descriptors they are trained with.
Single specific variants demonstrate notable sensitivity to the discrepancy between \Lxtest and \Lxtrain. \looseness=-1

\begin{table}[t]
  \centering
  \input{tab/ablation}
  \caption{\textbf{Impact of \ours components} on performance and memory/image (KB). 
  \label{tab:ablation}
  \vspace{-15pt} %
  }
\end{table}

\textbf{Ablation study. }
Table~\ref{tab:ablation} reports performance and memory for an ablation study.
Asymmetric similarity is beneficial, while binarization slightly compromises the performance but provides large benefits in memory requirements. Distillation improves performance without compromising memory. Compressing the global descriptor with PQ8 saves memory with negligible performance loss.

\begin{figure}[!t]
\begin{minipage}[c]{0.49\textwidth}
  \centering
  \input{tab/asymmetry}
  \vspace{0pt} %
  \captionof{table}{
  \textbf{Impact of asymmetry} on performance.
  Experiment with varying number of descriptors \Lqtest (query) and \Lxtest (database) during testing. 
  Global similarity performance is shown for reference.
  \label{tab:asymmetry}
  \vspace{0pt}
  }
\end{minipage}%
\hspace{3pt}
\begin{minipage}[!t]{0.49\textwidth}
\centering
  \input{tab/combos}
  \vspace{0pt}
  \captionof{table}{
  \textbf{Performance of \ours for small memory footprint per image} using various global-local descriptor combinations. 
  Global descriptor quantization and the number of local descriptors vary. %
  \label{tab:combos}
  \vspace{0pt}
  }
\end{minipage}%
\end{figure}

\textbf{Impact of asymmetry.} 
Table~\ref{tab:asymmetry} reports the results for symmetric ($\Lxtest\hspace{-3pt}=\hspace{-3pt}\Lqtest$) and asymmetric ($\Lxtest\hspace{-3pt}<\hspace{-2pt}\Lqtest$) similarity.
The number of descriptors varies only on the query side and remains fixed on the database side.
Increasing $L_q$ brings a significant performance increase in all cases.  
Figure~\ref{fig:teaser} shows the importance of local descriptors based on the dot product similarity of the respective tokens and the matching token at the output of the $N$-th transformer block. 
In the asymmetric setting, only a few descriptors that correspond to the query landmark are present, to which our model correctly assigns high importance.
\looseness=-1

\textbf{Small memory footprints. }
Table~\ref{tab:combos} shows results for different combinations of local/global descriptors to achieve a target memory per image. 
Observe that it is worth compromising the precision of the global descriptor as an exchange for a larger number of additional local descriptors (PQ1/64 is worse than PQ8/176 at 3KB, and PQ4/32 is worse than PQ8/48 at 1KB).

\begin{figure}[t]
  \centering
  \vspace{10pt}
  \hspace{-10pt}
  \input{fig/tradeoff_dino_cvnet_revop.tex}
  \hspace{-115pt}
  \input{fig/tradeoff_dino_cvnet_gld.tex}
  \vspace{-10pt}
  \caption{\textbf{\ours with different backbones for global and local representation} on $\cR$OP+1M (left) and GLDv2 (right).
  \vspace{-10pt}
  \label{fig:tradeoff_dino}
  }
\end{figure}
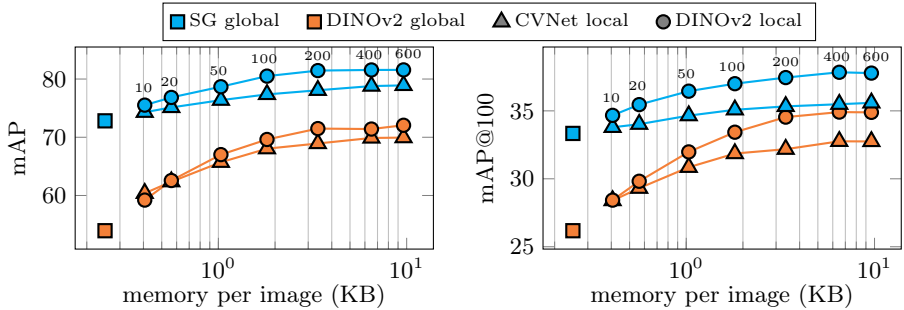

\textbf{CVNet \vs DINOv2 for global and local descriptors.}
In Figure~\ref{fig:tradeoff_dino}, we show performance for four different combinations of global and local descriptors. 
CVNet, and consequently SG, are the outcome of task-specific training, while DINOv2 is a foundational model.
Observe how SG is noticeably better as a global descriptor, while DINOv2 as a local descriptor.
Their combination, \ie SG global with DINOv2 local, is the winning entry of Table~\ref{tab:sota}.
\looseness=-1

\begin{figure}[t]
  \centering
    \input{fig/timings_revop.tex}
    \vspace{-10pt}
    \caption{\textbf{Performance \vs time} on $\cR$OP+1M for varying \Lqtest (shown in text labels) and two values of \Lxtest (\rtf and RRT use $\Lxtest=50$) using batches of 70 pairs. Global similarity takes 0.03 ms. 
    All three models use standard PyTorch modules. 
    \vspace{-10pt}
    }
    \label{fig:timings}    
\end{figure}
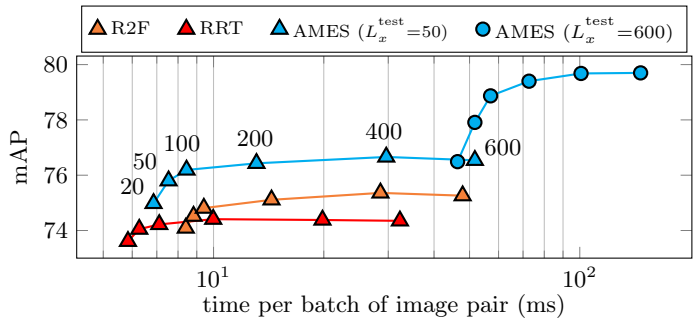

\textbf{Inference speed and GPU memory complexity.}
Inference time is reported in Fig~\ref{fig:timings}, showing how \Lqtest and \Lxtest impact latency and that \ours provides a better trade-off.
Regarding GPU memory, the maximum batch size during testing on an 80GB GPU with $\Lqtest=600$ and $\Lxtest=50$ is 6.1k, 4.3k, and 5.2k pairs for RRT, \rtf, and \ours, respectively.

%% file: fig/tradeoff_revop.tex
\pgfmathsetmacro{\teasermarkersize}{2.5}
\begin{tikzpicture}
\begin{axis}[%
  width=0.9\linewidth,
  height=0.36\linewidth,
  ylabel={\footnotesize mAP},
  tick label style={font=\footnotesize},
  ylabel near ticks, xlabel near ticks, 
  xticklabels={},
  legend columns=4,
  legend style={
            at={(0.5,1.32)},
            anchor=north,
            inner sep=1pt,
            row sep = 0pt,
            only marks,
            font=\scriptsize,
            /tikz/every even column/.append style={column sep=0.5cm}
            },  
  grid=minor,
  xlabel style={yshift=1ex},
  xmode=log,
  ]
    \addlegendimage{globalfull} \addlegendentry{Global};
    \addlegendimage{rtf} \addlegendentry{R2F~\cite{zyc+23}};
    \addlegendimage{asmk} \addlegendentry{ASMK~\cite{taj15}};
    \addlegendimage{rrt} \addlegendentry{RRT~\cite{tyo+21}};
    \addlegendimage{oursfull} \addlegendentry{\ours (fp)};
    \addlegendimage{oursbin} \addlegendentry{\ours (bin)};
    \addlegendimage{oursdist} \addlegendentry{\ours (bin,distill)};
    
    \addplot[globalfull] coordinates {(0.25, 72.82)};

    \addplot[oursbin] coordinates {(0.40625,73.89)   %
                                   (0.5625,74.36)    %
                                   (1.03125,75.49)   %
                                   (1.8125,76.28)    %
                                   (3.375,76.82)     %
                                   (6.5,77.44)       %
                                   (9.625,77.46)};   %

    \addplot[oursdist] coordinates {(0.40625,74.3)   %
                                   (0.5625,75.12)    %
                                   (1.03125,76.37)   %
                                   (1.8125,77.36)    %
                                   (3.375,78.07)     %
                                   (6.5,78.79)       %
                                   (9.625,78.91)};   %
                            
    \addplot[asmk] coordinates {(0.40029296875,72.92) %
                                (0.53599609375,73.08) %
                                (0.918671875,73.34)   %
                                (1.513515625,73.59)   %
                                (2.61619140625,73.43) %
                                (4.62958984375,73.27) %
                                (6.4991836,72.93)};   %
              
    \addplot[rrt] coordinates {(2.75,73.19)      %
                               (5.25,73.64)      %
                               (12.75,74.23)     %
                               (25.25,74.74)     %
                               (50.25,75.14)     %
                               (100.25,75.51)     %
                               (150.25,75.77)};   %

    \addplot[oursreal] coordinates {(2.75,74.43)      %
                                    (5.25,75.31)      %
                                    (12.75,76.72)     %
                                    (25.25,77.83)     %
                                    (50.25,78.88)     %
                                    (100.25,79.4)     %
                                    (150.25,79.98)};   %

    \addplot[rtf] coordinates      {(2.75,73.79)      %
                                    (5.25,74.13)      %
                                    (12.75,75.05)     %
                                    (25.25,75.71)      %
                                    (50.25,76.44)     %
                                    (100.25,76.99)    %
                                    (150.25,77.1)};  %

    \node [above] at (axis cs:  0.4, 74.5) {\footnotesize \textcolor{black}{10}};
    \node [above] at (axis cs:  0.55,75.3) {\footnotesize \textcolor{black}{20}};
    \node [above] at (axis cs:  1.0,76.45) {\footnotesize \textcolor{black}{50}};
    \node [above] at (axis cs:  1.75,77.5) {\footnotesize \textcolor{black}{100}};
    \node [above] at (axis cs:  3.37,78.2) {\footnotesize \textcolor{black}{200}};
    \node [above] at (axis cs:  6.3,78.9) {\footnotesize \textcolor{black}{400}};
    \node [above] at (axis cs:  10.2,79.07) {\footnotesize \textcolor{black}{600}};

\end{axis}
\end{tikzpicture}

%% file: fig/tradeoff_gld.tex
\pgfmathsetmacro{\teasermarkersize}{2.5}
\begin{tikzpicture}
\begin{axis}[%
  width=0.9\linewidth,
  height=0.36\linewidth,
  ylabel={\footnotesize mAP@100},
  xlabel={\footnotesize memory per image (KB)},
  tick label style={font=\footnotesize},
  ymax=36.1,
  ylabel near ticks, xlabel near ticks, 
  ytick={33,34,35,36},
  grid=minor,
  xlabel style={yshift=1ex},
  xmode=log,
  ]

    \addplot[globalfull] coordinates {(0.25, 33.337)};
    
    \addplot[asmk] coordinates {(0.395546875,33.37)   %
                                (0.52474609375,33.37) %
                                (0.8845703125,33.58)  %
                                (1.4396875,33.55)     %
                                (2.46853515625,33.51) %
                                (4.33216796875,33.43) %
                                (5.885722656,33.37)}; %

    \addplot[oursbin] coordinates {(0.40625,33.72)   %
                                   (0.5625,33.86)    %
                                   (1.03125,34.32)   %
                                   (1.8125,34.71)    %
                                   (3.375,34.96)     %
                                   (6.5,35.13)       %
                                   (9.625,35.13)};   %

    \addplot[oursdist] coordinates {(0.40625,33.79)  %
                                   (0.5625,34.02)    %
                                   (1.03125,34.65)   %
                                   (1.8125,35.08)    %
                                   (3.375,35.33)     %
                                   (6.5,35.49)       %
                                   (9.625,35.59)};   %

    \addplot[rtf] coordinates {(2.75,33.58)     %
                               (5.25,33.69)     %
                               (12.75,34.2)    %
                               (25.25,34.6)    %
                               (50.25,34.81)     %
                               (100.25,34.93)    %
                               (150.25,34.89)};  %

    \addplot[rrt] coordinates {(2.75,33.52)     %
                               (5.25,33.65)     %
                               (12.75,33.86)    %
                               (25.25,34.1)    %
                               (50.25,34.26)     %
                               (100.25,34.4)    %
                               (150.25,34.4)};  %

    \addplot[oursreal] coordinates {(2.75,33.87)   %
                               (5.25,34.24)     %
                               (12.75,34.8)   %
                               (25.25,35.42)   %
                               (50.25,35.65)   %
                               (100.25,35.84)  %
                               (150.25,35.78)}; %
                      
    \node [above] at (axis cs:  0.4,33.85) {\footnotesize \textcolor{black}{10}};
    \node [above] at (axis cs:  0.55,34.07) {\footnotesize \textcolor{black}{20}};
    \node [above] at (axis cs:  1.0,34.7) {\footnotesize \textcolor{black}{50}};
    \node [above] at (axis cs:  1.75,35.13) {\footnotesize \textcolor{black}{100}};
    \node [above] at (axis cs:  3.37,35.39) {\footnotesize \textcolor{black}{200}};
    \node [above] at (axis cs:  6.3,35.55) {\footnotesize \textcolor{black}{400}};
    \node [above] at (axis cs:  10.2,35.63) {\footnotesize \textcolor{black}{600}};

\end{axis}
\end{tikzpicture}

%% file: tab/sota.tex
\scalebox{0.75}{
\begin{tabular}{@{\xssp}l @{\lsp}l@{\lsp} l @{\lsp} c @{\ssp}c@{\ssp}cc @{\xssp}}
    \toprule
    \multirow{2}{*}{\textbf{global desc.}} & \multirow{2}{*}{\textbf{local desc.}} & \multicolumn{2}{c}{\textbf{re-ranking}} & \multirow{2}{*}{} & \multirow{2}{*}{\textbf{$\mathcal{R}$OP+1M}} & \multirow{2}{*}{\textbf{GLDv2}} \\ \cmidrule(lr){3-4}
    & & \textbf{approach} & \textbf{top}-$m$ & &
    \\ \toprule
        \multirow{3}{*}{{ \textbf{DELG}~\cite{cas20}}}  & {n/a} & {n/a} & {n/a} & $\heartsuit$& {51.6} & {24.1}  \\ 
        & {\textbf{GV}~\cite{cas20}}       & {\textbf{DELG}~\cite{cas20}}  & {100} & $\heartsuit$& {56.5} & {24.3}  \\
        & {\textbf{RRT}~\cite{tyo+21}}     & {\textbf{DELG}~\cite{cas20}} & {100} & $\heartsuit$& {57.6} & {--}  \\  \midrule
        \multirow{7}{*}{\textbf{CVNet}~\cite{lsl+22}}
        & n/a & n/a & n/a &$\heartsuit$& 67.6 & 32.5  \\ 
        & \multirow{6}{*}{\textbf{CVNet}~\cite{lsl+22}} 
        & \textbf{CVNet re-rank}~\cite{lsl+22}    & 100 &$\heartsuit$& 73.0 & 34.9 \\
        & & \textbf{AMES} (fp)                    & 100 &$\spadesuit$& 72.2 & 35.3 \\ 
        & & \textbf{AMES} (bin,distill)           & 100 &$\spadesuit$& 72.0 & 35.1 \\ 
        & & \textbf{CVNet re-rank}~\cite{lsl+22}  & 400 &$\heartsuit$& 75.6 & --   \\ 
        & & \textbf{AMES} (fp)                    & 400 &$\spadesuit$& \textbf{75.8} & \textbf{35.5} \\ 
        & & \textbf{AMES} (bin,distill)           & 400 &$\spadesuit$& 75.4 & 35.1 \\  
        \midrule
        \multirow{10}{*}{\textbf{SG}~\cite{sck+23}}\hspace{2pt} & n/a & n/a & n/a &$\spadesuit$& 72.9 & 33.3  \\
        & \multirow{2}{*}{\textbf{CVNet}~\cite{lsl+22}} 
        & \textbf{AMES} (fp) & 1600 &$\spadesuit$& 80.0 & 35.8  \\
        & & \textbf{AMES} (bin,distill)                 & 1600 &$\spadesuit$& 78.9 & 35.6  \\ 
        & \multirow{2}{*}{\textbf{DINOv2}~\cite{odm+24}}  
        & \textbf{AMES} (fp)                            & 1600 &$\spadesuit$& 83.5 & 38.3  \\
        & & \textbf{AMES} (bin,distill)                 & 1600 &$\spadesuit$& 81.8 & 37.8  \\  \cmidrule{2-6}
        & n/a & \textbf{SG re-rank}~\cite{sck+23} & 1600 &$\spadesuit$& 80.5 & 34.1 \\
        & \multirow{2}{*}{\textbf{CVNet}~\cite{lsl+22}} 
        & \textbf{SG re-rank}~\cite{sck+23}~+~\textbf{AMES} (fp)                  & 1600 &$\spadesuit$& 82.3 & 36.0 \\ 
        & & \textbf{SG re-rank}~\cite{sck+23}~+~\textbf{AMES} (bin,distill)       & 1600 &$\spadesuit$& 81.9 & 35.8 \\ 
        & \multirow{2}{*}{\textbf{DINOv2}~\cite{odm+24}} 
        & \textbf{SG re-rank}~\cite{sck+23}~+~\textbf{AMES} (fp)                  & 1600 &$\spadesuit$& \textbf{84.5} & \textbf{38.5} \\ 
        & & \textbf{SG re-rank}~\cite{sck+23}~+~\textbf{AMES} (bin,distill)       & 1600 &$\spadesuit$& 83.3 & 38.0  \\
        \bottomrule
\end{tabular}
}

%% file: fig/asymmetry_revop.tex
\pgfplotsset{
compat=1.11,
legend image code/.code={
\draw[mark repeat=2,mark phase=2]
plot coordinates {
(0cm,0cm)
(0.15cm,0cm)        %
(0.3cm,0cm)         %
};%
}
}
\pgfmathsetmacro{\teasermarkersize}{2.5}
\begin{tikzpicture}
\begin{axis}[%
  width=.8\linewidth,
  height=0.4\linewidth,
  ylabel={\small mAP},
  xlabel={\small number of local descriptors  \Lxtest },
  tick label style={font=\scriptsize},
  ylabel near ticks, xlabel near ticks, 
  legend pos=north west,
  legend style={opacity = .7,
            inner sep=0pt,
            cells={anchor=west},
            only marks,
            legend entries={only marks,{only marks, sharp plot}},
            font=\scriptsize,
            row sep = -2pt,
            },
  log ticks with fixed point,
  xlabel style={yshift=1ex},
  xtick={10,20,50,100,200,400},
  xmode=log,
  ]
    \addlegendimage{specific} \addlegendentry{many specific};
    \addlegendimage{asymlarge} \addlegendentry{single specific (400)};
    \addlegendimage{asymsmall} \addlegendentry{single specific (20)};
    \addlegendimage{ours-univ} \addlegendentry{single universal};
    
   \addplot[asymlarge] coordinates {(10,73.12)
                                    (20,73.36)
                                    (50,73.36)
                                    (100,74.69)
                                    (200,76.9)
                                    (400,77.28)
                                    (600,77.41)};

   \addplot[asymsmall] coordinates {(10,73.88)
                                    (20,74.41)
                                    (50,74.84)
                                    (100,74.76)
                                    (200,74.25)
                                    (400,73.97)
                                    (600,73.68)};

   \addplot[specific] coordinates  {(10,73.86)
                                    (20,74.41)
                                    (50,75.41)
                                    (100,76.31)
                                    (200,76.99)
                                    (400,77.28)
                                    (600,77.31)};

   \addplot[ours-univ] coordinates {(10,73.89)
                                    (20,74.36)
                                    (50,75.49)
                                    (100,76.28)
                                    (200,76.82)
                                    (400,77.44)
                                    (600,77.46)};

    \end{axis}
\end{tikzpicture}

%% file: tab/ablation.tex
\renewcommand{\arraystretch}{.83}
\setlength\tabcolsep{3.5pt}
\scalebox{0.87}{
\begin{tabular}{l l l c c l c c r}
    \toprule
    architecture & \Lxtest & \Lqtest & bin & dis & global & \textbf{$\mathcal{R}$OP+1M} & \textbf{GLDv2} & \textbf{mem.}
    \\ \midrule
    \textbf{AMES}   & 50 & 50  & -- & -- & full & 75.9 & 34.4 & 20.5  \\
    \textbf{AMES}   & 50 & 600 & -- & -- & full & 76.7 & 34.8 & 20.5  \\
    \textbf{AMES}   & 50 & 600 & \checkmark & -- & full & 75.6 & 34.3 & 8.8 \\
    \textbf{AMES}   & 50 & 600 & \checkmark & \checkmark & full & 76.5 & 34.6 & 8.8 \\
    \textbf{AMES}   & 50 & 600 & \checkmark & \checkmark & PQ8  & 76.4 & 34.7 & 1.0 \\ \bottomrule
\end{tabular}
}

%% file: tab/asymmetry.tex
\setlength\tabcolsep{5pt}
\renewcommand{\arraystretch}{.85}
\scalebox{0.90}{
\begin{tabular}{l l  c c}
    \toprule
    \Lxtest & \Lqtest & \textbf{$\mathcal{R}$OP+1M} & \textbf{GLDv2} \\ \midrule
    \multirow{2}{*}{50} 
     &  600  & 75.5 & 34.3 \\
     &  50   & 74.8 & 34.0 \\ \midrule
    \multirow{2}{*}{20} 
     &  600  & 74.4 & 33.9 \\
     &  20   & 73.6 & 33.5 \\ \midrule
    \multirow{2}{*}{10} 
     &  600  & 73.9 & 33.7 \\
     &  10   & 73.2 & 33.4 \\ \midrule
    \multicolumn{2}{c}{\color{gray} global-PQ8} & {\color{gray} 72.8} & {\color{gray} 33.3} \\
    \bottomrule
\end{tabular}
}

%% file: tab/combos.tex
\renewcommand{\arraystretch}{.83}
\scalebox{0.88}{
\begin{tabular}{l l l  c c c c c c c}
    \toprule
    \textbf{mem.} & global & \Lxtest & \textbf{$\mathcal{R}$OP+1M} & \textbf{GLDv2} \\ \midrule
    \multirow{3}{*}{3KB}     & PQ1  & 64 & 76.9 & 34.8 \\
                             & PQ4  & 160 & 77.8 & 35.2 \\ 
                             & PQ8  & 176 & 78.0 & 35.3 \\ \midrule
    \multirow{3}{*}{2KB}     & PQ1  & 0   & 72.8 & 33.3 \\ 
                             & PQ4  & 96  & 77.4 & 35.1 \\
                             & PQ8  & 112 & 77.4 & 35.2 \\ \midrule
    \multirow{2}{*}{1KB}     & PQ4  & 32  & 75.9 & 34.3 \\
                             & PQ8  & 48  & 76.4 & 34.6 \\ \bottomrule
\end{tabular}
}

%% file: fig/tradeoff_dino_cvnet_revop.tex
\pgfmathsetmacro{\teasermarkersize}{2.5}
\begin{tikzpicture}
\begin{axis}[%
  width=0.52\linewidth,
  height=0.35\linewidth,
  ylabel={\small mAP},
  xlabel={\small memory per image (KB)},
  tick label style={font=\small},
  ylabel near ticks, xlabel near ticks, 
  grid=minor,
  xlabel style={yshift=1ex},
  xmode=log,
  ymax=85.5
  ]
    
    \addplot[sgfull] coordinates {(0.25, 72.82)};
    \addplot[dinofull] coordinates {(0.25, 53.95)};

    \addplot[rtf] coordinates     {(0.40625,60.38)   %
                                   (0.5625 ,62.37)    %
                                   (1.03125,65.71)   %
                                   (1.8125 ,68.04)    %
                                   (3.375  ,68.93)     %
                                   (6.5    ,69.88)       %
                                   (9.625  ,69.94)};   %

    \addplot[rtfdist] coordinates {(0.40625,59.21)   %
                                    (0.5625 ,62.57)    %
                                    (1.03125,67.02)   %
                                    (1.8125 ,69.62)    %
                                    (3.375  ,71.49)     %
                                    (6.5    ,71.41)       %
                                    (9.625  ,72.06)};   %

    \addplot[oursreal] coordinates {(0.40625,74.3)   %
                                   (0.5625 ,75.12)    %
                                   (1.03125,76.37)   %
                                   (1.8125 ,77.36)    %
                                   (3.375  ,78.07)     %
                                   (6.5    ,78.79)       %
                                   (9.625  ,78.91)};   %

    \addplot[oursdist] coordinates {(0.40625,75.52)   %
                                    (0.5625 ,76.84)    %
                                    (1.03125,78.68)   %
                                    (1.8125 ,80.5)    %
                                    (3.375  ,81.45)     %
                                    (6.5    ,81.56)       %
                                    (9.625  ,81.57)};   %

    \node [above] at (axis cs:  0.4,76.25) {\tiny \textcolor{black}{10}};
    \node [above] at (axis cs:  0.55,77.35) {\tiny \textcolor{black}{20}};
    \node [above] at (axis cs:  1.0,79.2) {\tiny \textcolor{black}{50}};
    \node [above] at (axis cs:  1.75,81.33) {\tiny \textcolor{black}{100}};
    \node [above] at (axis cs:  3.37,81.79) {\tiny \textcolor{black}{200}};
    \node [above] at (axis cs:  6.3,82.0) {\tiny \textcolor{black}{400}};
    \node [above] at (axis cs:  10.2,82.10) {\tiny \textcolor{black}{600}};
    
\end{axis}
\end{tikzpicture}

%% file: fig/tradeoff_dino_cvnet_gld.tex
\pgfmathsetmacro{\teasermarkersize}{2.5}
\begin{tikzpicture}
\begin{axis}[%
  width=0.52\linewidth,
  height=0.35\linewidth,
  ylabel={\small mAP@100},
  xlabel={\small memory per image (KB)},
  tick label style={font=\small},
  ylabel near ticks, xlabel near ticks, 
  legend columns=4,
  legend style={
            at={(-0.15,1.22)},
            anchor=north,
            only marks,
            /tikz/every even column/.append style={column sep=0.3cm}
            },  
  grid=minor,
  xlabel style={yshift=1ex},
  xmode=log,
  ymax=39.7
  ]

    \addlegendimage{sgfull} \addlegendentry{\scriptsize SG global};
    \addlegendimage{dinofull} \addlegendentry{\scriptsize DINOv2 global};
    \addlegendimage{mark=triangle*, mark options={solid, fill=gray}, mark size=3.5 pt} \addlegendentry{\scriptsize CVNet local};
    \addlegendimage{mark=*, mark options={solid, fill=gray}, mark size=\tradeoffmarkersizeb pt} \addlegendentry{\scriptsize DINOv2 local};

    \addplot[sgfull] coordinates {(0.25, 33.34)};
    \addplot[dinofull] coordinates {(0.25, 26.18)};                                 

    \addplot[rtf] coordinates {(0.40625,28.4)   %
                                   (0.5625 ,29.31)    %
                                   (1.03125,30.85)   %
                                   (1.8125 ,31.86)    %
                                   (3.375  ,32.18)     %
                                   (6.5    ,32.76)       %
                                   (9.625  ,32.75)};   %
    
    \addplot[rtfdist] coordinates {(0.40625 ,28.43)   %
                                    (0.5625  ,29.82)    %
                                    (1.03125 ,31.98)   %
                                    (1.8125  ,33.43)    %
                                    (3.375   ,34.54)     %
                                    (6.5     ,34.91)       %
                                    (9.625   ,34.89)};   %

    \addplot[oursreal] coordinates {(0.40625,33.79)   %
                                   (0.5625 ,34.02)    %
                                   (1.03125,34.65)   %
                                   (1.8125 ,35.08)    %
                                   (3.375  ,35.33)     %
                                   (6.5    ,35.49)       %
                                   (9.625  ,35.59)};   %

    \addplot[oursdist] coordinates {(0.40625,34.67)   %
                                    (0.5625 ,35.46)    %
                                    (1.03125,36.45)   %
                                    (1.8125 ,37.00)    %
                                    (3.375  ,37.45)     %
                                    (6.5    ,37.84)       %
                                    (9.625  ,37.78)};   %

    \node [above] at (axis cs:  0.4,35.05) {\tiny \textcolor{black}{10}};
    \node [above] at (axis cs:  0.55,35.87) {\tiny \textcolor{black}{20}};
    \node [above] at (axis cs:  1.0,36.75) {\tiny \textcolor{black}{50}};
    \node [above] at (axis cs:  1.75,37.30) {\tiny \textcolor{black}{100}};
    \node [above] at (axis cs:  3.37,37.69) {\tiny \textcolor{black}{200}};
    \node [above] at (axis cs:  6.3,38.02) {\tiny \textcolor{black}{400}};
    \node [above] at (axis cs:  10.2,38.00) {\tiny \textcolor{black}{600}};
    
\end{axis}
\end{tikzpicture}

%% file: fig/timings_revop.tex
\pgfmathsetmacro{\teasermarkersize}{2.5}
\begin{tikzpicture}
\begin{axis}[%
  width=0.8\linewidth,
  height=0.35\linewidth,
  ylabel={\footnotesize mAP},
  xlabel={\footnotesize time per batch of image pair (ms)},
  tick label style={font=\footnotesize},
  ylabel near ticks, xlabel near ticks, 
  legend columns=4,
  legend style={
            at={(0.5,1.25)},
            anchor=north,
            only marks,
            /tikz/every even column/.append style={column sep=0.3cm}
            },  
  grid=minor,
  xlabel style={yshift=1.0ex},
  xmode=log,
  ]
    \addlegendimage{rtf} \addlegendentry{R2F};
    \addlegendimage{rrt} \addlegendentry{RRT};
    \addlegendimage{oursfull} \addlegendentry{\scalebox{0.95}{\ours (\scalebox{0.95}{$\Lxtest\hspace{-2pt}=\hspace{-2pt}50)$}}};
    \addlegendimage{oursdist} \addlegendentry{\ours ($\Lxtest\hspace{-2pt}=\hspace{-2pt}600)$};

    \addplot[rrt] coordinates {%
                               (5.84,73.61)      %
                               (6.27,74.05)      %
                               (7.11,74.22)      %
                               (9.98,74.41)      %
                               (19.84,74.38)     %
                               (32.28,74.35)};   %

    \addplot[oursreal] coordinates {%
                                    (6.86,74.98)      %
                                    (7.54,75.80)      %
                                    (8.44,76.19)      %
                                    (13.10,76.43)     %
                                    (29.62,76.66)     %
                                    (51.71,76.54)};   %

    \addplot[rtf] coordinates      {%
                                    (8.40,74.09)      %
                                    (8.82,74.52)      %
                                    (9.41,74.81)      %
                                    (14.40,75.11)     %
                                    (28.56,75.36)     %
                                    (47.88,75.26)};   %

    \addplot[oursdist] coordinates {%
                                    (46.38,76.49)      %
                                    (51.71,77.91)      %
                                    (57.08,78.87)      %
                                    (72.69,79.40)      %
                                    (100.80,79.68)     %
                                    (146.59,79.70)};   %
                                    
    \node [above] at (axis cs:  6.0,75.0) {\footnotesize \textcolor{black}{20}};
    \node [above] at (axis cs:  6.5,75.9) {\footnotesize \textcolor{black}{50}};
    \node [above] at (axis cs:  8.2,76.55) {\footnotesize \textcolor{black}{100}};
    \node [above] at (axis cs:  13.0,76.75) {\footnotesize \textcolor{black}{200}};
    \node [above] at (axis cs:  29.2,77.0) {\footnotesize \textcolor{black}{400}};
    \node [above] at (axis cs:  61.7,76.4) {\footnotesize \textcolor{black}{600}};
\end{axis}
\end{tikzpicture}

%% file: tex/conclusions.tex
\section{Conclusions}
\label{sec:conclusions}
This work highlights the practical importance of keeping the memory requirements in retrieval low and provides an experimental benchmark to allow future comparisons.
We propose a new image-to-image similarity model that optimizes the performance \vs memory trade-off in the best way compared to existing approaches.
The comparisons are performed in a direct and fair way based on the same representation backbone and using the same implementation framework whenever possible. 
We anticipate local similarity will be a key ingredient for tasks at a scale much larger than the current largest databases.

%% file: tex/supplementary.tex
\section{Additional results}

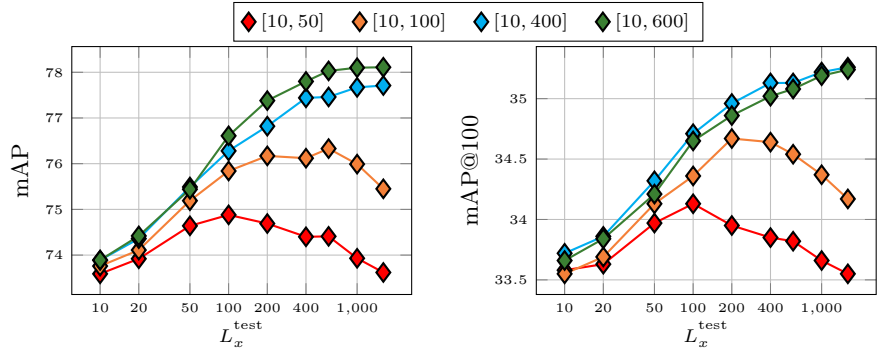
\begin{figure}[h!]
  \vspace{-10pt}
  \centering
  \hspace{-10pt}
  \input{fig/range_revop.tex}
  \hspace{-80pt}
  \input{fig/range_gld.tex}
  \vspace{-10pt}
  \caption{\textbf{Impact of $\Lxtrain$, $\Lqtrain$ sampling range.} \ours is trained 
  by limiting the descriptor set size to a particular range during training while testing for all set sizes, both within and outside the range used for training.
  Performance evaluated on $\cR$OP+1M (left) and GLDv2 (right) for \ours.
  All runs use global descriptors with PQ8 for initial ranking and ensemble similarity to re-rank $m\hspace{-2pt}=\hspace{-2pt}1600$ images. 
  \vspace{-30pt}
  \label{fig:range_revop_gld}
  }
\end{figure}

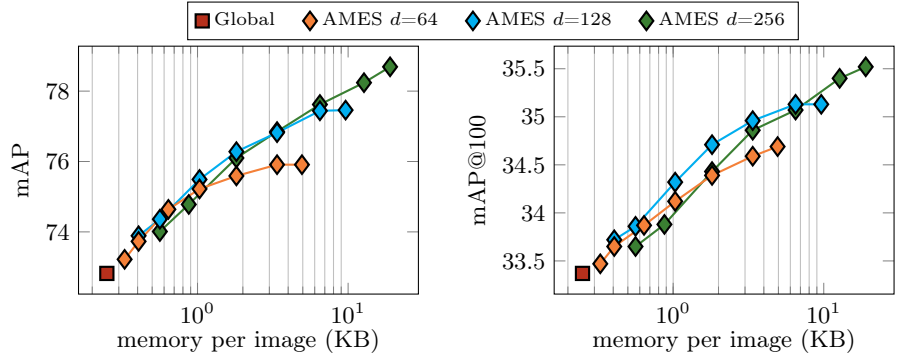
\begin{figure}[h!]
  \centering
  \vspace{10pt}
  \hspace{-10pt}
  \input{fig/dim_revop.tex}
  \hspace{-100pt}
  \input{fig/dim_gld.tex}
  \caption{\textbf{Impact of dimensionality $d$}. 
  Performance evaluated on $\cR$OP+1M (left) and GLDv2 (right) for \ours.
  All runs use global descriptors with PQ8 for initial ranking and ensemble similarity to re-rank $m\hspace{-2pt}=\hspace{-2pt}1600$ images.
  \label{fig:dim_revop_gld}
  }
\end{figure}

\clearpage
\newpage

\begin{table}[t]
\vspace{-5pt}
  \centering
  \input{tab/dist_beta}
  \captionof{table}{\textbf{Impact of distillation}. $Z_N$ stands for distillation with \l2 loss on the output tokens (proposed method). $S$ stands for distillation with \l2 loss on the final local similarity scores. $Z_N$ and $\beta=10$ are the default choices for \ours.
  \label{tab:dist_beta}
  \vspace{-20pt}
  }
\end{table}

\begin{figure}[h!]
    \begin{minipage}[!t]{0.49\textwidth}
      \centering
      \input{tab/model_size}
      \vspace{-10pt}
      \captionof{table}{
      \textbf{Impact of network depth}. $N=0$ stands for the performance of the global-only similarity for reference. $N=5$ is the default choice for \ours.
      \label{tab:model_size}
      \vspace{-0pt} %
      }
    \end{minipage}%
    \hspace{3pt}
    \begin{minipage}[!t]{0.49\textwidth}
      \centering
      \vspace{6pt}
      \input{tab/binarization}
      \vspace{-10pt}
      \captionof{table}{
      \textbf{Impact of binarization settings}. \textbf{ITQ} stands for initialization with ITQ~\cite{glg+12}. \textbf{FT} stands for fine-tuning $W$, as opposed to keeping it frozen. ITQ and FT are the defaults for \ours.
      \label{tab:binarization}
      }
    \end{minipage}%
\vspace{-10pt}    
\end{figure}

In this section, we provide additional experimental results with different model settings and training hyper-parameters. Unless specified otherwise, $\Lqtest\hspace{-2pt}=\hspace{-2pt}600$ and $\Lxtest\hspace{-2pt}=\hspace{-2pt}50$, global descriptors with PQ8, $m\hspace{-2pt}=\hspace{-2pt}1600$ and \ours binary without distillation are used.

\textbf{Impact of $\Lxtrain, \Lqtrain$ sampling range.} 
Figure~\ref{fig:range_revop_gld} displays the results of \ours trained with four different sampling ranges. In the smaller sampling ranges, performance increases until \Lxtest is close to the maximum range used during training. %
After that point, performance almost consistently drops or remains the same. This behavior aligns with the observation in prior work in different research field~\cite{vb21}, where transformer-based models are trained and tested with different input sequence lengths. In the larger sampling ranges, performance steadily improves as \Lxtest increases. It saturates for the larger descriptor set sizes, \ie performance gains are marginal for \Lxtest greater than 600 with significant memory and computation overhead. The two larger sampling ranges report very close results; hence, we select $[10, 400]$ in our default settings for better efficiency in training time and memory allocation. Nevertheless, training with an even larger sampling range could potentially yield even better performance.

\textbf{Impact of dimensionality $d$.} 
To further investigate the memory footprint and performance trade-off, we explore two additional dimensionalities of the binary local descriptors in Figure~\ref{fig:dim_revop_gld}.
In low memory regimes, using more local descriptors with lower dimensions is advantageous. On the other hand, it is preferable to use fewer higher-dimensional descriptors instead of including all available ones in the high memory regime. A subset of the descriptors carries most of the necessary information, and adding more introduces redundancy or noise in the matching.

\textbf{Impact of distillation.} 
In Table~\ref{tab:dist_beta}, we evaluate the impact of the hyper-parameter $\beta$, and we compare our distillation scheme with another alternative that applies distillation on the output similarity scores, commonly used in the literature~\cite{pkl+19,ktp+22}. Different values of $\beta$ do not significantly affect performance. Our distillation scheme considerably outperforms the similarity-based approach. This is expected considering that the latter may oppose the supervision loss, whereas ours is complementary.

\textbf{Impact of network depth.} 
Table~\ref{tab:model_size} reports the performance of \ours implemented with various numbers of blocks $N$. The performance saturates after using more than three transformer blocks.

\textbf{Impact of binarization settings.} 
In Table~\ref{tab:binarization}, we assess our implementation choices for our binarization scheme, \ie the initialization with ITQ and the fine-tuning of the trainable parameters of the matrix $W$. Both choices are necessary to achieve the final performance, with the proper initialization having a significant impact since the model does not achieve similar performance while learning $W$ from scratch.

\textbf{Performance mean and standard deviation.}
For future reference, we provide the numeric results of our \ours model on the different $\mathcal{R}$Oxford and $\mathcal{R}$Paris settings, as well as the private split of GLDv2, in Table~\ref{tab:600-50_std}. 
Results of \ours in Table~\ref{tab:sota} are repeated in Table~\ref{tab:sota_std} with their standard deviation.
The mean and standard deviation of three training runs with different seeds are reported.

\begin{table*}[!t]
  \centering
  \vspace{10pt}
  \scalebox{0.68}{
  \input{tab/600-50_std}
  }
  \caption{\textbf{\ours performance for different settings} reported separately per dataset with standard deviations across three experiments using a different seed. mAP used for $\cR$Oxf and $\cR$Par. mAP@100 used for GLDv2.
  \label{tab:600-50_std}
  \vspace{-10pt}
  }
\end{table*}

\begin{table*}[!t]
  \centering
  \scalebox{0.6}{
  \input{tab/sota_std}
  }
  \caption{\textbf{\ours performance for different backbones} reported separately per dataset with standard deviations across three experiments using a different seed. mAP used for $\cR$Oxf and $\cR$Par. mAP@100 used for GLDv2. 
  Global descriptors are in full precision.
  \label{tab:sota_std}
  \vspace{-10pt}
  }
\end{table*}

\begin{figure}[t]
  \centering
  \vspace{5pt}
  \input{fig/topk_revop}
  \vspace{-5pt}
  \caption{\textbf{Impact of re-ranking} the top-$m$ images obtained by global similarity.
  The global and local similarity ensemble is effective, but local similarity fails by itself. Results on $\cR$OP+1M.
  \label{fig:topk}
  \vspace{-5pt}
  }
\end{figure}
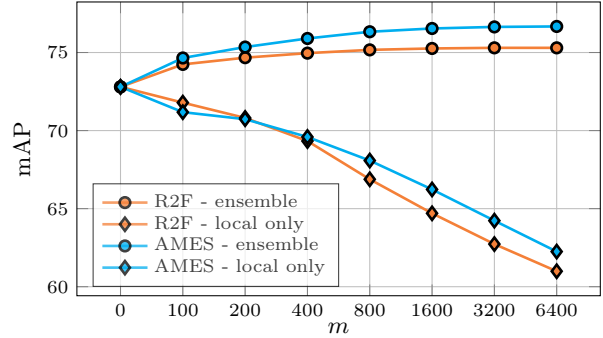

\textbf{Top-$m$ re-ranking. }
Figure~\ref{fig:topk} shows the retrieval performance for re-ranking using local similarity only or with the global-local ensemble. 
Local similarity does not work by itself; it harms the performance compared to the global similarity ranking.
By contrast, the ensemble similarity consistently increases performance as the number of re-ranked images increases. The same holds both for our method and \rtf.
This observation is likely a byproduct of the high-performing global descriptors used in our work and is not fully aligned with the original findings by other methods using older descriptors, such as RRT using DELG.

\begin{figure}[t]
  \centering
  \input{fig/sa_vs_ca_revop}
  \vspace{-5pt}
  \caption{\textbf{\ours \vs RRT} for varying \Lxtest, and $\Lqtest=600$ on $\cR$OP+1M.
  RRT is modified to remove the global descriptor from its input tokens.
  \label{fig:arch}
  \vspace{-5pt} %
  }
\end{figure}

\textbf{Transformer architecture comparison. }
In our experiments with RRT, we observe a small performance decrease by including the global descriptor in the input token set, which is part of the original RRT architecture.
We attribute this to the difficulty of mapping both descriptors in the same space and the insignificant value of one extra token compared to the many tokens of the local descriptors.
The results of Figure~\ref{fig:tradeoff_revop_gld} are obtained with the original RRT setup.
In addition to that, we use the RRT model architecture and train with the same input token set as \ours, \ie excluding the global descriptor.
Results are presented for the binarized variants of both models in Figure~\ref{fig:arch}, where \ours seems to consistently outperform RRT even after our fix to it.

\begin{figure}[!t]
  \centering
    \vspace{20pt}
    \input{fig/sim_global_local/q56_OxfH+1m} 
    \input{fig/sim_global_local/q26_OxfH+1m}
    \vspace{5pt}
    \hspace*{-12pt}
    \input{fig/sim_global_local/q65_OxfH+1m}
    \hspace{-12pt}
    \input{fig/sim_global_local/q60_OxfH+1m}
    \vspace{5pt}
    \hspace*{-10pt}
    \input{fig/sim_global_local/q18_OxfH+1m}
    \hspace*{-5pt}
    \input{fig/sim_global_local/q69_OxfH+1m}
  \caption{\textbf{Comparison of global and local similarity.} Each figure shows the top 400 retrieved images after the initial ranking for a single query from $\mathcal{R}$Oxf+1M in the hard setting. $\Delta$AP represents the change in average precision for the specific query between using global similarities only and our ensemble similarities.
  Matching (positive) and non-matching (negative) are according to the query image ground truth.
  \label{fig:sim_global_local}}
\end{figure}
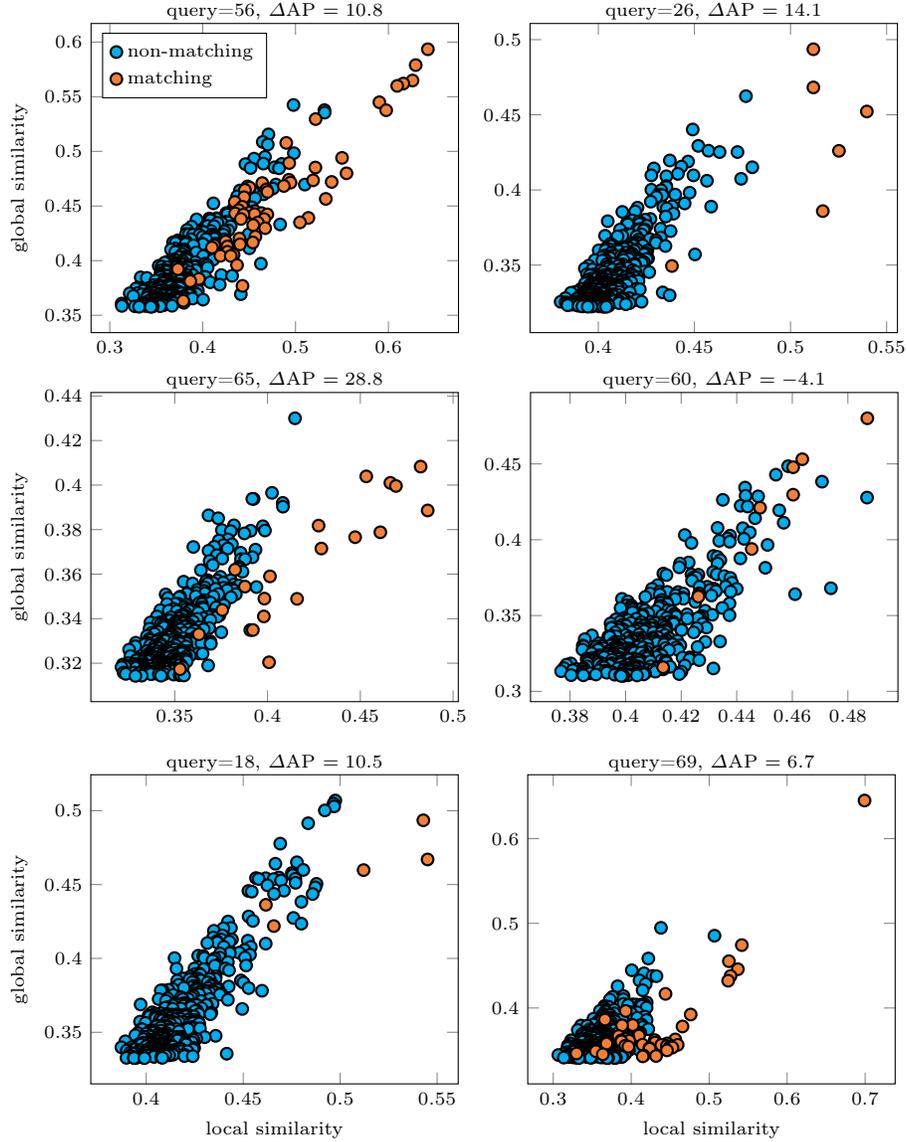

\textbf{In-depth analysis of global-local similarity}
Using an ensemble of global and local models to compute the final similarity for each pair of images is clearly beneficial based on quantitative results. In Figure~\ref{fig:sim_global_local}, we present the global and local similarities, estimated by the corresponding models separately, for six selected queries.
In several cases, the two types of similarities are not linearly correlated, making it easier to separate matching and non-matching pairs. 

\section{Additional implementation details}

\begin{figure}[t]
\vspace{15pt}
  \centering
  \includegraphics[width=0.99\linewidth]{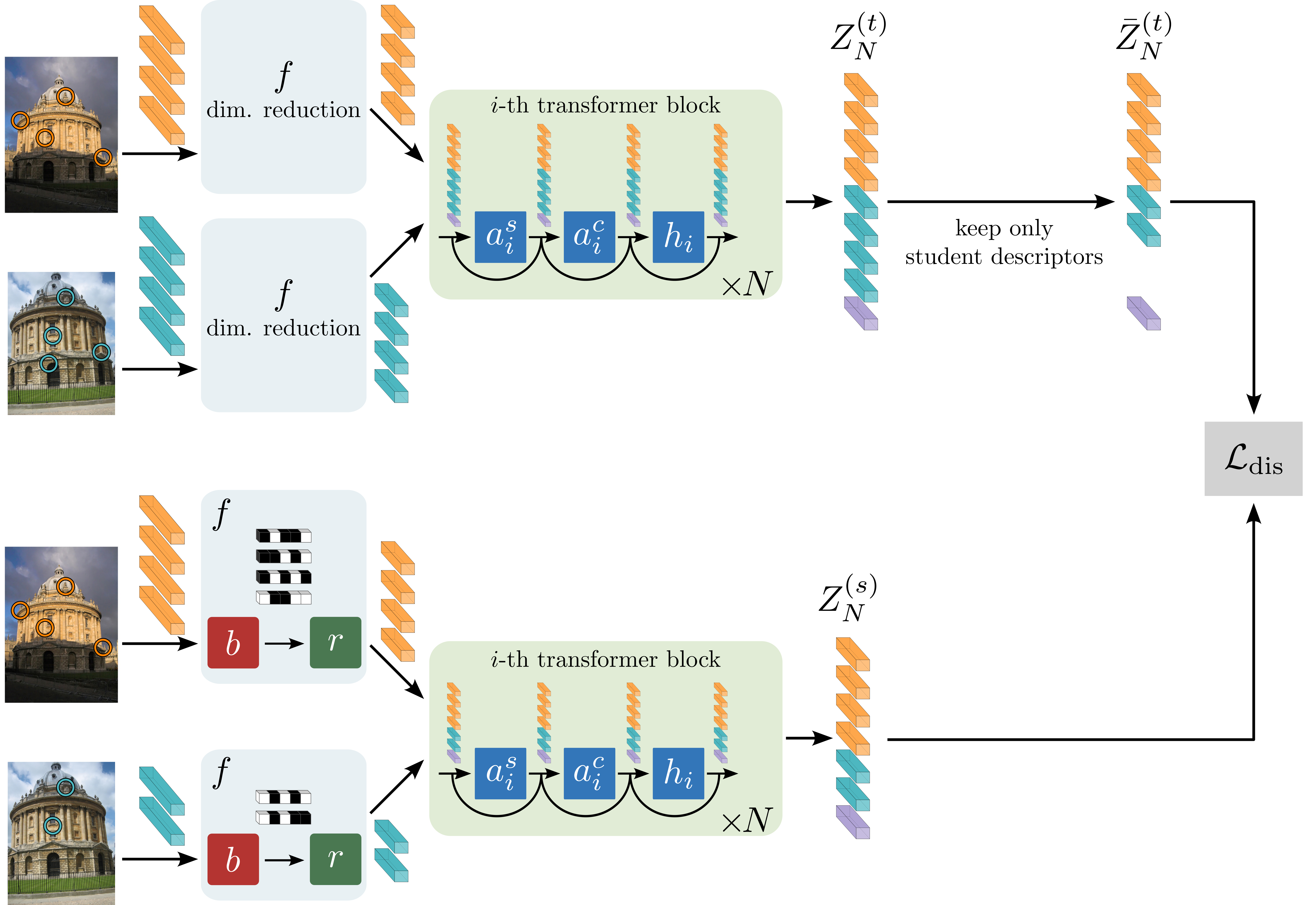}
  \caption{\textbf{The distillation process of \ours.} 
  The teacher model (top) distills its output tokens to the student model (bottom)  via the root mean squared error loss. 
  The student operates in an asymmetric way on a subset of the local descriptors used by the teacher for the database image. 
  The loss is applied to the intersection of the two sets.
  Function $f$ refers to a different function for the teacher (dimensionality reduction by a linear layer) and the student (dimensionality reduction, binarization, and re-mapping to the real coordinate space). 
  The distillation loss is combined with the binary cross-entropy loss.
  \vspace{15pt}
  \label{fig:distill}
  \vspace{5pt}}
\end{figure}

\textbf{Backbone architecture and training process.}
Our network is trained for 15 epochs with 300 batch size. Every batch consists of 100 triplets, in which one training sample acts as an anchor. All training samples are used as an anchor exactly once per epoch. The rest of the triplet consists of a matching sample from the same class as the anchor and a non-matching sample. Both are randomly drawn from the anchor's 300 nearest neighbors with a distribution proportional to the cube of their global similarities. The network is optimized with AdamW~\cite{lh+19} and cosine learning rate scheduler with the initial value of $2\cdot10^{-4}$. The variance $\delta$ for the binarization layer is set to $10^{-3}$, following~\cite{ktp+22}.
Figure~\ref{fig:distill} shows the distillation process between the teacher and the student models. 

In the experiments with CVNet, we follow the extraction pipeline from the original paper. We feed the input images in 3 resolutions and extract the local descriptors from the penultimate layer feature map. When using SuperGlobal, we skip Scale-GeM and ReLUP to reproduce performance close to the reported one.
SG is reported to achieve 73.4 and 33.4 on $\cR$OP+1M and GLDv2, respectively, without re-ranking and 81.2 and 35.0 with re-ranking $m\hspace{-2pt}=\hspace{-2pt}1600$ images; see Table 1 for our reproduction. 
For DINOv2~\cite{odm+24} experiments, we use its ViT-B/14 variant with registers to extract patch tokens and CLS tokens as our local and global descriptors. We resize the input images such that the larger image side equals 518 pixels and pad the rest of the image to a square. 

The local feature detector architecture is adopted from DOLG~\cite{yhf+21}. It consists of two $1\times 1$ convolution layers followed by a Softplus activation. There is a BatchNorm~\cite{is15} and ReLU in between the two convolutions.
It is applied on top of a dense 3D activation map to provide a weight per local descriptor.
For CVNet, the detector is applied on top of the third ResNet101 block output of 1024 dimensions (3D activation map depth). 
For DINOv2, the detector is applied on top of the ViT output, \ie the set of patch tokens of 768 dimensions.
During training, we use these local descriptor weights to extract a global descriptor by performing weighted average pooling and train with triplet loss. 
The triplets are sampled in the same manner as in \ours; the margin for the loss is set to 0.9. We train only the detector part on top of a frozen backbone for 20k iterations with a batch size of 10. We use AdamW optimizer and cosine learning rate scheduler with an initial value equal to $10^{-4}$.
During testing, we use the weights to select the $L$  strongest local descriptors per image. This acts in the form of a classical local feature detector.

\textbf{Estimation of mAP.} We rely on two different implementations for estimating mAP on \roxf/\rpar or mAP@100 on GLDv2. We choose the corresponding implementations that use the trapezoids for the area under the curve or average precision values, respectively, in our effort to better match the standard practice in the prior work. 

\textbf{Global-local ensemble parameter tuning.}
The optimal parameters for each (\Lqtest,\Lxtest) setting are tuned independently. We conduct a hyper-parameter tuning via grid search on the validation set to find the values for $\lambda$ and $\gamma$ from the global-local ensemble and local similarity, respectively. We use the public split of the GLDv2 as the validation set for tuning and re-rank $m=400$ images. 
Table~\ref{tab:hyperparam_tuning} illustrates grid search results of an example run with $\Lqtest\hspace{-2pt}=\hspace{-2pt}600$ and $\Lxtest\hspace{-2pt}=\hspace{-2pt}600$. Higher values of $\lambda$ are usually paired with higher values of $\gamma$. Consequently, even when $\lambda$ is large, its most confident predictions are accounted for in the final ensemble. The differences for $\lambda=0$ among different rows are due to the ties in the ranking of database images; many local similarities are either 0 or 1 with very large $\gamma$.

\begin{table}[t]
  \centering
   \vspace{5pt}
  \input{tab/hyperparam_tuning}
  \caption{\textbf{Global-local ensemble tuning} via grid search on the validation set (GLDv2 public retrieval benchmark). 
  Performance measured with mAP@100.
  \label{tab:hyperparam_tuning}}
  \vspace{-15pt}
\end{table}

\textbf{Competing methods.}
We use the publicly available implementations for all the competitors and employed methods, \ie ASMK\footnote{\rurl{github.com/jenicek/asmk}}, RRT\footnote{\rurl{github.com/uvavision/RerankingTransformer}}, \rtf\footnote{\rurl{github.com/bytedance/R2Former}}, CVNet\footnote{\rurl{github.com/sungonce/CVNet}}, and SuperGlobal\footnote{\rurl{github.com/ShihaoShao-GH/SuperGlobal}}. For a fair comparison in our trade-off evaluation, we follow a similar tuning process for all competing approaches. %

ASMK includes an internal quantization process that aggregates vectors per cell; therefore, the effective number of local descriptors that need to be stored is typically lower than the input number. Also, a visual word \emph{id} needs to be stored for each effective descriptor, which amounts to 2 bytes per descriptor if stored as an unsigned integer, using delta coding. The total memory footprint is derived from the sum of the effective descriptor vectors and their word \emph{ids}. The binary and simplified variant~\cite{tjc20} is used, while local similarity is estimated for all the database images as this is more of an indexing than a re-ranking approach. As the input to ASMK, we perform dimensionality reduction down to 128 dimensions by PCA whitening~\cite{jc12} learned on the training set, while its internal representation is comprised of 128-bit vectors.

RRT and \rtf are trained with our implementation framework with varying descriptor set sizes and the same projection function $f$ at their input, as in our fp variant. In both cases, we use $d\hspace{-2pt}=\hspace{-2pt}128$ for the reduction of local descriptors. We use a similar reduction layer for RRT on the global descriptors to map them to the same dimension space as the local ones.
In this way, we train those on exactly the same input local descriptors, training dataset, and training process as \ours.

CVNet's memory footprint of the re-ranker is computed based on the quantized variant of the approach, as reported in Lee \etal~\cite{lsl+22} since it demonstrates a marginal performance drop compared to the full precision variant.

\begin{figure}[t]
	\centering
    \includegraphics[clip, width=1.0\textwidth]{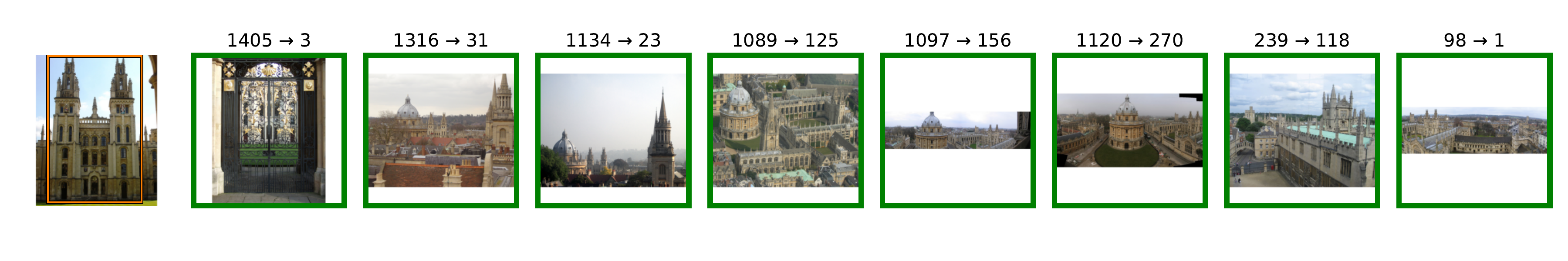}
    \includegraphics[clip, width=1.0\textwidth]{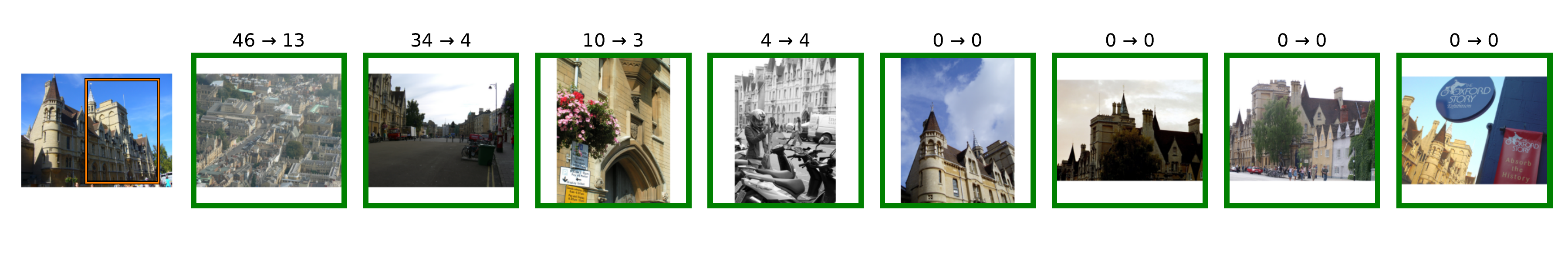}
    \includegraphics[clip, width=1.0\textwidth]{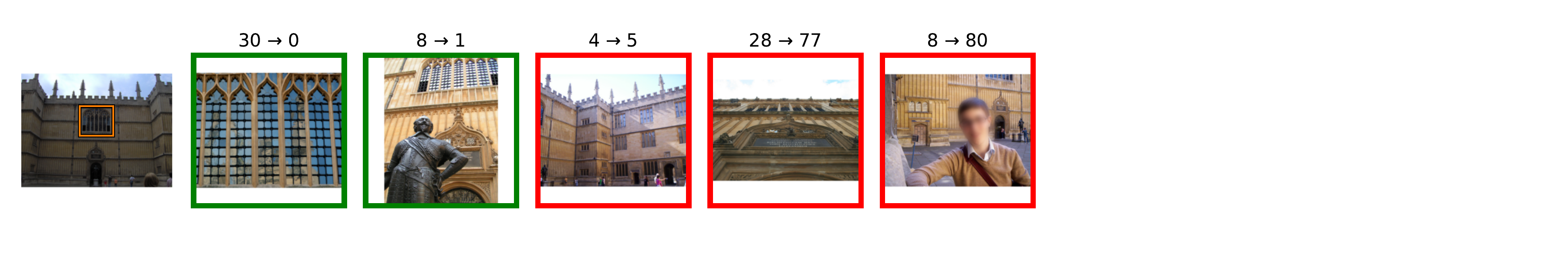}
    \includegraphics[clip, width=1.0\textwidth]{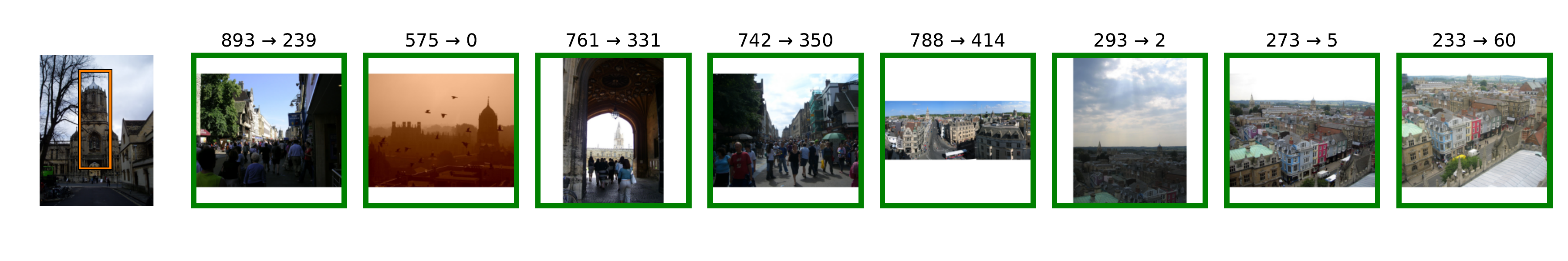}
    \includegraphics[clip, width=1.0\textwidth]{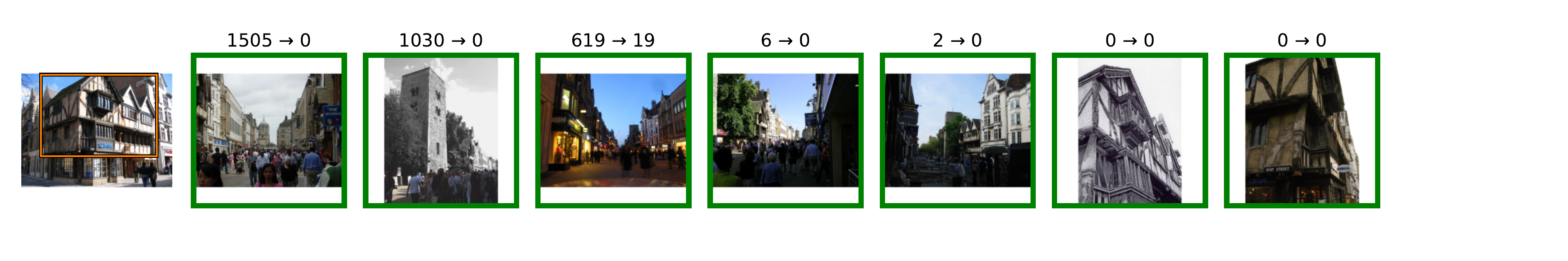}
    \includegraphics[clip, width=1.0\textwidth]{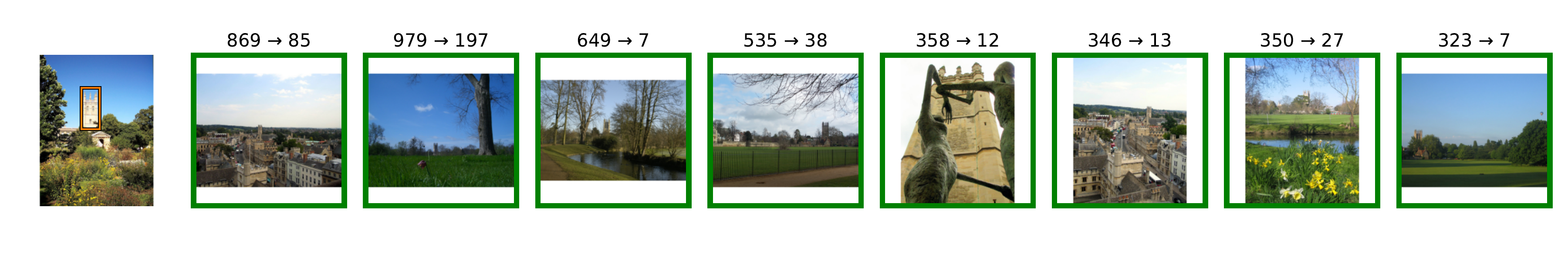}
    \includegraphics[clip, width=1.0\textwidth]{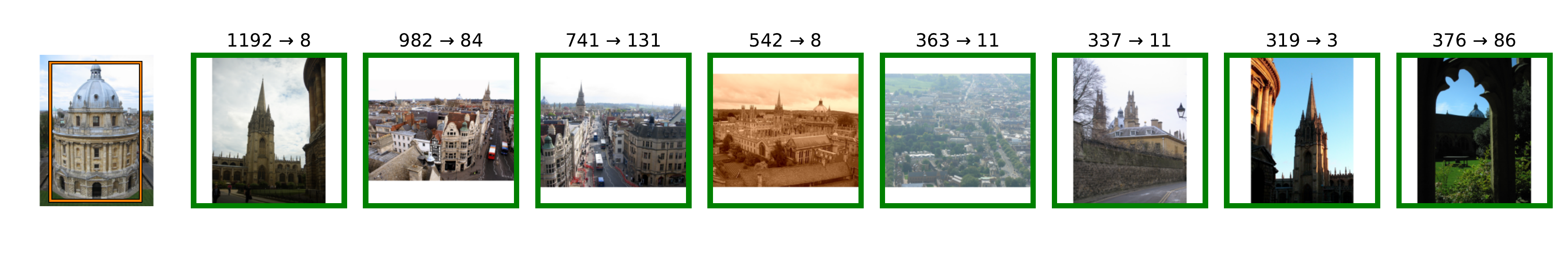}
    \includegraphics[clip, width=1.0\textwidth]{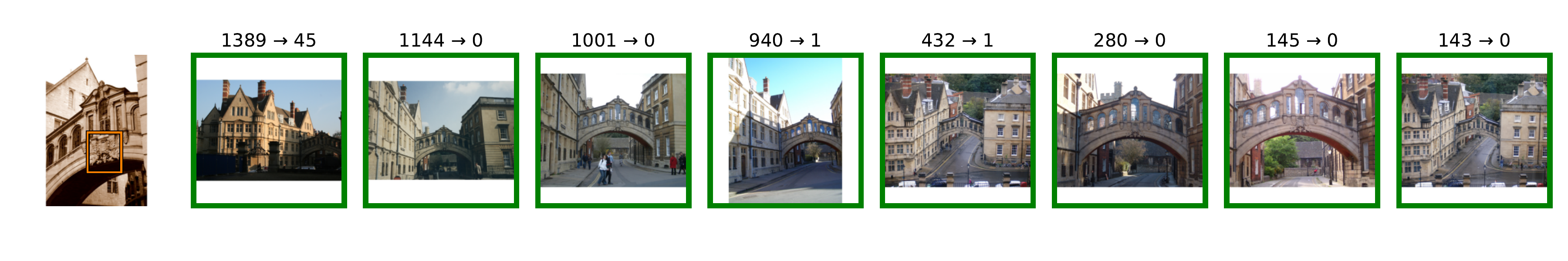}
    \includegraphics[clip, width=1.0\textwidth]{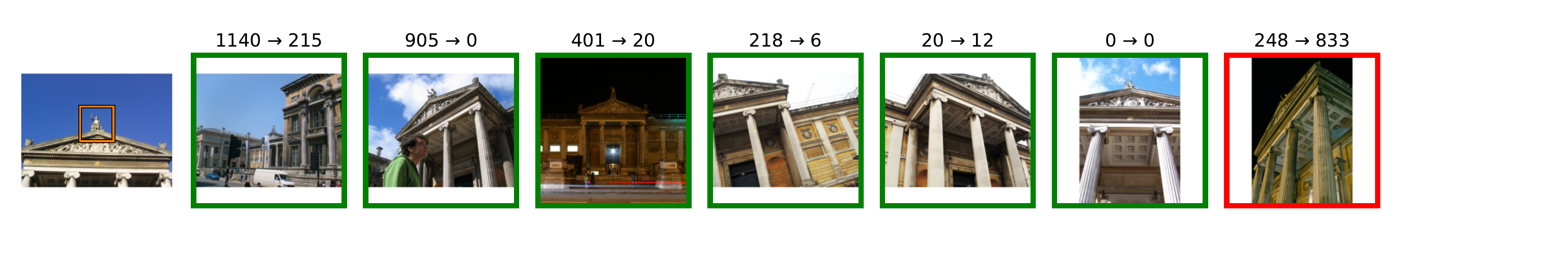}
	\caption{\textbf{The impact of re-ranking with \ours.} We show a query  (region within the orange bounding box) per row and its hard positives from the database. We show the number of negative images ranked before the positive using only global similarity (1) and after re-ranking with \ours (2), by (1)→(2). The positives are ordered based on the difference (1)-(2) in descending order. Green (red) border denotes improved (harmed) re-ranking. 
 Retrieval on  $\cR$Oxford +1M.\label{fig:ranking}}
\end{figure}

\section{Qualitative results}
In Figure~\ref{fig:ranking}, we show examples of hard positive images whose ranking is significantly improved using \ours for re-ranking. The most prominent examples include small objects among severe background clutter. We conclude that local similarity is essential in handling clutter, but current benchmarks only include a small number of such cases. 

\begin{figure}[h!]
\vspace{20pt}
    \begin{tabular}{c c}
      \centering
      \begin{tabular}{c@{\msp}c@{\xlsp}}
        \includegraphics[clip, trim=0 0 0 0, height=0.25\linewidth]{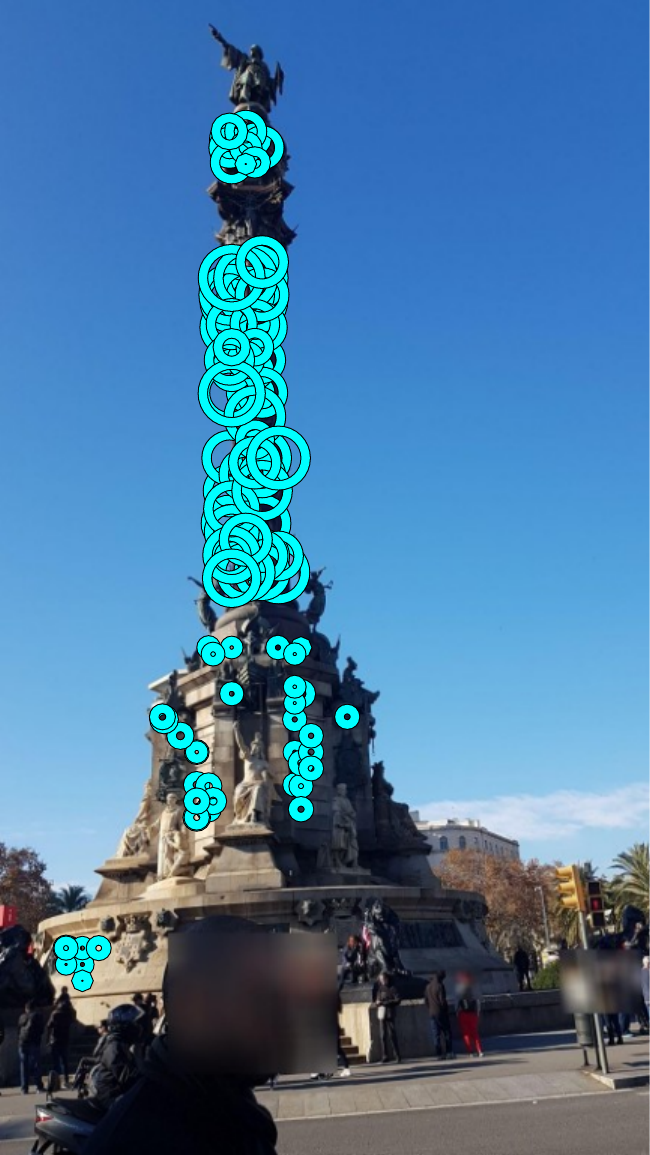}&
        \includegraphics[clip, trim=0 0 0 0, height=0.25\linewidth]{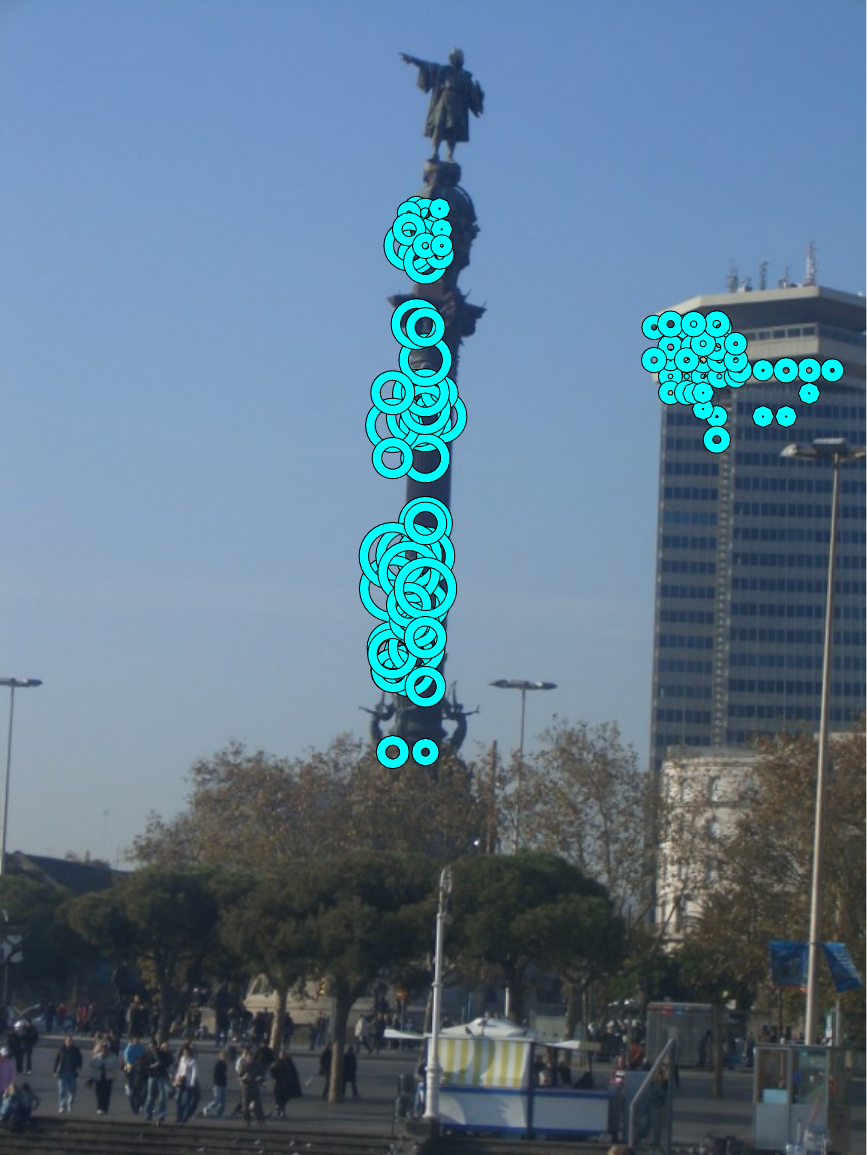}\\
        \includegraphics[clip, trim=0 0 0 0, height=0.25\linewidth]{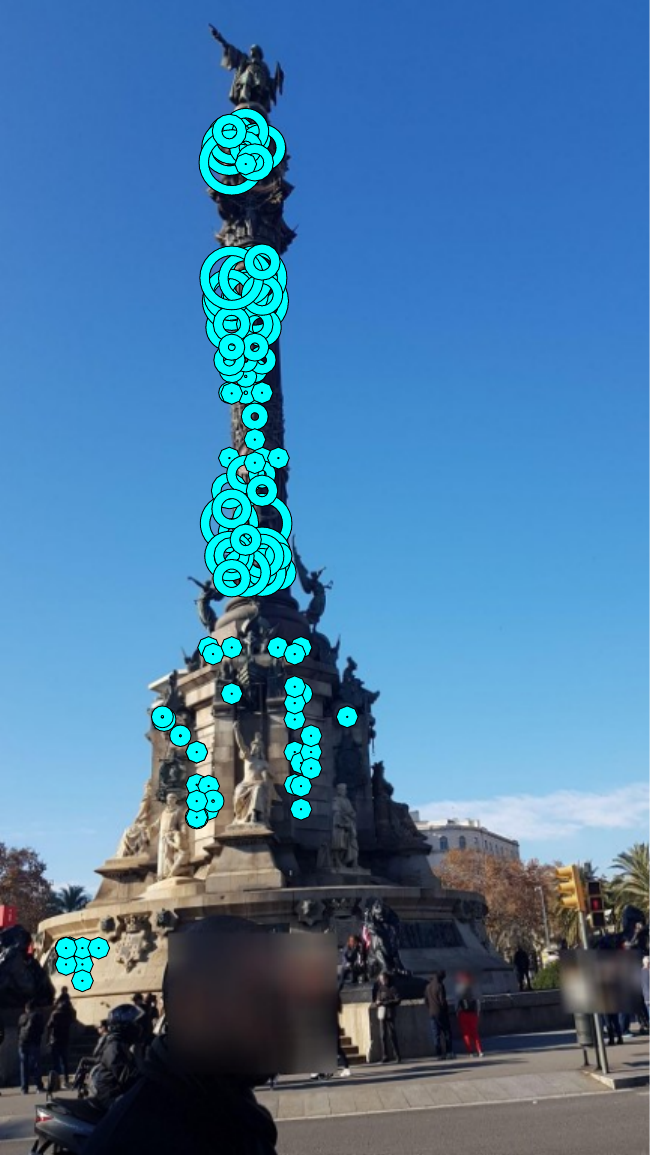}&
        \includegraphics[clip, trim=0 0 0 0, height=0.25\linewidth]{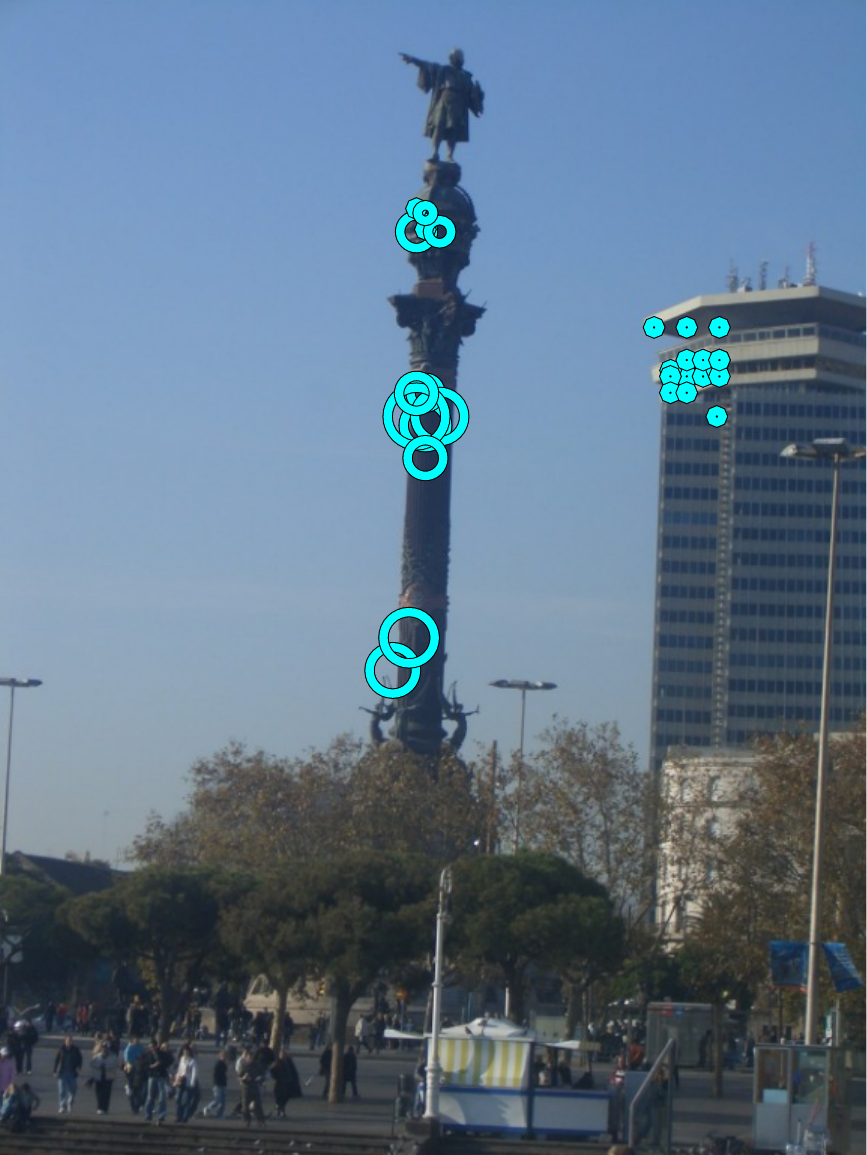}
      \end{tabular} & 
      \begin{tabular}{c@{\msp}c}
        \includegraphics[clip, trim=0 150 0 0, height=0.25\linewidth]{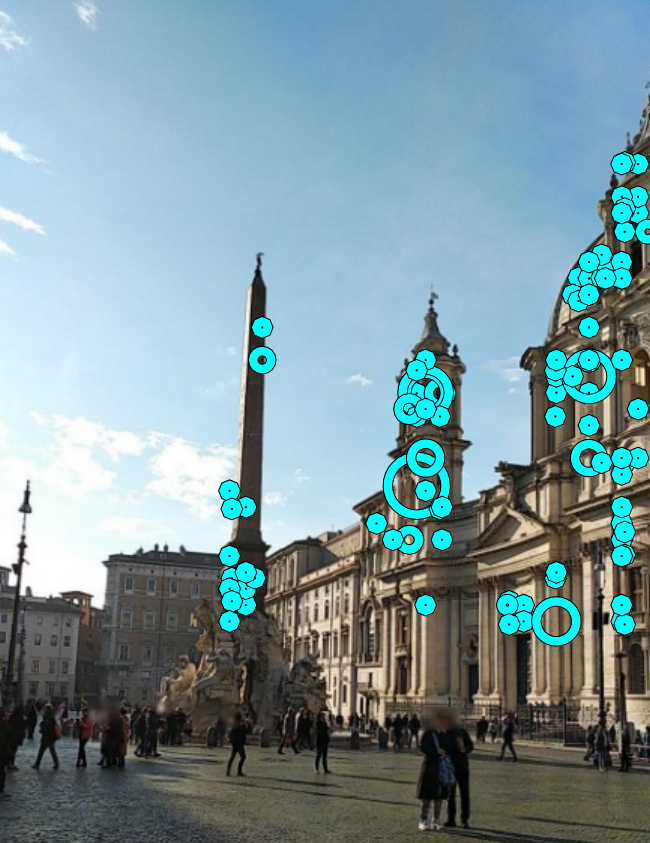}&
        \includegraphics[clip, trim=0 0 0 0, height=0.25\linewidth]{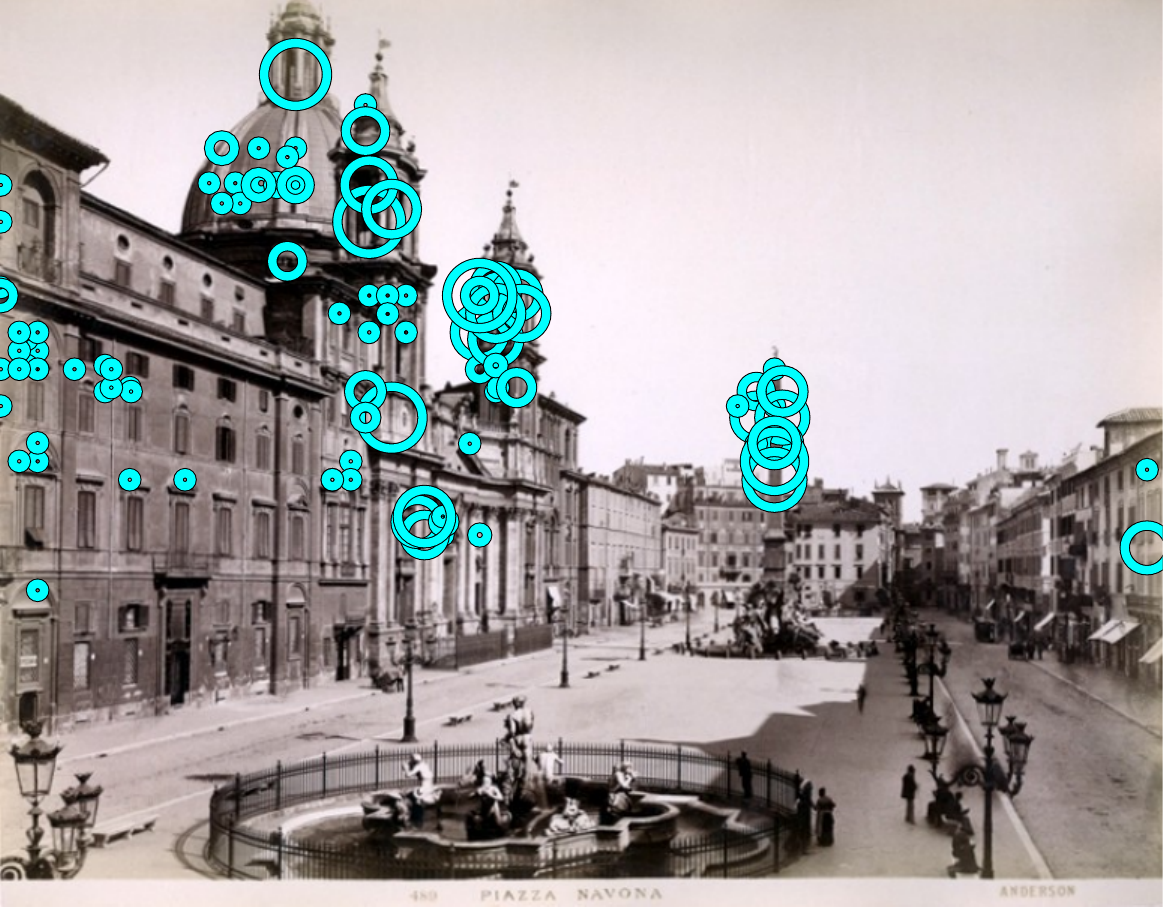}\\
        \includegraphics[clip, trim=0 150 0 0, height=0.25\linewidth]{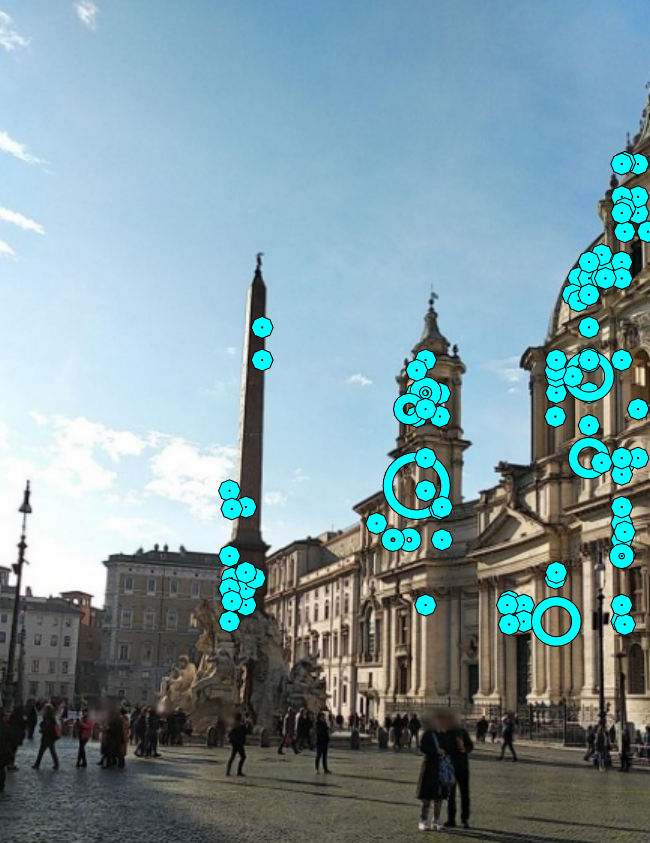}&
        \includegraphics[clip, trim=0 0 0 0, height=0.25\linewidth]{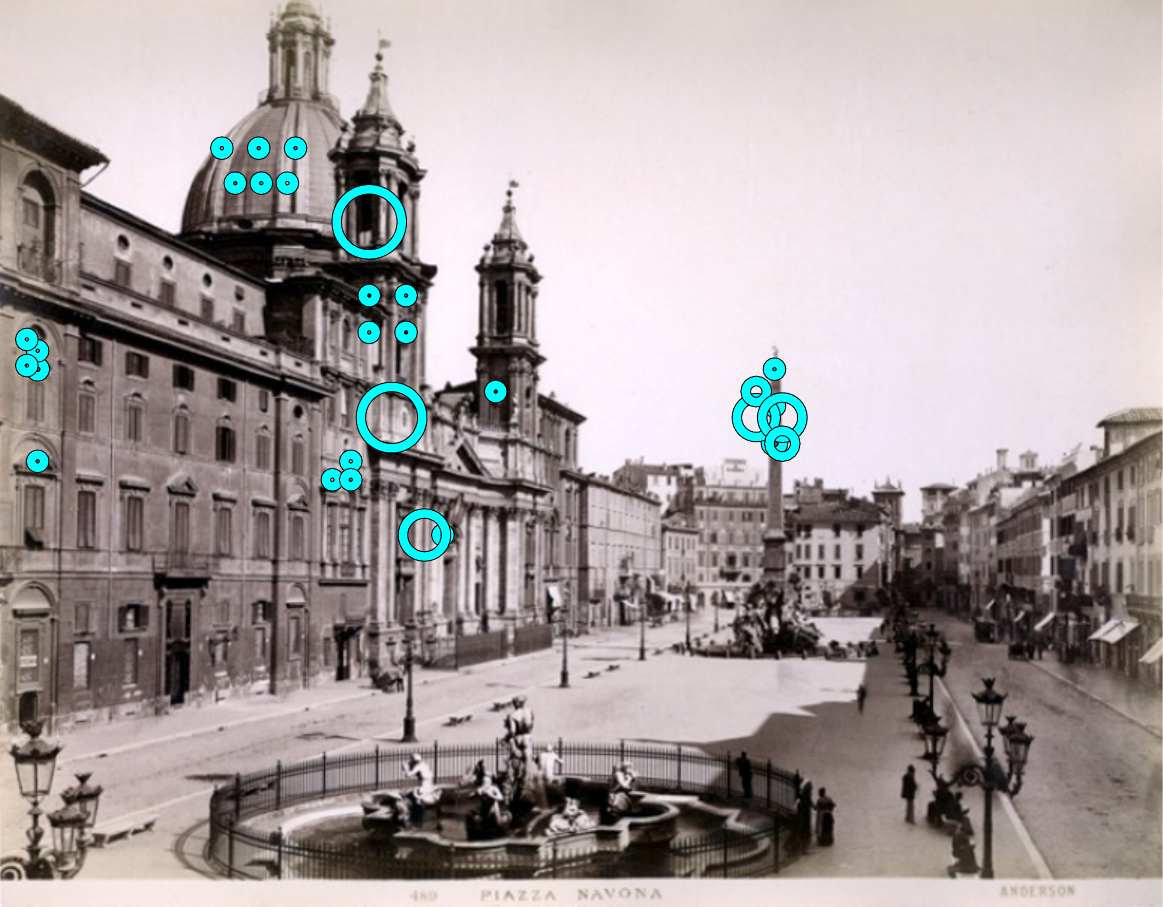}
      \end{tabular} \\[5pt]
      (a) & (b) \\[20pt]
      \multicolumn{2}{c}{
      \begin{tabular}{c@{\msp}c}
          \includegraphics[clip, trim=18 120 25 40, height=0.25\linewidth]{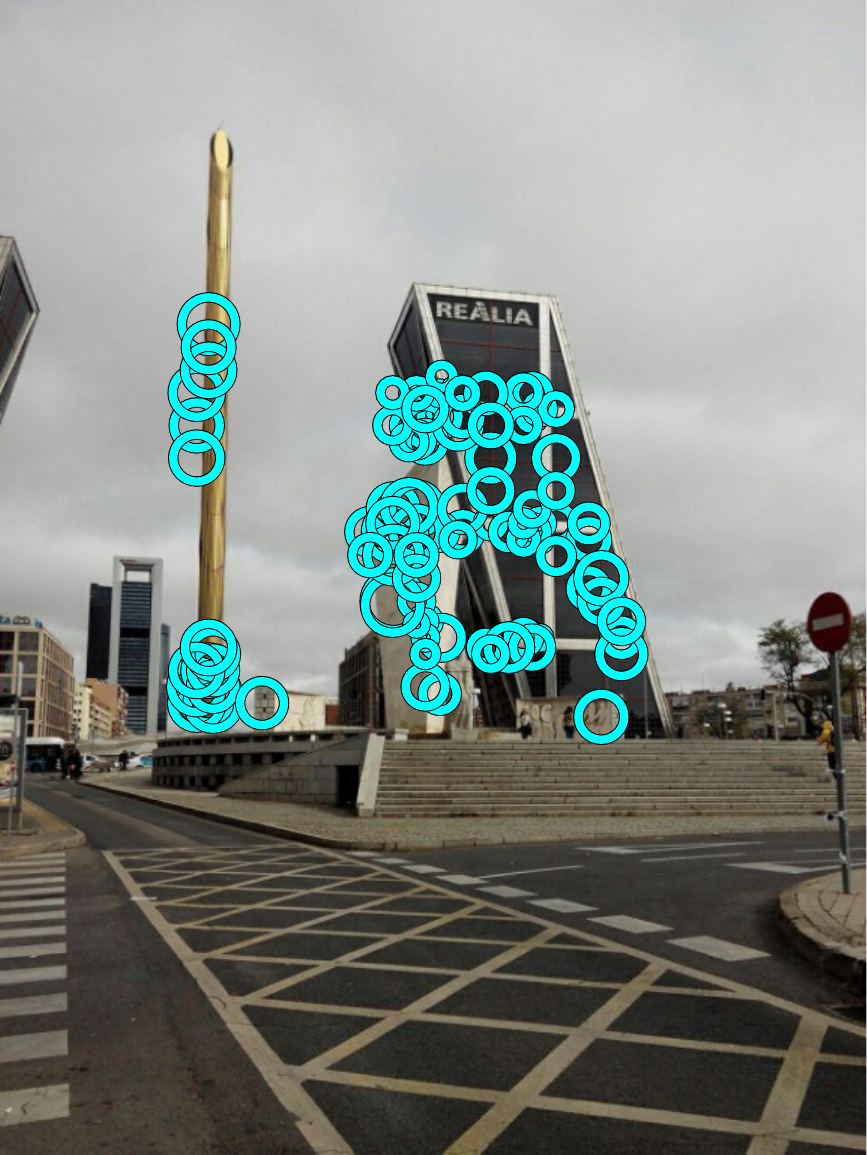}&
          \includegraphics[clip, trim=0 60 30 0, height=0.25\linewidth]{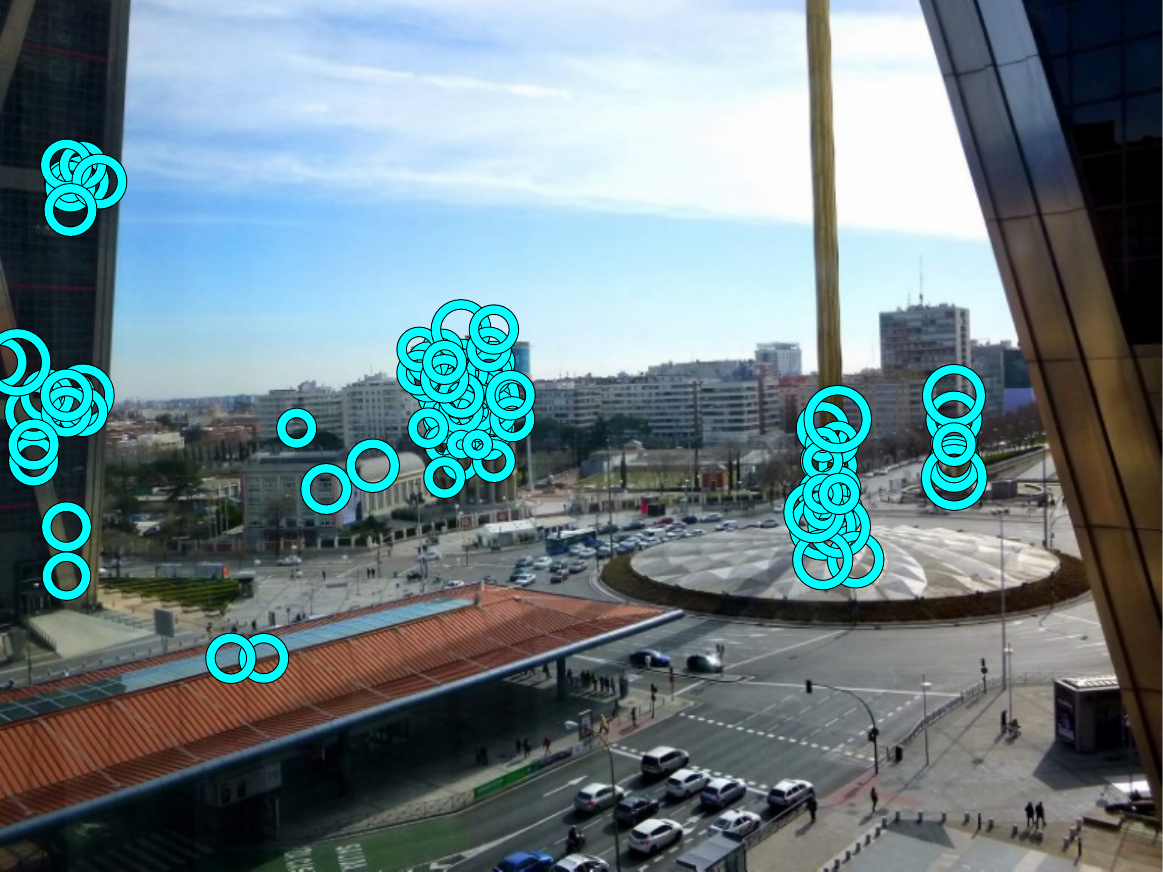}\\
          \includegraphics[clip, trim=18 120 25 40, height=0.25\linewidth]{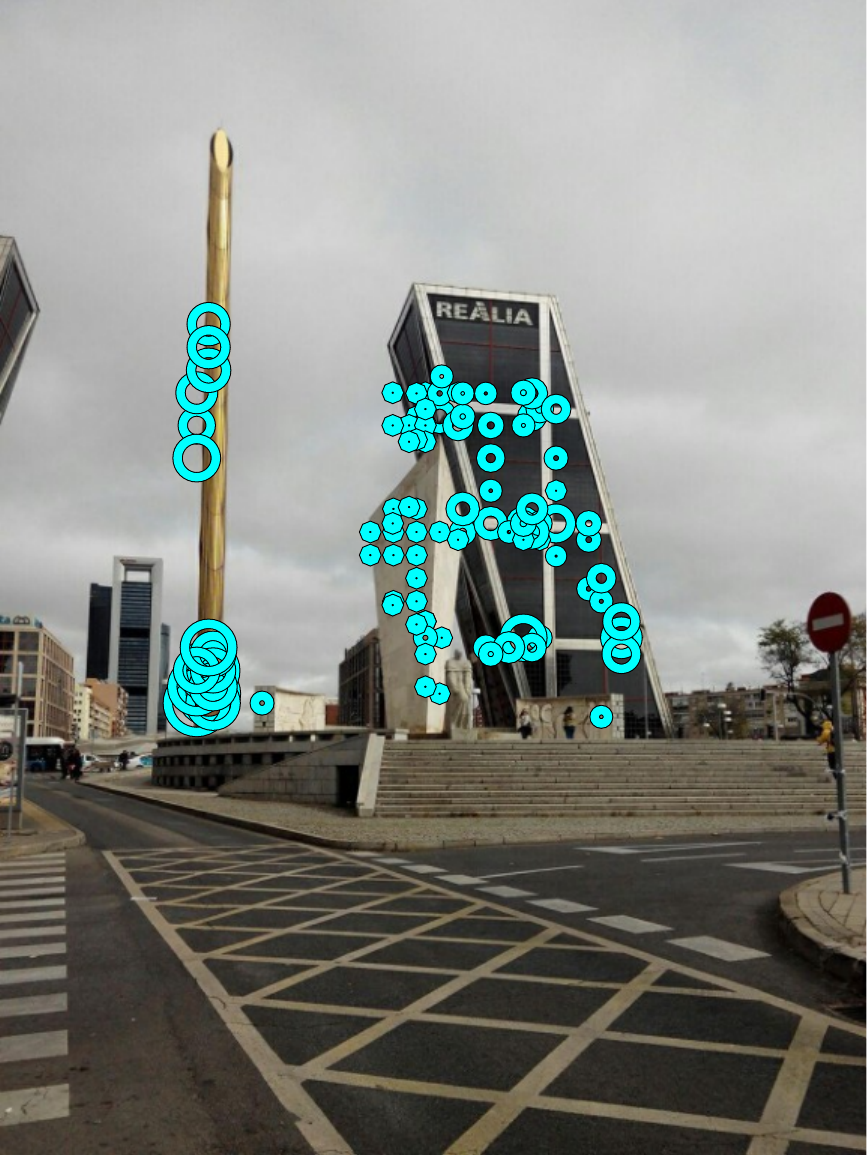}&
          \includegraphics[clip, trim=0 60 30 0, height=0.25\linewidth]{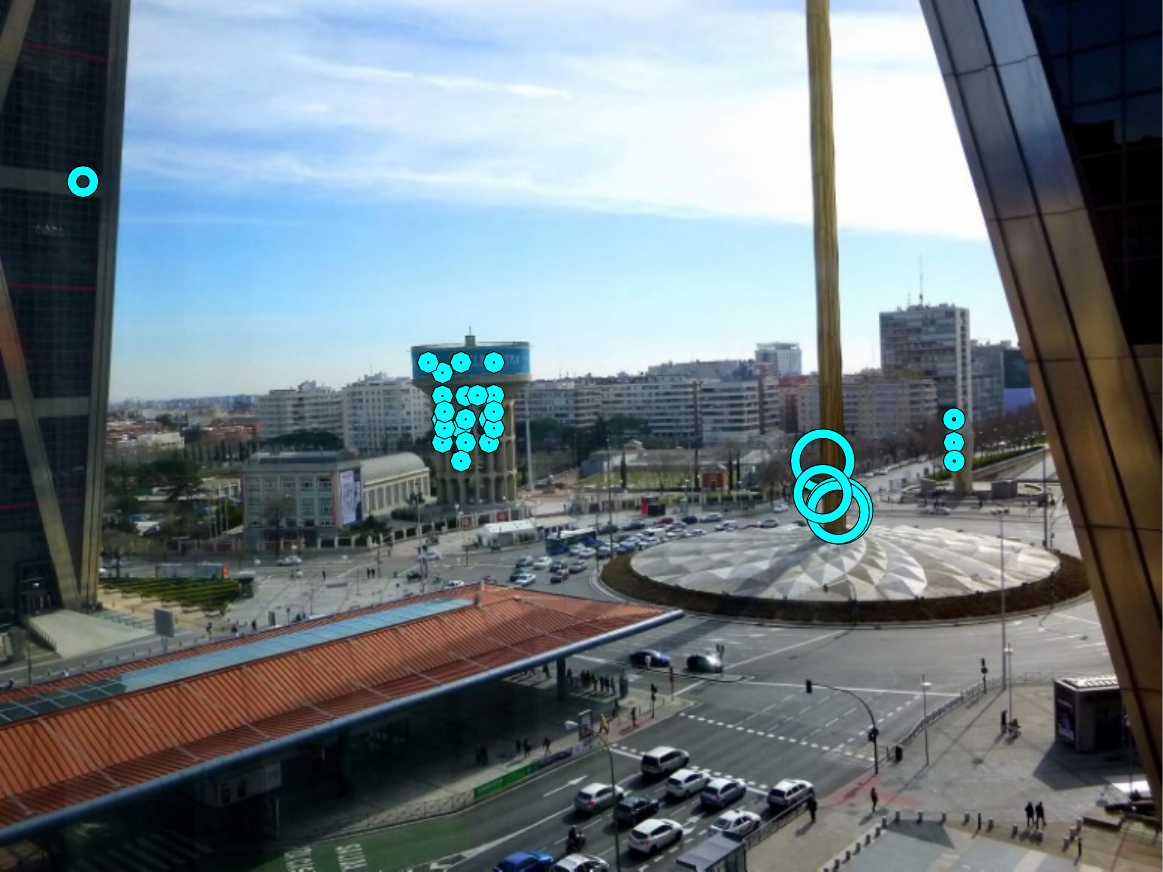}
      \end{tabular}} \\[5pt]
      \multicolumn{2}{c}{(c)}
  \end{tabular}
  \vspace{10pt}
  \caption{Our model estimates the image pair similarity with a low number of local descriptors.
  Circle size reflects importance of the corresponding local descriptor within the model. 
  Top: 100 (query) \vs 100 (database) local descriptors. 
  Bottom: memory efficient asymmetric similarity with 100 \vs 30 local descriptors.
  Descriptors of the common object (other objects) are taken more (less) into account even with the lightweight and asymmetric variant.
  \label{fig:local_asymmetry}
}
\end{figure}

In Figure~\ref{fig:local_asymmetry}, we illustrate several matching image pairs, \ie a query and a database image, and visualize the locations and importance of the local descriptors for the local similarity estimation with our \ours model. We experiment with two different values for \Lxtest, corresponding to symmetric and asymmetric matching. We measure the importance of the local descriptors based on the dot product similarity between the output matching token $t_N$ and the output tokens $X_N$ and $Q_N$ of the two images. The network has learned to ignore the background descriptors that do not have a matching pair in both images. This is prevalent mainly in the limited memory settings, but it is also noticeable in the settings with more descriptors. Note that on the side of the query image, the importance of descriptors also changes between the symmetric and asymmetric settings, which reflects the availability of matching descriptors on the side of the database image. All images are from the GLDv2 test set.

\section{Limitations}
We discuss limitations of \ours and possible improvements of those.
(i) \ours relies on external sources/models for image descriptors, for both global and local. Hence, the quality of the employed descriptors is an important factor for the final performance, regardless of the performance gains \ours introduces. Nevertheless, \ours is agnostic to the type of local descriptors and applicable in an off-the-shelf way. Training both \ours and the backbone representation in an end-to-end manner is possible, but we do not pursue this direction in this work.
(ii) \ours lacks of modeling image geometry. It does not contain any mechanism that encodes the spatial structure of the images. We consider it a promising direction for future work for better generalization of the proposed approach. Our preliminary trials using conventional positional encodings do not bring any performance improvements.

%% file: fig/range_revop.tex
\pgfmathsetmacro{\teasermarkersize}{2.5}
\begin{tikzpicture}
\begin{axis}[%
  width=0.50\linewidth,
  height=0.4\linewidth,
  ylabel={\small mAP},
  xlabel={\scriptsize \Lxtest },
  tick label style={font=\tiny},
  ylabel near ticks, xlabel near ticks, 
  legend pos=south east,
  log ticks with fixed point,
  xlabel style={yshift=1ex},
  xtick={10,20,50,100,200,400, 1000},
  xmode=log,
  ]

    \addplot[ourssmall] coordinates {(10,73.59)
                               (20,73.92)
                               (50,74.64)
                               (100,74.88)
                               (200,74.69)
                               (400,74.4)
                               (600,74.41)
                               (1000,73.93)
                               (1600,73.62)};
    
    \addplot[oursmed] coordinates {(10,73.76)
                                   (20,74.11)
                                   (50,75.19)
                                   (100,75.84)
                                   (200,76.17)
                                   (400,76.12)
                                   (600,76.33)
                                   (1000,75.99)
                                   (1600,75.45)};
                                   
    \addplot[oursbin] coordinates {(10,73.89)
                                   (20,74.36)
                                   (50,75.49)
                                   (100,76.28)
                                   (200,76.82)
                                   (400,77.44)
                                   (600,77.46)
                                   (1000,77.67)
                                   (1600,77.71)};

    \addplot[oursbig] coordinates {(10,73.89)
                               (20,74.42)
                               (50,75.44)
                               (100,76.61)
                               (200,77.38)
                               (400,77.8)
                               (600,78.03)
                               (1000,78.10)
                               (1600,78.11)};
\end{axis}
\end{tikzpicture}

%% file: fig/range_gld.tex
\pgfmathsetmacro{\teasermarkersize}{2.5}
\begin{tikzpicture}
\begin{axis}[%
  width=0.50\linewidth,
  height=0.4\linewidth,
  ylabel={\small mAP@100},
  xlabel={\scriptsize \Lxtest },
  tick label style={font=\tiny},
  ylabel near ticks, xlabel near ticks, 
  legend columns=4,
  legend style={
            at={(-0.2,1.18)},
            anchor=north,
            only marks,
            /tikz/every even column/.append style={column sep=0.3cm}
            },  
  log ticks with fixed point,
  xlabel style={yshift=1ex},
  xtick={10,20,50,100,200,400, 1000},
  xmode=log,
  ]

    \addlegendimage{ourssmall} \addlegendentry{$[10,50]$};
    \addlegendimage{oursmed} \addlegendentry{$[10,100]$};
    \addlegendimage{oursbin} \addlegendentry{$[10,400]$};
    \addlegendimage{oursbig} \addlegendentry{$[10,600]$};

    \addplot[ourssmall] coordinates {(10,33.58)
                                    (20,33.63)
                                    (50,33.97)
                                    (100,34.13)
                                    (200,33.95)
                                    (400,33.85)
                                    (600,33.82)
                                    (1000,33.66)
                                    (1600,33.55)};

    \addplot[oursmed] coordinates {(10,33.55)
                                   (20,33.69)
                                   (50,34.13)
                                   (100,34.36)
                                   (200,34.67)
                                   (400,34.64)
                                   (600,34.54)
                                   (1000,34.37)
                                   (1600,34.17)};

    \addplot[oursbin] coordinates {(10,33.72)
                                   (20,33.86)
                                   (50,34.32)
                                   (100,34.71)
                                   (200,34.96)
                                   (400,35.13)
                                   (600,35.13)
                                   (1000,35.22)
                                   (1600,35.26)};

    \addplot[oursbig] coordinates {(10,33.66)
                                   (20,33.84)
                                   (50,34.21)
                                   (100,34.65)
                                   (200,34.86)
                                   (400,35.02)
                                   (600,35.08)
                                   (1000,35.19)
                                   (1600,35.24)};

\end{axis}
\end{tikzpicture}

%% file: fig/dim_revop.tex
\pgfmathsetmacro{\teasermarkersize}{2.5}
\begin{tikzpicture}
\begin{axis}[%
  width=0.50\linewidth,
  height=0.4\linewidth,
  ylabel={\small mAP},
  xlabel={\small memory per image (KB)},
  tick label style={font=\small},
  ylabel near ticks, xlabel near ticks, 
  grid=minor,
  xlabel style={yshift=1ex},
  xmode=log,
  ]

    \addplot[globalfull] coordinates {(0.25, 72.82)};

    \addplot[oursbig] coordinates {(0.5625,74.01)    %
                                (0.875,74.78)     %
                                (1.8125,76.1)     %
                                (3.375,76.85)     %
                                (6.5,77.62)       %
                                (12.75,78.24)     %
                                (19,78.69)};      %

    \addplot[oursbin] coordinates {(0.40625,73.89)   %
                                   (0.5625,74.36)    %
                                   (1.03125,75.49)   %
                                   (1.8125,76.28)    %
                                   (3.375,76.82)     %
                                   (6.5,77.44)       %
                                   (9.625,77.46)};   %
    \addplot[oursmed] coordinates {(0.328125,73.22) %
                                (0.40625,73.73) %
                                (0.640625,74.64) %
                                (1.03125,75.22) %
                                (1.8125,75.59) %
                                (3.375,75.91) %
                                (4.9375 ,75.91)}; %

\end{axis}
\end{tikzpicture}

%% file: fig/dim_gld.tex
\pgfmathsetmacro{\teasermarkersize}{2.5}
\begin{tikzpicture}
\begin{axis}[%
  width=0.50\linewidth,
  height=0.4\linewidth,
  ylabel={\small mAP@100},
  xlabel={\small memory per image (KB)},
  tick label style={font=\small},
  ylabel near ticks, xlabel near ticks, 
  legend columns=4,
  legend style={
            at={(-0.18,1.18)},
            anchor=north,
            only marks,
            /tikz/every even column/.append style={column sep=0.3cm}
            },  
  grid=minor,
  xlabel style={yshift=1ex},
  xmode=log,
  ]

    \addlegendimage{globalfull} \addlegendentry{Global};
    \addlegendimage{oursmed} \addlegendentry{\ours $d$=64};
    \addlegendimage{oursbin} \addlegendentry{\ours $d$=128};
    \addlegendimage{oursbig} \addlegendentry{\ours $d$=256};
    
    \addplot[globalfull] coordinates {(0.25, 33.37)};

    \addplot[oursbig] coordinates {(0.5625,33.65)    %
                                (0.875,33.88)     %
                                (1.8125,34.43)    %
                                (3.375,34.86)     %
                                (6.5,35.07)       %
                                (12.75,35.4)      %
                                (19,35.52)};      %

    \addplot[oursbin] coordinates {(0.40625,33.72)   %
                                   (0.5625,33.86)    %
                                   (1.03125,34.32)   %
                                   (1.8125,34.71)    %
                                   (3.375,34.96)     %
                                   (6.5,35.13)       %
                                   (9.625,35.13)};   %
    \addplot[oursmed] coordinates {(0.328125,33.47)    %
                                (0.40625,33.65)    %
                                (0.640625,33.87)   %
                                (1.03125,34.12)    %
                                (1.8125,34.39)     %
                                (3.375,34.59)      %
                                (4.9375 ,34.69)};  %

\end{axis}
\end{tikzpicture}

%% file: tab/dist_beta.tex
\small
\begin{tabular}{l l c c}
    \toprule
    \textbf{\l2 loss} & \textbf{$\beta$} & \textbf{$\mathcal{R}$OP+1M} & \textbf{GLDv2} \\ 
    \midrule
    --                                & 0   & 75.5 & 34.3 \\ \midrule
    \multirow{3}{*}{$Z_N$}            & 1   & 76.3 & 34.6 \\
                                      & 10  & 76.4 & 34.7 \\ 
                                      & 100 & 76.3 & 34.5 \\ \midrule
    \multirow{3}{*}{$S$}              & 1   & 75.7 & 34.4 \\
                                      & 10  & 75.9 & 34.4 \\ 
                                      & 100 & 75.9 & 34.4 \\
    \bottomrule
\end{tabular}

%% file: tab/model_size.tex
\small
\begin{tabular}{l c c}
    \toprule
    \textbf{$N$} & \textbf{$\mathcal{R}$OP+1M} & \textbf{GLDv2} \\ 
    \midrule
    0  & 72.8 & 33.3 \\
    1  & 73.0 & 33.3 \\
    3  & 75.1 & 34.1 \\ 
    5  & 75.5 & 34.3 \\
    7  & 75.6 & 34.3 \\
    \bottomrule
\end{tabular}

%% file: tab/binarization.tex
\def\arraystretch{.9}
\small
\begin{tabular}{c c c c }
    \toprule
    \textbf{ITQ} & \textbf{FT} & \textbf{$\mathcal{R}$OP+1M} & \textbf{GLDv2} \\ \midrule
    -- & --                  & 74.6 & 34.1 \\
    -- & \checkmark          & 74.8 & 34.0 \\
    \checkmark & --          & 75.1 & 34.3 \\
    \checkmark & \checkmark  & 75.5 & 34.3 \\
    \bottomrule
\end{tabular}

%% file: tab/600-50_std.tex
\setlength\tabcolsep{4pt}
\small
\begin{tabular}{l l  c c c  c c c c  c c c c  c}
    \toprule
    &  &  &  &  &\multicolumn{4}{c}{\textbf{Medium}} & \multicolumn{4}{c}{\textbf{Hard}} & \multirow{2}{*}{\textbf{GLDv2}} \\ \cmidrule(lr){6-9} \cmidrule(lr){10-13}
    \Lqtest & \Lxtest & bin & dist & global & $\mathcal{R}$Oxf & +1M & $\mathcal{R}$Par & +1M & $\mathcal{R}$Oxf & +1M & $\mathcal{R}$Par & +1M & \\ 
    \midrule 
    600 & 600 & \checkmark & \checkmark & PQ8  & 88.3\scriptsize{$\pm0.4$} & 83.5\scriptsize{$\pm0.4$} & 93.2\scriptsize{$\pm0.2$} & 86.9\scriptsize{$\pm0.1$} & 77.0\scriptsize{$\pm0.7$} & 69.6\scriptsize{$\pm0.5$} & 86.1\scriptsize{$\pm0.4$} & 75.7\scriptsize{$\pm0.1$} & 35.6\scriptsize{$\pm0.1$} \\
    600 & 50  & \checkmark & \checkmark & PQ8  & 87.0\scriptsize{$\pm0.3$} & 81.1\scriptsize{$\pm0.2$} & 92.6\scriptsize{$\pm0.0$} & 85.4\scriptsize{$\pm0.0$} & 75.0\scriptsize{$\pm0.5$} & 66.0\scriptsize{$\pm0.3$} & 84.7\scriptsize{$\pm0.0$} & 72.9\scriptsize{$\pm0.1$} & 34.7\scriptsize{$\pm0.1$} \\
    50 &  50  & \checkmark & \checkmark & PQ8 & 86.9\scriptsize{$\pm0.2$} & 80.9\scriptsize{$\pm0.3$} & 92.4\scriptsize{$\pm0.1$} & 85.0\scriptsize{$\pm0.1$} & 74.7\scriptsize{$\pm0.4$} & 65.4\scriptsize{$\pm0.4$} & 84.3\scriptsize{$\pm0.1$} & 72.0\scriptsize{$\pm0.2$} & 34.3\scriptsize{$\pm0.1$} \\
    \midrule
    600 &  600 & \checkmark & -- & PQ8 & 87.7\scriptsize{$\pm0.4$} & 82.4\scriptsize{$\pm0.3$} & 92.7\scriptsize{$\pm0.3$} & 86.2\scriptsize{$\pm0.4$} & 75.6\scriptsize{$\pm0.7$} & 67.3\scriptsize{$\pm0.5$} & 85.1\scriptsize{$\pm0.6$} & 74.0\scriptsize{$\pm0.8$} & 35.1\scriptsize{$\pm0.0$} \\        
    600 &  50  & \checkmark & -- & PQ8 & 86.7\scriptsize{$\pm0.5$} & 80.6\scriptsize{$\pm0.3$} & 92.2\scriptsize{$\pm0.2$} & 84.7\scriptsize{$\pm0.2$} & 74.4\scriptsize{$\pm0.9$} & 65.2\scriptsize{$\pm0.5$} & 83.9\scriptsize{$\pm0.4$} & 71.4\scriptsize{$\pm0.5$} & 34.3\scriptsize{$\pm0.1$} \\        
    50  &  50  & \checkmark & -- & PQ8 & 86.5\scriptsize{$\pm0.2$} & 80.2\scriptsize{$\pm0.3$} & 92.0\scriptsize{$\pm0.2$} & 84.2\scriptsize{$\pm0.2$} & 73.9\scriptsize{$\pm0.3$} & 64.4\scriptsize{$\pm0.3$} & 83.4\scriptsize{$\pm0.3$} & 70.5\scriptsize{$\pm0.4$} & 34.0\scriptsize{$\pm0.0$} \\        
    \midrule
    600 &  600  & -- & -- & PQ8 & 89.3\scriptsize{$\pm0.2$} & 84.7\scriptsize{$\pm0.4$} & 93.3\scriptsize{$\pm0.0$} & 87.2\scriptsize{$\pm0.2$} & 78.1\scriptsize{$\pm1.0$} & 71.5\scriptsize{$\pm0.7$} & 86.6\scriptsize{$\pm0.0$} & 76.5\scriptsize{$\pm0.4$} & 35.8\scriptsize{$\pm0.3$} \\               
    600 &  50   & -- & -- & PQ8 & 87.2\scriptsize{$\pm0.2$} & 81.5\scriptsize{$\pm0.2$} & 92.6\scriptsize{$\pm0.1$} & 85.4\scriptsize{$\pm0.1$} & 75.5\scriptsize{$\pm0.8$} & 66.9\scriptsize{$\pm0.7$} & 84.8\scriptsize{$\pm0.3$} & 73.1\scriptsize{$\pm0.4$} & 34.8\scriptsize{$\pm0.2$} \\               
    50  &  50   & -- & -- & PQ8 & 86.8\scriptsize{$\pm0.2$} & 80.7\scriptsize{$\pm0.1$} & 92.3\scriptsize{$\pm0.1$} & 85.0\scriptsize{$\pm0.1$} & 74.4\scriptsize{$\pm0.7$} & 65.1\scriptsize{$\pm0.4$} & 84.1\scriptsize{$\pm0.2$} & 71.9\scriptsize{$\pm0.2$} & 34.4\scriptsize{$\pm0.1$} \\           
    \midrule                                                                                                                                           
    600 & 600 & -- & -- & full & 89.1\scriptsize{$\pm0.1$} & 84.4\scriptsize{$\pm0.4$} & 93.2\scriptsize{$\pm0.1$} & 87.1\scriptsize{$\pm0.2$} & 78.3\scriptsize{$\pm0.3$} & 71.2\scriptsize{$\pm0.5$} & 86.3\scriptsize{$\pm0.2$} & 76.2\scriptsize{$\pm0.3$} & 35.9\scriptsize{$\pm0.2$} \\
    600 & 50  & -- & -- & full & 87.3\scriptsize{$\pm0.3$} & 81.3\scriptsize{$\pm0.2$} & 93.0\scriptsize{$\pm0.1$} & 85.6\scriptsize{$\pm0.2$} & 75.8\scriptsize{$\pm0.8$} & 66.2\scriptsize{$\pm0.7$} & 85.6\scriptsize{$\pm0.1$} & 73.3\scriptsize{$\pm0.2$} & 34.7\scriptsize{$\pm0.2$} \\
    50  & 50  & -- & -- & full & 87.3\scriptsize{$\pm0.2$} & 80.9\scriptsize{$\pm0.0$} & 92.6\scriptsize{$\pm0.0$} & 85.0\scriptsize{$\pm0.1$} & 75.5\scriptsize{$\pm0.6$} & 65.5\scriptsize{$\pm0.2$} & 84.7\scriptsize{$\pm0.2$} & 72.0\scriptsize{$\pm0.2$} & 34.4\scriptsize{$\pm0.1$} \\             
    \bottomrule
\end{tabular}

%% file: tab/sota_std.tex
\setlength\tabcolsep{4pt}
\small
\begin{tabular}{l l c @{\ssp} c @{\ssp} c  c c c c c  c c c c  c}
    \toprule
    \multirow{2}{*}{\textbf{global desc.}} & \multirow{2}{*}{\textbf{local desc.}} & \multicolumn{3}{c}{\textbf{re-ranking}} &\multicolumn{4}{c}{\textbf{Medium}} & \multicolumn{4}{c}{\textbf{Hard}} & \multirow{2}{*}{\textbf{GLDv2}} \\ \cmidrule(lr){3-5} \cmidrule(lr){6-9} \cmidrule(lr){10-13}
    & & \textbf{bin} & \textbf{dist} & \textbf{top}-$m$ & $\mathcal{R}$Oxf & +1M & $\mathcal{R}$Par & +1M & $\mathcal{R}$Oxf & +1M & $\mathcal{R}$Par & +1M & \\ 
    \midrule 
        \multirow{4}{*}{\textbf{CVNet}~\cite{lsl+22}} & \multirow{4}{*}{\textbf{CVNet}~\cite{lsl+22}}
        & -- & -- &                   100 & 84.9\scriptsize{$\pm0.2$} & 78.6\scriptsize{$\pm0.2$} & 90.6\scriptsize{$\pm0.0$} & 81.3\scriptsize{$\pm0.0$} & 71.1\scriptsize{$\pm0.5$} & 62.4\scriptsize{$\pm0.7$} & 81.6\scriptsize{$\pm0.0$} & 66.5\scriptsize{$\pm0.1$} & 35.3\scriptsize{$\pm0.2$} \\
        & & \checkmark & \checkmark & 100 & 84.6\scriptsize{$\pm0.7$} & 78.4\scriptsize{$\pm0.6$} & 90.6\scriptsize{$\pm0.0$} & 81.3\scriptsize{$\pm0.0$} & 70.8\scriptsize{$\pm1.4$} & 61.9\scriptsize{$\pm1.0$} & 81.5\scriptsize{$\pm0.1$} & 66.3\scriptsize{$\pm0.2$} & 35.0\scriptsize{$\pm0.2$} \\
        & & -- & -- &                 400 & 86.8\scriptsize{$\pm0.3$} & 81.3\scriptsize{$\pm0.2$} & 91.9\scriptsize{$\pm0.0$} & 84.1\scriptsize{$\pm0.1$} & 74.2\scriptsize{$\pm0.6$} & 66.3\scriptsize{$\pm0.5$} & 84.1\scriptsize{$\pm0.1$} & 71.5\scriptsize{$\pm0.1$} & 35.5\scriptsize{$\pm0.2$} \\
        & & \checkmark & \checkmark & 400 & 86.3\scriptsize{$\pm1.0$} & 80.9\scriptsize{$\pm0.8$} & 91.9\scriptsize{$\pm0.1$} & 84.0\scriptsize{$\pm0.0$} & 73.7\scriptsize{$\pm1.6$} & 65.5\scriptsize{$\pm1.2$} & 84.1\scriptsize{$\pm0.3$} & 71.2\scriptsize{$\pm0.3$} & 35.1\scriptsize{$\pm0.2$} \\
        \midrule
        \multirow{9}{*}{\textbf{SG}~\cite{sck+23}}
        & \multirow{2}{*}{\textbf{CVNet}~\cite{lsl+22}}
        & -- & -- &                   1600 & 89.1\scriptsize{$\pm0.1$} & 84.4\scriptsize{$\pm0.4$} & 93.2\scriptsize{$\pm0.1$} & 87.1\scriptsize{$\pm0.2$} & 78.3\scriptsize{$\pm0.3$} & 71.2\scriptsize{$\pm0.5$} & 86.3\scriptsize{$\pm0.2$} & 76.2\scriptsize{$\pm0.3$} & 35.9\scriptsize{$\pm0.2$} \\
        & & \checkmark & \checkmark & 1600 & 88.5\scriptsize{$\pm0.4$} & 83.6\scriptsize{$\pm0.4$} & 93.2\scriptsize{$\pm0.2$} & 87.0\scriptsize{$\pm0.1$} & 77.2\scriptsize{$\pm0.9$} & 69.8\scriptsize{$\pm0.6$} & 86.3\scriptsize{$\pm0.4$} & 75.9\scriptsize{$\pm0.3$} & 35.5\scriptsize{$\pm0.2$} \\
        & \multirow{2}{*}{\textbf{DINOv2}~\cite{odm+24}}
        & -- & -- &                   1600 & 92.4\scriptsize{$\pm0.9$} & 87.1\scriptsize{$\pm0.5$} & 95.2\scriptsize{$\pm0.1$} & 89.8\scriptsize{$\pm0.1$} & 83.1\scriptsize{$\pm1.1$} & 76.1\scriptsize{$\pm0.7$} & 90.2\scriptsize{$\pm0.4$} & 81.0\scriptsize{$\pm0.3$} & 38.3\scriptsize{$\pm0.2$} \\
        & & \checkmark & \checkmark & 1600 & 90.7\scriptsize{$\pm0.3$} & 85.1\scriptsize{$\pm0.2$} & 94.9\scriptsize{$\pm0.1$} & 89.3\scriptsize{$\pm0.2$} & 80.0\scriptsize{$\pm0.9$} & 72.6\scriptsize{$\pm0.8$} & 89.7\scriptsize{$\pm0.2$} & 80.0\scriptsize{$\pm0.4$} & 37.8\scriptsize{$\pm0.0$} \\
        \cmidrule{2-14}
        & \multicolumn{13}{c}{\textbf{with SG re-rank}~\cite{sck+23}} \\
        \cmidrule{2-14}
        & \multirow{2}{*}{\textbf{CVNet}~\cite{lsl+22}}
        & -- & -- &                   1600 & 91.1\scriptsize{$\pm0.2$} & 86.6\scriptsize{$\pm0.4$} & 94.3\scriptsize{$\pm0.1$} & 88.8\scriptsize{$\pm0.1$} & 80.4\scriptsize{$\pm0.7$} & 74.1\scriptsize{$\pm0.6$} & 88.6\scriptsize{$\pm0.3$} & 79.9\scriptsize{$\pm0.2$} & 36.0\scriptsize{$\pm0.2$} \\
        & & \checkmark & \checkmark & 1600 & 90.7\scriptsize{$\pm0.3$} & 85.9\scriptsize{$\pm0.3$} & 94.3\scriptsize{$\pm0.1$} & 88.9\scriptsize{$\pm0.1$} & 79.4\scriptsize{$\pm0.7$} & 72.9\scriptsize{$\pm0.6$} & 88.7\scriptsize{$\pm0.4$} & 79.8\scriptsize{$\pm0.2$} & 35.8\scriptsize{$\pm0.1$} \\
        & \multirow{2}{*}{\textbf{DINOv2}~\cite{odm+24}}
        & -- & -- &                   1600 & 93.6\scriptsize{$\pm0.6$} & 88.2\scriptsize{$\pm0.4$} & 95.3\scriptsize{$\pm0.1$} & 90.1\scriptsize{$\pm0.1$} & 84.8\scriptsize{$\pm0.9$} & 77.7\scriptsize{$\pm0.8$} & 90.7\scriptsize{$\pm0.4$} & 82.0\scriptsize{$\pm0.2$} & 38.5\scriptsize{$\pm0.2$} \\
        & & \checkmark & \checkmark & 1600 & 92.7\scriptsize{$\pm0.3$} & 86.7\scriptsize{$\pm0.1$} & 95.2\scriptsize{$\pm0.2$} & 89.8\scriptsize{$\pm0.2$} & 83.0\scriptsize{$\pm0.6$} & 75.4\scriptsize{$\pm0.5$} & 90.4\scriptsize{$\pm0.3$} & 81.4\scriptsize{$\pm0.5$} & 38.0\scriptsize{$\pm0.1$} \\
    \bottomrule
\end{tabular}

%% file: fig/topk_revop.tex
\pgfplotstableread{
		l		ours   ours-l0  rtf-l0  rtf
        50      72.80  72.80    72.80   72.80   
		100		74.65  71.18    71.79   74.24
		200		75.35  70.73    70.81   74.67
		400		75.90  69.59    69.33   74.96
		800		76.33  68.08    66.88   75.17 
		1600	76.54  66.23    64.70   75.26 
        3200    76.64  64.23    62.73   75.30
        6400    76.67  62.25    60.99   75.30
	}{\topk}

\pgfmathsetmacro{\teasermarkersize}{2.5}
\begin{tikzpicture}
\begin{axis}[%
  width=.7\linewidth,
  height=0.45\linewidth,
  ylabel={\small mAP},
  xlabel={\small $m$},
  tick label style={font=\scriptsize},
  legend pos=south west,
  legend style = {opacity = 0.75, row sep = -2pt, font=\scriptsize},
  ylabel near ticks, xlabel near ticks, 
  xlabel style={yshift=1ex},
  xmode=log,
  log ticks with fixed point,
  xtick={50,100,200,400,800,1600,3200,6400},
  xticklabels = {0,100,200,400,800,1600,3200,6400},
  ]

    \addplot[mark options={draw=black}, color=\rtfcol, solid, mark=*,  mark size=2., line width=1.0] table[x=l, y expr={\thisrow{rtf}}] \topk;

    \addplot[mark options={draw=black}, color=\rtfcol, solid, mark=\binmark,  mark size=2.5, line width=1.0] table[x=l, y expr={\thisrow{rtf-l0}}] \topk;
    
    \addplot[mark options={draw=black}, color=\ourscol, solid, mark=*,  mark size=2., line width=1.0] table[x=l, y expr={\thisrow{ours}}]  \topk;

    \addplot[mark options={draw=black}, color=\ourscol, solid, mark=\binmark,  mark size=2.5, line width=1.0] table[x=l, y expr={\thisrow{ours-l0}}] \topk;
    
    \addlegendimage{rtf} \addlegendentry{R2F - ensemble};
    \addlegendimage{rtf-l0} \addlegendentry{R2F - local only};
    
    \addlegendimage{ours} \addlegendentry{\ours~- ensemble};
    \addlegendimage{ours-l0} \addlegendentry{\ours~- local only};
    
\end{axis}
\end{tikzpicture}

%% file: fig/sa_vs_ca_revop.tex
\pgfplotsset{
compat=1.11,
legend image code/.code={
\draw[mark repeat=2,mark phase=2]
plot coordinates {
(0cm,0cm)
(0.15cm,0cm)        %
(0.3cm,0cm)         %
};%
}
}
\pgfmathsetmacro{\teasermarkersize}{2.5}
\begin{tikzpicture}
\begin{axis}[%
  width=0.7\linewidth,
  height=0.45\linewidth,
  ylabel={\small mAP},
  xlabel={ \small local descriptors number (\Lxtest) },
  tick label style={font=\scriptsize},
  ylabel near ticks, xlabel near ticks, 
  legend pos=north west,
  legend style={opacity = .7,
            inner sep=0pt,
            cells={anchor=west},
            only marks,
            legend entries={only marks,{only marks, sharp plot}},
            font=\scriptsize,
            row sep = -2pt,
            },
  log ticks with fixed point,
  xlabel style={yshift=1ex},
  xtick={10,20,50,100,200,400},
  xmode=log,
  ]
    \addlegendimage{rrt-univ} \addlegendentry{RRT w/o global};
    \addlegendimage{ours-univ} \addlegendentry{\ours};

   \addplot[rrt-univ] coordinates  {(10,73.60)
                                    (20,73.88)
                                    (50,74.93)
                                    (100,75.70)
                                    (200,76.15)
                                    (400,76.54)
                                    (600,76.82)};
                          
   \addplot[ours-univ] coordinates {(10,73.89)
                                    (20,74.36)
                                    (50,75.49)
                                    (100,76.28)
                                    (200,76.82)
                                    (400,77.44)
                                    (600,77.46)};

\end{axis}
\end{tikzpicture}

%% file: fig/sim_global_local/q56_OxfH+1m.tex
\pgfmathsetmacro{\teasermarkersize}{2.2}
\begin{tikzpicture}
\begin{axis}[%
  scale=0.46,
  ylabel={\scriptsize global similarity},
  tick label style={font=\scriptsize},
  ylabel near ticks, xlabel near ticks, 
  grid=none,
  legend pos=north west,
  title={query=56, $\Delta$AP $=10.8$},
  title style={yshift=-2.0ex, font=\scriptsize},
  legend style={font=\scriptsize}
  ]
  
    \addplot[only marks, mark options={draw=black}, color=\ourscol, solid, mark=*,  mark size=2.2] table[x expr={\thisrow{localneg}}, y expr={\thisrow{globalneg}}]  \queryfivesix;
    \addlegendentry{non-matching}
    
	\addplot[only marks, mark options={draw=black}, color=\rtfcol, solid, mark=*,  mark size=2.2] table[x expr={\thisrow{localpos}}, y expr={\thisrow{globalpos}}]  \queryfivesix;
    \addlegendentry{matching}
 
\end{axis}
\end{tikzpicture}

%% file: fig/sim_global_local/q26_OxfH+1m.tex
\begin{tikzpicture}
\begin{axis}[%
  scale=0.46,
  tick label style={font=\scriptsize},
  ylabel near ticks, xlabel near ticks, 
  grid=none,
  legend pos=north west,
  title={query=26, $\Delta$AP $=14.1$},
  title style={yshift=-2.0ex, font=\scriptsize},
  ]
    
    \addplot[mark options={draw=black}, fill opacity=1.0, color=\ourscol, solid, mark=*,  mark size=2.2, draw=none] table[x expr={\thisrow{localneg}}, y expr={\thisrow{globalneg}}]  \querytwosix;
    
	\addplot[mark options={draw=black}, fill opacity=1.0, color=\rtfcol, solid, mark=*,  mark size=2.2, draw=none] table[x expr={\thisrow{localpos}}, y expr={\thisrow{globalpos}}]  \querytwosix;
 
\end{axis}
\end{tikzpicture}

%% file: fig/sim_global_local/q65_OxfH+1m.tex
\pgfmathsetmacro{\teasermarkersize}{2.2}
\begin{tikzpicture}
\begin{axis}[%
  scale=0.46,
  ylabel={\scriptsize global similarity},
  tick label style={font=\scriptsize},
  ylabel near ticks, xlabel near ticks,
  legend pos=north west,
  grid=none,
  title={query=65, $\Delta$AP $=28.8$},
  title style={yshift=-2.0ex, font=\scriptsize},
  ]
    \addplot[only marks, mark options={draw=black}, color=\ourscol, solid, mark=*,  mark size=2.2] table[x expr={\thisrow{localneg}}, y expr={\thisrow{globalneg}}]  \querysixfive;
    
	\addplot[only marks, mark options={draw=black}, color=\rtfcol, solid, mark=*,  mark size=2.2] table[x expr={\thisrow{localpos}}, y expr={\thisrow{globalpos}}]  \querysixfive; 

 
\end{axis}
\end{tikzpicture}

%% file: fig/sim_global_local/q60_OxfH+1m.tex
\pgfmathsetmacro{\teasermarkersize}{2.2}
\begin{tikzpicture}
\begin{axis}[%
  scale=0.46,
  tick label style={font=\scriptsize},
  ylabel near ticks, xlabel near ticks, 
  grid=none,
  legend pos=north west,
  title={query=60, $\Delta$AP $=-4.1$},
  title style={yshift=-2.0ex, font=\scriptsize},
  ]
    
    \addplot[mark options={draw=black}, fill opacity=1.0, color=\ourscol, solid, mark=*,  mark size=2.2, draw=none] table[x expr={\thisrow{localneg}}, y expr={\thisrow{globalneg}}]  \querysixzero;
    
	\addplot[mark options={draw=black}, fill opacity=1.0, color=\rtfcol, solid, mark=*,  mark size=2.2, draw=none] table[x expr={\thisrow{localpos}}, y expr={\thisrow{globalpos}}]  \querysixzero;
 
\end{axis}
\end{tikzpicture}

%% file: fig/sim_global_local/q18_OxfH+1m.tex
\pgfmathsetmacro{\teasermarkersize}{2.2}
\begin{tikzpicture}
\begin{axis}[%
  scale=0.46,
  xlabel={\scriptsize local similarity},
  ylabel={\scriptsize global similarity},
  tick label style={font=\scriptsize},
  ylabel near ticks, xlabel near ticks, 
  grid=none,
  legend pos=north west,
  title={query=18, $\Delta$AP $=10.5$},
  title style={yshift=-2.0ex, font=\scriptsize},
  ]
    
    \addplot[mark options={draw=black}, fill opacity=1.0, color=\ourscol, solid, mark=*,  mark size=2.2, draw=none] table[x expr={\thisrow{localneg}}, y expr={\thisrow{globalneg}}]  \queryoneeight;
    
	\addplot[mark options={draw=black}, fill opacity=1.0, color=\rtfcol, solid, mark=*,  mark size=2.2, draw=none] table[x expr={\thisrow{localpos}}, y expr={\thisrow{globalpos}}]  \queryoneeight;
 
\end{axis}
\end{tikzpicture}

%% file: fig/sim_global_local/q69_OxfH+1m.tex
\pgfmathsetmacro{\teasermarkersize}{2.2}
\begin{tikzpicture}
\begin{axis}[%
  scale=0.46,
  tick label style={font=\scriptsize},
  ylabel near ticks, xlabel near ticks, 
  xlabel={\scriptsize local similarity},
  grid=none,
  legend pos=north west,
  title={query=69, $\Delta$AP $=6.7$},
  title style={yshift=-2.0ex, font=\scriptsize}
  ]
    
    \addplot[mark options={draw=black}, fill opacity=1.0, color=\ourscol, solid, mark=*,  mark size=2.2, draw=none] table[x expr={\thisrow{localneg}}, y expr={\thisrow{globalneg}}]  \querysixnine;
    
	\addplot[mark options={draw=black}, fill opacity=1.0, color=\rtfcol, solid, mark=*,  mark size=2.2, draw=none] table[x expr={\thisrow{localpos}}, y expr={\thisrow{globalpos}}]  \querysixnine;
 
\end{axis}
\end{tikzpicture}

%% file: tab/hyperparam_tuning.tex
\footnotesize
\def\arraystretch{1.4}%
\newcommand*{\MinNumber}{26.0}
\newcommand*{\MaxNumber}{33.9}
\newcommand{\col}[1]{%
    \pgfmathsetmacro{\PercentColor}{100.0*((#1-\MinNumber)^(2)+0.0*(\MaxNumber-\MinNumber))/((\MaxNumber-\MinNumber)^(2)+0.0*(\MaxNumber-\MinNumber))}%
    \xdef\PercentColor{\PercentColor}%
    \cellcolor{\rtfcol!\PercentColor}{#1}%
}
\scalebox{0.67}{
\tabcolsep=3pt
\begin{tabular}{@{\ssp}l@{\msp}r@{\ssp}|c@{\msp}c@{\msp}c@{\msp}c@{\msp}c@{\msp}c@{\msp}c@{\msp}c@{\msp}c@{\msp}c@{\msp}c@{\msp}c@{\msp}c@{\msp}c@{\msp}c@{\msp}c@{\msp}c@{\msp}c@{\msp}c@{\msp}c@{\msp}c|}
    \multicolumn{2}{c}{} & \multicolumn{20}{c}{\large$\lambda$} \\
    \multicolumn{2}{c}{} & 0.00 & 0.05 & 0.10 & 0.15 & 0.20 & 0.25 & 0.30 & 0.35 & 0.40 & 0.45 & 0.50 & 0.55 & 0.60 & 0.65 & 0.70 & 0.75 & 0.80 & 0.85 & 0.90 & 0.95 & \multicolumn{1}{c}{\hspace{-3pt}1.00} \\ 
    \cline{3-23}
    \cline{3-23}
    \multirow{6}{*}{\large$\gamma$}
    & 1e-4 & \col{28.8} & \col{31.0} & \col{31.0} & \col{31.0} & \col{31.0} & \col{31.0} & \col{31.0} & \col{31.0} & \col{31.0} & \col{31.0} & \col{31.0} & \col{31.0} & \col{31.0} & \col{31.0} & \col{31.0} & \col{31.0} & \col{31.0} & \col{31.0} & \col{31.0} & \col{31.0} & \col{31.0} \\
    & 1e-3 & \col{28.7} & \col{31.9} & \col{31.4} & \col{31.3} & \col{31.2} & \col{31.2} & \col{31.1} & \col{31.0} & \col{31.0} & \col{31.1} & \col{31.0} & \col{31.0} & \col{31.0} & \col{31.0} & \col{31.0} & \col{31.0} & \col{31.0} & \col{31.0} & \col{31.0} & \col{31.0} & \col{31.0} \\
    & 1e-2 & \col{28.7} & \col{32.6} & \col{32.9} & \col{32.6} & \col{32.3} & \col{32.1} & \col{32.0} & \col{31.9} & \col{31.8} & \col{31.6} & \col{31.5} & \col{31.4} & \col{31.4} & \col{31.3} & \col{31.2} & \col{31.2} & \col{31.1} & \col{31.0} & \col{31.1} & \col{31.0} & \col{31.0} \\
    & 1e-1 & \col{28.7} & \col{29.7} & \col{30.3} & \col{31.1} & \col{31.7} & \col{32.1} & \col{32.4} & \col{32.7} & \col{32.9} & \col{32.9} & \col{33.0} & \col{32.9} & \col{32.7} & \col{32.5} & \col{32.3} & \col{32.2} & \col{32.0} & \col{31.9} & \col{31.5} & \col{31.3} & \col{31.0} \\
    & 1e0 & \col{28.7} & \col{28.9} & \col{29.1} & \col{29.2} & \col{29.4} & \col{29.7} & \col{29.9} & \col{30.0} & \col{30.2} & \col{30.3} & \col{30.6} & \col{31.0} & \col{31.4} & \col{31.8} & \col{32.2} & \col{32.4} & \col{32.6} & \col{32.8} & \col{32.7} & \col{32.1} & \col{31.0} \\ 
    & 1e1 & \col{27.6} & \col{30.3} & \col{30.4} & \col{30.5} & \col{30.6} & \col{30.7} & \col{30.8} & \col{30.8} & \col{30.9} & \col{30.9} & \col{30.9} & \col{31.0} & \col{31.0} & \col{31.1} & \col{31.2} & \col{31.3} & \col{31.5} & \col{31.9} & \col{32.2} & \col{32.2} & \col{31.0} \\ 
    \cline{3-23} 
\end{tabular}
}